\documentclass[twoside,11pt]{article}

\usepackage{microtype}
\usepackage{graphicx}
\usepackage{subfigure}
\usepackage{booktabs} 
\usepackage{amsmath,amssymb,amsthm,bm}
\usepackage{subfigure}
\usepackage{algorithmic}
\usepackage{algorithm}

\usepackage{multirow}

\usepackage[abbrvbib, preprint]{jmlr2e}

\usepackage[misc]{ifsym}

\allowdisplaybreaks[4]

\def\g{{\bf g}}
\def\h{{\bf h}}

\def\u{{\bf u}}

\def\w{{\bf w}}
\def\x{{\bf x}}
\def\y{{\bf y}}

\def\G{{\bf G}}

\def\I{{\bf I}}

\def\0{{\bf 0}}
\def\1{{\bf 1}}
\def\2{{\bf 2}}
\def\3{{\bf 3}}
\def\4{{\bf 4}}
\def\5{{\bf 5}}
\def\6{{\bf 6}}
\def\7{{\bf 7}}
\def\8{{\bf 8}}
\def\9{{\bf 9}}

\def\DM{{\mathcal D}}

\def\HM{{\mathcal H}}

\def\LM{{\mathcal L}}
\def\MM{{\mathcal M}}
\def\NM{{\mathcal N}}

\def\SM{{\mathcal S}}

\def\EB{{\mathbb E}}

\def\RB{{\mathbb R}}

\def\ZB{{\mathbb Z}}

\newtheorem{assumption}{Assumption}

\ShortHeadings{Buffered Asynchronous SGD for Byzantine Learning}{Yang and Li}
\firstpageno{1}

\begin{document}

\title{Buffered Asynchronous SGD for Byzantine Learning}

\author{\name Yi-Rui Yang \email yangyr@smail.nju.edu.cn \\
        \name \Letter\  Wu-Jun Li \email liwujun@nju.edu.cn \\
       \addr National Key Laboratory for Novel Software Technology\\
       Department of Computer Science and Technology\\
       Nanjing University, Nanjing 210023, China}

\editor{Kevin Murphy and Bernhard Sch{\"o}lkopf}

\maketitle

\begin{abstract}
  Distributed learning has become a hot research topic due to its wide application in cluster-based large-scale learning, federated learning, edge computing and so on. Most traditional distributed learning methods typically assume no failure or attack. However, many unexpected cases, such as communication failure and even malicious attack, may happen in real applications. Hence, Byzantine learning~(BL), which refers to distributed learning with failure or attack, has recently attracted much attention. Most existing BL methods are synchronous, which are impractical in some applications due to heterogeneous or offline workers. In these cases, asynchronous BL~(ABL) is usually preferred. 
    In this paper, we propose a novel method, called \underline{b}uffered \underline{a}synchronous \underline{s}tochastic \underline{g}radient \underline{d}escent~(\mbox{BASGD}), for ABL. 
    To the best of our knowledge, BASGD is the first ABL method that can resist non-omniscient attacks without storing any instances on server. Furthermore, we also propose an improved variant of BASGD, called BASGD with momentum (BASGDm), by introducing momentum into BASGD. BASGDm can resist both non-omniscient and omniscient attacks. Compared with those methods which need to store instances on server, \mbox{BASGD} and \mbox{BASGDm} have a wider scope of application.
    Both \mbox{BASGD} and \mbox{BASGDm} are compatible with various aggregation rules. Moreover, both \mbox{BASGD} and \mbox{BASGDm} are proved to be convergent and be able to resist failure or attack.
    Empirical results show that our methods significantly outperform existing ABL baselines when there exists failure or attack on workers. 
\end{abstract}

\begin{keywords}
  distributed machine learning, momentum, asynchronous Byzantine learning, buffer, stochastic gradient descent
\end{keywords}

\section{Introduction}

Due to the wide application in cluster-based large-scale learning,
federated learning~\citep{kairouz2019advances_FLoverview,federated_learning_2016_google},
edge computing~\citep{edge_computing_2016} and so on,
distributed learning has recently become a hot research
topic~\citep{zinkevich2010parallelized,yang2013trading,
jaggi2014communication,shamir2014communication,zhang2014asynchronous_AD_ADMM,
ma2015adding,lee2017distributed,D_PSGD_2017_lian,zhao2017scope,
sun2018slim,wangni2018gradient,zhao2018proximal_pSCOPE,zhou2018distributed,
yu2019linear_PRSGDM,PR_SGD_yu2019,haddadpour2019trading,9217472_AsynOverview1,nokleby2020_AsynOverview2}.
Most traditional distributed learning methods are based on stochastic gradient descent~(SGD)
and its variants~\citep{bottou2010large,xiao2010dual,duchi2011adaptive,johnson2013accelerating,
shalev2013stochastic,zhang2013linear,lin2014accelerated,schmidt2017minimizing,zheng2017asynchronous_delaycompensation,zhao2018proximal_pSCOPE},
and typically assume no failure or attack.

However, in distributed learning applications with multiple networked machines~(nodes),
different kinds of hardware or software failure may happen. Representative failure includes bit-flipping
in the communication media and the memory of some workers~\citep{xie2019zeno}.
In this case, small failure on some machines~(workers) might cause a distributed learning method to fail.
In addition, malicious attack should not be neglected in an open network where the manager~(or server) generally has not much control on the workers, such as the cases of edge computing and federated learning. Malicious workers may behave arbitrarily or even adversarially. Hence, \emph{Byzantine learning}~(BL), which refers to distributed learning with failure or attack,
has attracted much attention~
\citep{diakonikolas2017being_highDrobust,chen2017distributed_geoMed,blanchard2017machine_krum,damaskinos2018asynchronous_Kardam,baruch2019little_ByztAttack,diakonikolas2019recent}.

Existing BL methods can be divided into two main categories: synchronous BL~(SBL) methods and asynchronous BL~(ABL) methods. 
In SBL methods, the learning information, such as the gradient in SGD, of all workers will be aggregated in a synchronous way. 
On the contrary, in ABL methods the learning information of workers will be aggregated in an asynchronous way. 
Existing SBL methods mainly take two different ways to achieve resilience against \emph{Byzantine workers} 
which refer to those workers with failure or attack.
One way is to replace the simple averaging aggregation operation with some more robust aggregation operations, such as median \& trimmed-mean~\citep{yin2018byzantine_median}, geometric median~\citep{chen2017distributed_geoMed}, and centered-clipping~\citep{karimireddy2020_learning_history}. Krum~\citep{blanchard2017machine_krum} and \mbox{ByzantinePGD}~\citep{yin2019defending_byztPGD} take this way.
The other way is to filter the suspicious learning information~(gradients) before averaging. 
Representative examples include \mbox{ByzantineSGD}~\citep{alistarh2018byzantineSGD} and Zeno~\citep{xie2019zeno}. 
Furthermore, some recent works reveal that using history information can strengthen the Byzantine resilience in SBL~\citep{allen2020byzantine,el2020distributed_byzMomentum,karimireddy2020_learning_history}.

The advantage of SBL methods is that they are relatively simple and easy to be implemented. 
But SBL methods will result in slow convergence when there exist heterogeneous workers. 
Furthermore, in some applications like federated learning and edge computing, synchronization cannot even be performed most of the time due to the offline workers~(clients or edge servers). 
Hence, ABL is preferred in these cases.

To the best of our knowledge, there exist only two ABL methods: Kardam~\citep{damaskinos2018asynchronous_Kardam} and Zeno++~\citep{xie2020zeno++}.
Kardam introduces two filters to drop out suspicious learning information~(gradients), which can still achieve good performance
when the communication delay is heavy. However, when in face of malicious attack, some work~\citep{xie2020zeno++} finds that Kardam also drops out most correct gradients in order to filter all faulty~(failure) gradients. Hence, Kardam cannot resist malicious attack. Zeno++ needs to store some training instances on server for scoring. In some practical applications like federated learning~\citep{kairouz2019advances_FLoverview}, storing data on server will increase the risk of privacy leakage or even face legal risk. Therefore, under the general setting where server has no access to any training instances, there does not exist any ABL method that can resist malicious attack.

Moreover, in some recently proposed attacks~\citep{xie2020_FoE,baruch2019little_ByztAttack}, attackers are assumed to have access to all the information on other workers and use these information for attack. This type of attacks are called omniscient attacks, while the others are called non-omniscient attacks. As far as we know, there does not exist any ABL method that can resist the two omniscient attacks `Fall of Empires'~\citep{xie2020_FoE} and `A Little is Enough'~\citep{baruch2019little_ByztAttack}.



In this paper, we propose a novel method called \underline{b}uffered \underline{a}synchronous \underline{s}tochastic \underline{g}radient \underline{d}escent~(\mbox{BASGD}) and an improved variant of BASGD called BASGD with \underline{m}omentum~(\mbox{BASGDm}) for ABL. 
The main contributions are listed as follows:

\begin{itemize}
\item   To the best of our knowledge, BASGD is the first ABL method that can resist non-omniscient attacks without storing any instances on server. With the benefit of local momentum, BASGDm can resist both non-omniscient and omniscient attacks. Compared with those methods which need to store instances on server, \mbox{BASGD} and \mbox{BASGDm} have a wider scope of application.  

\item Both \mbox{BASGD} and \mbox{BASGDm} are compatible with various aggregation rules. Moreover, both \mbox{BASGD} and \mbox{BASGDm} are proved to be convergent and be able to resist failure or attack.

\item Empirical results show that our methods significantly outperform existing ABL baselines when there exists failure or attack on workers.
\end{itemize}

\section{Preliminary}

In this section, we present the preliminary of this paper, including the distributed learning framework used in this paper and the definition of Byzantine worker.


\subsection{Distributed Learning Framework}\label{subsec:framework}


Many machine learning models, such as logistic regression and deep neural networks, can be formulated as the following finite sum optimization problem:
\begin{equation} \label{eq:obj}
\min_{\w \in \RB^d}{F(\w)}=\frac{1}{n}\sum_{i=1}^n f(\w;z_i),
\end{equation}
where $\w$ is the parameter to learn,
$d$ is the dimension of parameter, $n$ is the number of training instances,
$f(\w;z_i)$ is the empirical loss on the instance $z_i$.
The goal of distributed learning is to solve the problem in~(\ref{eq:obj}) by designing learning algorithms based on multiple networked machines.

Although there have appeared many distributed learning frameworks, in this paper we focus on the widely used Parameter Server~(PS) framework~\citep{C_PSGD}. 
In a PS framework, there are several workers and one or more servers. Each worker can only communicate with server(s). There may exist more than one server in a PS framework, but for the problem of this paper servers can be logically conceived as a unity. 
Without loss of generality, we will assume there is only one server in this paper. Training instances are disjointedly distributed across $m$ workers.
Let $\DM_k$ denote the index set of training instances on worker\_$k$, we have $\cup_{k=1}^m\DM_k=\{1,2,\ldots,n\}$ and $\DM_k\cap\DM_{k'}=\emptyset$ if $k\neq k'$. In this paper, we assume that server has no access to any training instances.
If two instances have the same value,
they are still deemed as two distinct instances.
Namely, $z_i$ may equal $z_{i'}$~$(i\neq i')$.
One popular asynchronous method to solve the problem in~(\ref{eq:obj})
under the PS framework is ASGD~\citep{dean2012large_ASGD}~(see Appendix~\ref{appendix:alg_details} for details).
In this paper, we assume each worker samples one instance for gradient computation each time. The analysis of mini-batch case is similar.





In PS based ASGD, server is responsible for updating and maintaining the latest parameter.
The number of iterations that server has already executed is used
as the global logical clock of server. 
At the beginning, iteration number $t=0$. 
Each time a SGD step is executed, $t$ will increase by $1$ immediately.
The parameter after $t$ iterations is denoted as $\w^t$.
If server sends parameters to worker\_$k$ at iteration $t'$,
some SGD steps may have been excuted before server receives gradient from worker\_$k$ next time at iteration $t$.
Thus, we define the \emph{delay} of worker\_$k$ at iteration $t$ as
$\tau_k^t=t-t'$. Worker\_$k$ is \emph{heavily delayed} at iteration $t$
if $\tau_k^t>\tau_{max}$, where $\tau_{max}$ is a pre-defined non-negative constant.

\subsection{Byzantine Worker}\label{subsec:byzantine}

For workers that have sent gradients~(one or more) to server at iteration $t$,
we call worker\_$k$ \textit{loyal worker} if it has finished all the tasks without any fault and each sent gradient is correctly received by the server.
Otherwise, worker\_$k$ is called \textit{Byzantine worker}.
If worker\_$k$ is a Byzantine worker, it means the received gradient from worker\_$k$ is not credible, which can be an arbitrary value.
In ASGD, there is one received gradient at a time. 
Formally, we denote the gradient received from worker\_$k$ at iteration $t$ as $\g_{k}^t$. Then, we have:
\begin{equation*}
    \g_{k}^t=\left\{
    \begin{aligned}
        &\nabla f(\w^{t'};z_i),~~\text{if worker\_$k$ is loyal at iteration $t$};\\
        &~*,~~~~~~~~~~\text{if worker\_$k$ is Byzantine at iteration $t$},
    \end{aligned}
    \right .
\end{equation*}
where $0\leq t'\leq t$, and $i$ is randomly sampled from $\DM_k$. `$*$' represents an arbitrary value.
Our definition of Byzantine worker is consistent with
most previous works~\citep{blanchard2017machine_krum,xie2019zeno,xie2020zeno++}.
Either accidental failure or malicious attack will result in Byzantine workers. 


\begin{algorithm}[tb]
\caption{Buffered Asynchronous SGD~(BASGD)}
\label{alg:BASGD}
\begin{algorithmic}
\vskip 0.05in
\STATE {\bfseries  Server:}
\STATE {\bfseries Input:} learning rate $\eta$, reassignment interval $\Delta$,\\
~~~buffer number $B$, aggregation function: $Aggr(\cdot)$;
\STATE {\bfseries Initialization:} initial parameter $\w^0$, learning rate $\eta$;
\STATE Set buffer:~$\h_b\leftarrow \0$, $N_b^t\leftarrow 0$;
\STATE Initialize mapping table $\beta_s\leftarrow s$ $(s=0,1,\ldots,m-1)$;
\STATE Send initial $\w^0$ to all workers;
\STATE Set $t\leftarrow 0$, and start the timer;
\REPEAT
    \STATE Wait until receiving $\g$ from some worker\_$s$;
    \STATE Choose buffer: $b\leftarrow \beta_s~mod~B$;
    \STATE Let $N_b^t\leftarrow N_b^t + 1$, and $\h_b\leftarrow\frac{(N_b^t-1)\h_b+\g}{N_b^t}$;
    \IF{$N_b^t > 0$ for each $b\in[B]$}
        \STATE Aggregate: $\G^t = Aggr([\h_1,\ldots,\h_B])$;
        \STATE Execute SGD step: $\w^{t+1}\leftarrow\w^t-\eta\cdot\G^t$;
        \STATE Zero out buffers:~$\h_b\leftarrow \0$, $N_b^t\leftarrow 0$ $(b=1,\ldots,B)$;
        \STATE Set $t\leftarrow t+1$, and restart the timer;
    \ENDIF
    \IF{the timer has exceeded $\Delta$ seconds}
        \STATE Zero out buffers:~$\h_b\leftarrow \0$, $N_b^t\leftarrow 0$ $(b=1,\ldots,B)$;
        \STATE Modify the mapping table $\{\beta_s\}_{s=0}^{m-1}$ for buffer reassignment, and restart the timer;
    \ENDIF
    \STATE 
    Send back the latest parameters back to worker\_$s$, no matter whether a SGD step is executed or not.
\UNTIL{stop criterion is satisfied}
\STATE Notify all workers to stop;

\vskip 0.1in

\STATE{\bfseries  Worker\_$k$:} ~~$(k=0,1,...,m-1)$
\REPEAT
    \STATE{Wait until receiving the latest parameter $\w$ from server};
    \STATE{Randomly sample an index $i$ from $\DM_k$;}
    \STATE{Compute $\nabla f(\w;z_i)$;}
    \STATE{Send $\nabla f(\w;z_i)$ to server};
\UNTIL{receive server's notification to stop}
\end{algorithmic}
\end{algorithm}

\section{Buffered Asynchronous SGD}\label{sec:BASGD}

In synchronous BL, gradients from all workers are received at each iteration.
    We can compare the gradients with each other, and then filter suspicious ones,
    or use more robust aggregation rules such as median and trimmed-mean for updating. However, in asynchronous BL,
    only one gradient is received at a time. Without any training instances stored on server, it is difficult for server
    to identify whether a received gradient is credible or not.
    
    In order to deal with this problem in asynchronous BL, we propose a novel method called buffered asynchronous SGD~(BASGD). 
    BASGD introduces $B$ buffers~($0<B\leq m$) on server, and the gradient used for updating parameters will be aggregated from these buffers. 
    The detail of the learning procedure of BASGD is presented in Algorithm~\ref{alg:BASGD}.  
    In this section, we will first introduce the three key components of BASGD: 
    buffer, aggregation function, and mapping table.
    At the end of this section, we will also introduce an improved variant of BASGD which is called buffered asynchronous SGD with momentum~(BASGDm).

    \begin{figure}[t]
      \begin{center}
      \centerline{\includegraphics[width=0.65\linewidth]{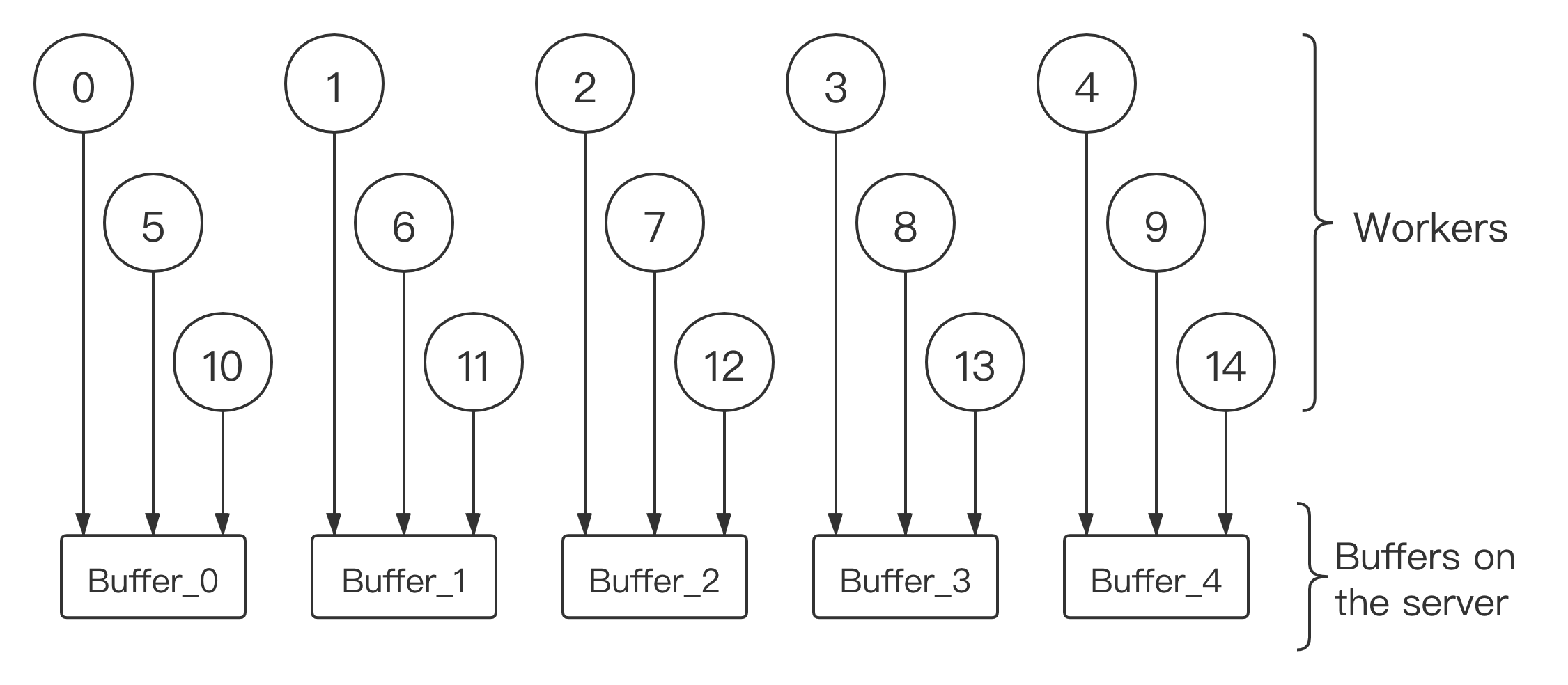}}
      \caption{An example of buffers. Circle represents worker,
      and the number is worker ID. There are $15$ workers and $5$ buffers.
      The gradient received from worker\_$s$ is stored in buffer\_\{$s$ mod $5$\}.
      }
      \label{fig:buffer}
      \end{center}
   \end{figure}

\subsection{Buffer}
    In BASGD, the $m$ workers do the same job as that in ASGD, while the updating rule on server is modified. More specifically, there are $B$ buffers~($0<B\leq m$) on server. 
    When a gradient $\g$ from worker\_$s$ is received,
    it will be temporarily stored in buffer $b$, where $b=s~mod~B$, as illustrated in Figure~\ref{fig:buffer}.
    Only when each buffer has stored at least one gradient, a new SGD step will be executed.
    Please note that no matter whether a SGD step is executed or not, 
    the server will immediately send the latest parameters back to the worker after receiving a gradient.
    Hence, BASGD introduces no barrier, and is an asynchronous algorithm. 

    For each buffer $b$, more than one gradient may have been received at iteration $t$. We will store the average of these gradients (denoted by $\h_b$) in buffer $b$. Assume that there are already $(N-1)$ gradients $\g_1,\g_2,\ldots,\g_{N-1}$ which should be stored in buffer $b$,
    and $\h_{b(old)}=\frac{1}{N-1}\sum_{i=1}^{N-1}\g_i.$
    When the $N$-th gradient $\g_{N}$ is received, the new average value is:
    $$\h_{b(new)}=\frac{1}{N}\sum_{i=1}^{N}\g_i
    =\frac{N-1}{N}\cdot\h_{b(old)}+\frac{1}{N}\cdot\g_{N}.$$
    This is the updating rule for each buffer $b$ when a gradient is received. We use $N_b^t$ to denote the total number of gradients stored in buffer $b$ at the $t$-th iteration. After the parameter $\w$ is updated, all buffers will be zeroed out at once. With the benefit of buffers, server has access to $B$ candidate gradients when updating parameter. Thus, a more reliable~(robust) gradient can be aggregated from the $B$ gradients of buffers, if a proper aggregation function $Aggr(\cdot)$ is chosen.


Please note that from the perspective of workers, BASGD is fully asynchronous, since a worker will immediately receive the latest parameter from the server after sending a gradient to the server, without waiting for other workers.
Meanwhile, from the perspective of server, BASGD is semi-asynchronous because the server will not update the model until all buffers are filled. However, it is a necessity to limit the updating frequency in ABL when server has no instances. If the server always updates the model when receiving a gradient, it will be easily foiled when Byzantine workers send gradients much more frequently than others. A similar conclusion has been proved in previous works~\citep{damaskinos2018asynchronous_Kardam}.

\subsection{Aggregation Function}
    When a SGD step is ready to be executed, there are $B$ buffers providing candidate gradients. An aggregation function is needed to get the final gradient for updating. A naive way is to take the mean of all candidate gradients. However, mean value is sensitive to outliers which are common in BL.
    For designing proper aggregation functions, we first define the $q$-Byzantine Robust~($q$-BR) condition to quantitatively describe the Byzantine resilience ability of an aggregation function.
    \begin{definition}[$q$-Byzantine Robust]\label{def:qbr}
        For an aggregation function $Aggr(\cdot)$:
        $Aggr([\h_1,\ldots,$ $\h_B])=\G$,
        where $\G=[G_1,\ldots,G_d]^T$ and $\h_b=[h_{b1},\ldots,h_{bd}]^T,\forall b \in [B]$,
        we call $Aggr(\cdot)$ \emph{$q$-Byzantine Robust}~($q\in \ZB,0<q<B/2$),
        if it satisfies the following two properties:
        \vskip 0.05in
        \noindent
        (a) $Aggr([\h_1+\h',\ldots,\h_B+\h'])=Aggr([\h_1,\ldots,\h_B])+\h',\ $
        $\forall \h_1,\ldots,\h_B\in\RB^d, \forall \h'\in\RB^d;$\\
        (b) $\min_{s\in\SM} \{h_{sj}\}\leq G_j\leq \max_{s\in\SM} \{h_{sj}\},\ $
        $\forall j \in [d]$, $\forall \SM \subset [B]$ with $|\SM|=B-q.$
    \end{definition}

    Intuitively, property (a) in Definition~\ref{def:qbr} says that if all candidate vectors $\h_i$ are added by a same vector $\h'$,
    the aggregated gradient will also be added by $\h'$.
    Property (b) says that for each coordinate $j$, the aggregated value $G_j$
    will be between the $(q+1)$-th smallest value and the $(q+1)$-th largest value
    among the $j$-th coordinates of all candidate vectors.
    Thus, the gradient aggregated by a $q$-BR function is insensitive to at least $q$ outliers.
    We can find that $q$-BR condition gets stronger when $q$ increases.
    Namely, if $Aggr(\cdot)$ is $q$-BR, then for any $0<q'<q$, $Aggr(\cdot)$ is also $q'$-BR.

    \begin{remark}
        When $B>1$, mean function is not $q$-Byzantine Robust for any $q>0$.
        We illustrate this by a one-dimension example:
        $h_1,\ldots,h_{B-1}\in[0,1]$, and $h_B=10\times B$.
        Then $\frac{1}{B}\sum_{b=1}^B h_b \geq \frac{h_B}{B} = 10 \not\in [0,1].$
        Namely, the mean is larger than any of the first $B-1$ values.
    \end{remark}
    The following two aggregation functions are both $q$-BR.

    \begin{definition}[Coordinate-wise median~\citep{yin2018byzantine_median}]\label{def:cwMedian}
        For candidate vectors $\h_1,\h_2,\ldots$, $\h_B \in \RB^d$,
        $\h_b=[h_{b1},h_{b2},\ldots,h_{bd}]^T$, $\forall b=1,2,\ldots,B$.
        \emph{Coordinate-wise median} is defined as:
        $$Med([\h_1,\ldots,\h_B])= [Med(h_{\cdot 1}),\ldots,Med(h_{\cdot d})]^T,$$
        where $Med(h_{\cdot j})$ is the scalar median of the $j$-th coordinates, 
        $\forall j=1,2,\ldots,d$.
    \end{definition}

    \begin{definition}[Coordinate-wise $q$-trimmed-mean~\citep{yin2018byzantine_median}]\label{def:cwTrmean}
        For any positive interger $q<B/2$ and
        candidate vectors $\h_1,\h_2,\ldots,\h_B \in \RB^d$,
        $\h_b=[h_{b1},h_{b2},\ldots,h_{bd}]^T$, $\forall b=1,2,\ldots,B$.
        \emph{Coordinate-wise $q$-trimmed-mean} is defined as:
        $$Trm([\h_1,\ldots,\h_B])= [Trm(h_{\cdot 1}),\ldots,Trm(h_{\cdot d})]^T,$$
        where $Trm(h_{\cdot j})=\frac{1}{B-2q}\sum_{b\in\MM_j}h_{bj}$ is the scalar $q$-trimmed-mean. $\MM_j$ is the subset of $\{h_{bj}\}_{b=1}^B$ obtained by removing
        the $q$ largest elements and $q$ smallest elements.
    \end{definition}

    In the following content, coordinate-wise median and coordinate-wise $q$-trimmed-mean
    are also called \emph{median} and \emph{trmean}, respectively. Proposition~\ref{prop:qbr} shows the $q$-BR property of these two functions.

    \begin{proposition}\label{prop:qbr}
        Coordinate-wise $q$-trmean is $q$-BR.
        Coordinate-wise median is $\lfloor\frac{B-1}{2}\rfloor$-BR.
            \end{proposition}

    Here, $\lfloor x \rfloor$ represents the maximum integer that is not larger than $x$. 
    According to Proposition~\ref{prop:qbr}, both median and trmean are proper choices for aggregation function in BASGD.
    The proof can be found in Appendix~\ref{appendix:proof_details}. 
    Now we define another class of aggregation functions, which is also important for the analysis in Section~\ref{sec:proof}.

    \begin{definition}[Stable aggregation function]\label{def:stable aggr}
       Aggregation function $Aggr(\cdot)$ is called \emph{stable} provided that
       $\forall \h_1,\ldots,\h_B$, $\tilde{\h}_1,\ldots,\tilde{\h}_B\in\RB^d$, 
       letting $\delta=(\sum_{b=1}^B\|\h_b-\tilde{\h}_b\|^2)^{\frac{1}{2}}$, we have:
       $$\|Aggr(\h_1,\ldots,\h_B)
           -Aggr(\tilde{\h}_1,\ldots,\tilde{\h}_B)\|
           \leq\delta.$$
       
    \end{definition}
    If $Aggr(\cdot)$ is a stable aggregation function, 
    it means that when there is a disturbance with $L_2$-norm $\delta$ on buffers,
    the disturbance of aggregated result will not be larger than $\delta$.
    
    \begin{definition}[Effective aggregation function]\label{def:effc_aggr}
      When there are at most $r$ Byzantine workers, stable aggregation function $Aggr(\cdot)$ is called an $(A_1,A_2)$-effective aggregation function, 
       provided that it satisfies the following two properties for all $\w^t \in \RB^d$ in cases without delay~($\tau_k^t =0,\ \forall t=0,1,\ldots,T-1$):
       \vskip 0.05in
       \noindent
       (a) $\EB[\nabla F(\w^t)^T\G_{syn}^t~|~\w^t] \geq \|\nabla F(\w^t)\|^2 - A_1$;\\
       (b) $\EB[\|\G_{syn}^t\|^2~|~\w^t]\leq (A_2)^2$;
       \vskip 0.02in

     where $A_1,A_2\in\RB_+$ are two non-negative constants,  $\G_{syn}^t$ is the aggregated result of $Aggr(\cdot)$ at the $t$-th iteration in cases without delay. 
    \end{definition}

    More specifically, $\G_{syn}^t$ can be the aggregated \emph{gradient} or \emph{momentum}. In the conference version~\citep{yang2020_basgd}, $\G_{syn}^t$ is the aggregated \emph{gradient}. We change the statement to make it compatible with BASGDm method, which we will introduce in Section~\ref{subsec:BASGDm}.
    
    For different aggregation functions, constants $A_1$ and $A_2$ may differ. $A_1$ and $A_2$ are related to loss function $F(\cdot)$, distribution of instances, buffer number $B$, maximum Byzantine worker number $r$. 
    Inequalities (a) and (b) in Definition~\ref{def:effc_aggr} are two important properties in convergence proof of synchronous Byzantine learning methods. 
    As revealed in \citep{yang2020adversary_BLoverview}, there are many existing aggregation rules for Byzantine learning. We find that most of them satisfy Definition~\ref{def:effc_aggr}.
    For example, Krum, median, and trimmed-mean have already been proved to satisfy these two properties~\citep{blanchard2017machine_krum,yin2018byzantine_median}. 
    SignSGD~\citep{bernstein2018signsgd} can be seen as a combination of 1-bit quantization and median aggregation, while median satisfies the properties in Definition~\ref{def:effc_aggr}. 

    Compared to Definition~\ref{def:qbr}, Definition~\ref{def:effc_aggr} can be used to obtain a tighter bound with respect to $A_1$ and $A_2$. However, it usually requires more effort to check the two properties in Definition~\ref{def:effc_aggr} than those in Definition~\ref{def:qbr}. 

    Please note that too large $B$ could lower the updating frequency and damage the performance, while too small $B$ may harm the Byzantine resilience. Thus, a moderate $B$ is usually preferred. In some practical applications, we could estimate the maximum number of Byzantine workers $r$, and set $B$ to make the aggregation function resilient to up to $r$ Byzantine workers.
    In particular, $B$ is suggested to be $(2r+1)$ for median, since median is $\lfloor\frac{B-1}{2}\rfloor$-BR.

    \begin{figure}[t]
      \begin{center}
      \centerline{\includegraphics[width=0.65\linewidth]{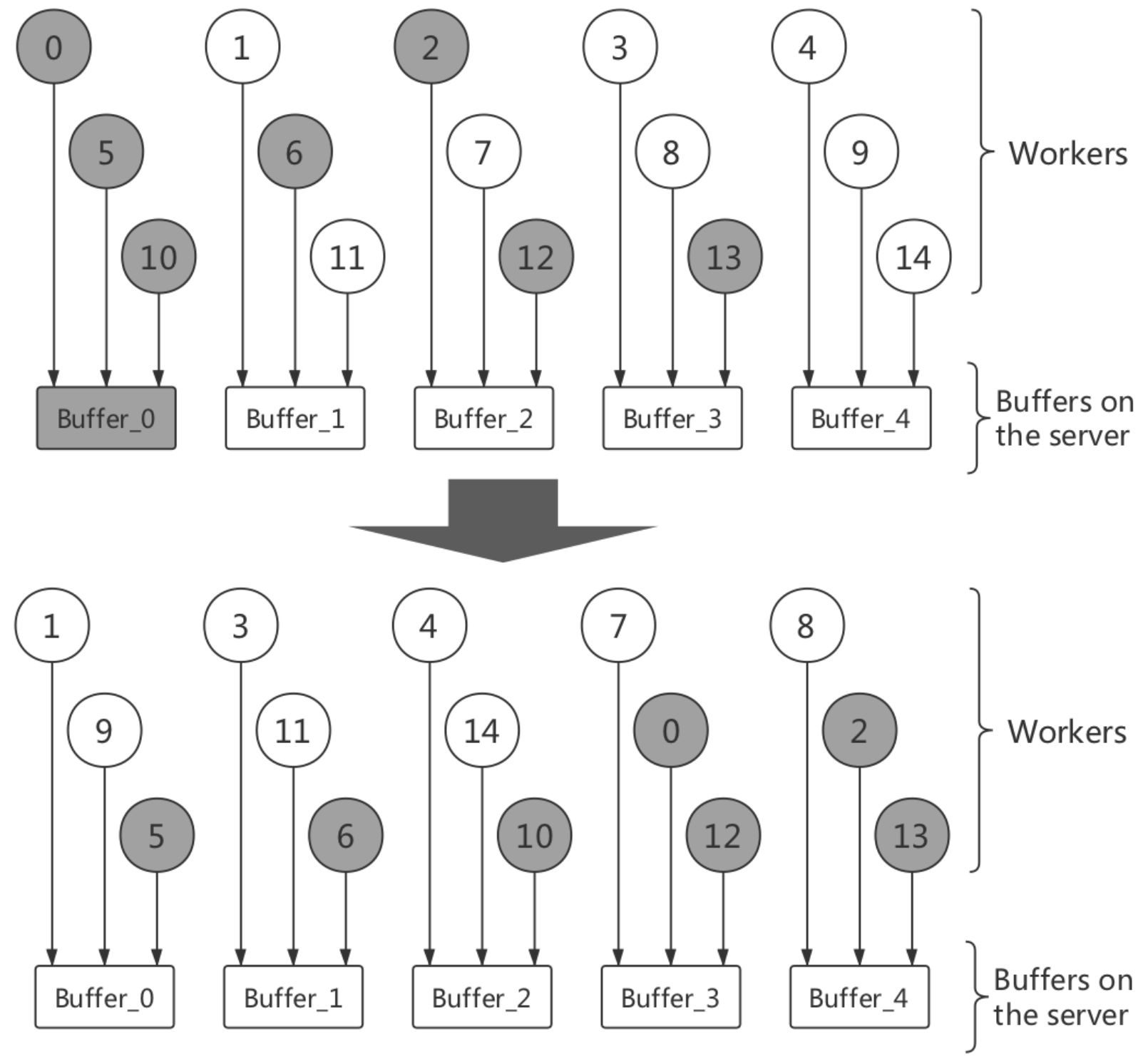}}
      \caption{An example of buffer reassignment. White circle represents active worker, and grey circle represents unresponsive worker. Before reassignment, buffer\_$0$ is a straggler. After reassignment, there is at least one active worker corresponding to each buffer.
      }
      \label{fig:reassignment}
      \end{center}
   \end{figure}
\subsection{Mapping Table}
    At each iteration of BASGD, buffer\_$b$ needs at least one gradient for aggregation. In the worst case, all the workers corresponding to buffer\_$b$ may be unresponsive. In this case, buffer\_$b$ will become the straggler, and slow down the whole learning process. To deal with this problem, we introduce the mapping table for buffer reassignment technique.

    We call a worker active worker if it has responsed at the current iteration.
    If SGD step has not been excuted for $\Delta$ seconds, the server immediately zeroes out stored gradients in all buffers, equally reassigns active workers to each buffer, and then continues the learning procedure. Hyper-parameter $\Delta$ is called reassignment interval.
    Figure~\ref{fig:reassignment} illustrates an example of reassignment. The grey circles represents unresponsive workers. After reassignment, there are at least one active worker corresponding to each buffer.

    Specifically, we introduce a mapping table $\{\beta_s\}_{s=0}^{m-1}$ for buffer reassignment. Initially, $\beta_s=s$ $(\forall s=0,1,\ldots,m-1)$. When reassigning buffers, the server only needs to modify the mapping table $\{\beta_s\}_{s=0}^{m-1}$, and then stores worker\_$s$'s gradients in buffer\_\{$\beta_s$ mod $B$\}, instead of buffer\_\{$s$ mod $B$\} any more. Please note that the server only needs to modify the mapping table for buffer reassignment, and there is no need to notify workers.

    Besides, a timer is used on the server for indicating when to reassign buffers.
    The timer is started at the beginning of BASGD, and is restarted immediately after each SGD step or buffer reassignment. When the timer exceeds $\Delta$ seconds, buffers will be zeroed out, and reassignment executed. Hyper-parameter $\Delta$ should be set properly. If $\Delta$ is too small, buffers will be zeroed out too frequently, which may slow down the learning process. If $\Delta$ is too large, straggler buffers could not be eliminated in time.

    \subsection{Buffered Asynchronous SGD with Momentum}\label{subsec:BASGDm}

    As previous works have revealed, history information can greatly help to resist Byzantine attacks~\citep{el2020distributed_byzMomentum, allen2020byzantine,karimireddy2020_learning_history}. Therefore, we introduce momentum into BASGD, and obtain the method buffered asynchronous SGD with momentum~(BASGDm). In BASGDm, the algorithm of server is exactly the same as that in BASGD. The only difference is that each worker maintains a local momentum, and sends local momentums to server instead of gradients. The detail of BASGDm is illustrated in Algorithm~\ref{alg:BASGDm}. With the benefit of momentum, BASGDm can achieve stronger Byzantine resilience. In particular, BASGDm can resist both non-omniscient and omniscient attacks, as we will show in Section~\ref{sec:exp}.

\begin{algorithm}[ht]
  \caption{Buffered Asynchronous SGD with Momentum~(BASGDm)}
  \label{alg:BASGDm}
  \begin{algorithmic}
  \vskip 0.05in
  \STATE {\bfseries  Server:}
  \STATE {\bfseries Input:} learning rate $\eta$, momentum hyper-parameter $\mu\ (0\leq \mu<1)$,\\
  ~~~reassignment interval $\Delta$, buffer number $B$, aggregation function: $Aggr(\cdot)$;
  \STATE {\bfseries Initialization:} initial parameter $\w^0$, learning rate $\eta$;
  \STATE Set buffer:~$\h_b\leftarrow \0$, $N_b^t\leftarrow 0$;
  \STATE Initialize mapping table $\beta_s\leftarrow s$ $(s=0,1,\ldots,m-1)$;
  \STATE Send initial $\w^0$ to all workers;
  \STATE Set $t\leftarrow 0$, and start the timer;
  \REPEAT
      \STATE Wait until receiving $\u$ from some worker\_$s$;
      \STATE Choose buffer: $b\leftarrow \beta_s~mod~B$;
      \STATE Let $N_b^t\leftarrow N_b^t + 1$, and $\h_b\leftarrow\frac{(N_b^t-1)\h_b+\u}{N_b^t}$;
      \IF{$N_b^t > 0$ for each $b\in[B]$}
          \STATE Aggregate: $\G^t = Aggr([\h_1,\ldots,\h_B])$;
          \STATE Execute SGD step: $\w^{t+1}\leftarrow\w^t-\eta\cdot\G^t$;
          \STATE Zero out buffers:~$\h_b\leftarrow \0$, $N_b^t\leftarrow 0$ $(b=1,\ldots,B)$;
          \STATE Set $t\leftarrow t+1$, and restart the timer;
      \ENDIF
      \IF{the timer has exceeded $\Delta$ seconds}
          \STATE Zero out buffers:~$\h_b\leftarrow \0$, $N_b^t\leftarrow 0$ $(b=1,\ldots,B)$;
          \STATE Modify the mapping table $\{\beta_s\}_{s=0}^{m-1}$ for buffer reassignment, and restart the timer;
      \ENDIF
      \STATE 
      Send back the latest parameters back to worker\_$s$, no matter whether a SGD step is executed or not.
  \UNTIL{stop criterion is satisfied}
  \STATE Notify all workers to stop;
  
  \vskip 0.1in
  
  \STATE{\bfseries  Worker\_$k$:} ~~$(k=0,1,...,m-1)$
  \STATE {\bfseries Initialization:} initial momentum $\u\leftarrow\0$;
  \REPEAT
      \STATE{Wait until receiving the latest parameter $\w$ from server};
      \STATE{Randomly sample an index $i$ from $\DM_k$;}
      \STATE{Compute stochastic gradient $\nabla f(\w;z_i)$;}
      \STATE{Update local momentum $\u\leftarrow \mu\cdot\u+(1-\mu)\cdot \nabla f(\w;z_i)$;}
      \STATE{Send $\u$ to server};
  \UNTIL{receive server's notification to stop}
  \end{algorithmic}
  \end{algorithm}

\section{Convergence}\label{sec:proof}
    In this section, we theoretically prove the convergence and
    resilience of BASGD and BASGDm against failure or attack. We will introduce three main theorems in this section. 
    The first two theorems are for BASGD. One presents a relatively loose but general bound for all $q$-BR aggregation functions, while the other one presents a relatively tight bound for each distinct $(A_1,A_2)$-effective aggregation function. Since the definition of $(A_1,A_2)$-effective aggregation function is usually more difficult to verify than $q$-BR property, the general bound is also useful.

    Similar to the second theorem, the last one is for BASGDm with $(A_1,A_2)$-effective aggregation function.
    Here we only present the results. Proof details are in Appendix~\ref{appendix:proof_details}.
    We first make the following assumptions, 
    which have been widely used in stochastic optimization.

    \begin{assumption}[Lower bound]\label{ass:l_bound}
        Global loss function $F(\w)$ is bounded below:
        $\exists F^*\in \RB, F(\w)\geq F^*, \forall \w\in\RB^d$.
    \end{assumption}

    \begin{assumption}[Bounded bias]\label{ass:limit_bias}
        For any loyal worker, it can use locally stored training instances
        to estimate global gradient with bounded bias $\kappa$:
        $\|\EB[\nabla f(\w;z_i)] - \nabla F(\w)\|\leq \kappa ,~\forall \w \in \RB^d.$
    \end{assumption}

    \begin{assumption}[Bounded gradient]\label{ass:bounded_gradient}
        $\nabla F(\w)$ is bounded: $\|\nabla F(\w)\|\leq D,\ \forall \w \in \RB^d$.
    \end{assumption}

    \begin{assumption}[Bounded variance]\label{ass:lim_variance}
        $\EB[||\nabla f(\w;z_i)-\EB[\nabla f(\w;z_i)|\w]||^2~|~\w]\leq \sigma^2,\ \forall \w \in \RB^d$.
    \end{assumption}

    \begin{assumption}[$L$-smoothness]\label{ass:L_smooth}
        Global loss function $F(\w)$ is differentiable and $L$-smooth:
        $||\nabla F(\w)-\nabla F(\w')||\leq L ||\w-\w'||,~\forall \w,\w' \in\RB^d.$
    \end{assumption}




    Let $N^{(t)}$ be the $(q+1)$-th smallest value in $\{N_b^t\}_{b\in[B]}$, where $N_b^t$ is the number of gradients stored in buffer $b$ at the $t$-th iteration. We define the constant $$\Lambda_{B,q,r}=\frac{(B-r)\sqrt{B-r+1}}{\sqrt{(B-q-1)(q-r+1)}},$$ which will appear in Lemma~\ref{lemma:variance_bound} and Lemma~\ref{lemma:expectation_bound}.

    \begin{lemma}\label{lemma:variance_bound}
        If $Aggr(\cdot)$ is $q$-BR, 
        and there are at most $r$ Byzantine workers~$(r\leq q)$, we have:
        $$\EB[||\G^t||^2~|~\w^t]\leq \Lambda_{B,q,r}d\cdot(D^2+\sigma^2 / N^{(t)}).$$
    \end{lemma}

    \begin{lemma}\label{lemma:expectation_bound}
        If $Aggr(\cdot)$ is $q$-BR, and the total number of 
        heavily delayed workers and Byzantine workers is not larger than $r~(r\leq q)$, we have:
        \begin{align*}
          ||\EB[\G^t-\nabla F(\w^t)~|~\w^t]||
          \leq \Lambda_{B,q,r}d (\tau_{max}L\cdot[\Lambda_{B,q,r}d(D^2+\sigma^2/N^{(t)})]^\frac{1}{2} + \sigma +\kappa) .
        \end{align*}
    \end{lemma}

    \begin{theorem}\label{thm:main_theorem}
        Let $\tilde{D}=\frac{1}{T}\sum_{t=0}^{T-1}(D^2 +\sigma^2/N^{(t)})^\frac{1}{2}$.
        If $Aggr(\cdot)$ is $q$-BR, $B=O(r)$, and the total number of 
        heavily delayed workers and Byzantine workers is not larger than $r~(r\leq q)$, with learning rate $\eta=O(\frac{1}{L\sqrt{T}})$, we have:
        \begin{align*}
          &\frac{\sum_{t=0}^{T-1} \EB[||\nabla F(\w^t)||^2]}{T}
          \leq O\left(\frac{L[F(\w^0)-F^*]}{T^\frac{1}{2}}
          \right)
          +O\left(\frac{rd\tilde{D}}{T^\frac{1}{2}(q-r+1)^\frac{1}{2}}\right) \\ & \qquad \qquad \qquad
          +O\left(\frac{rD d\sigma}{(q-r+1)^\frac{1}{2}}\right)
          +O\left(\frac{rD d\kappa}{(q-r+1)^\frac{1}{2}}\right)
          +O\left(\frac{r^\frac{3}{2}LD\tilde{D}d^\frac{3}{2}\tau_{max}}{(q-r+1)^\frac{3}{4}}\right).
        \end{align*}
    \end{theorem}
    Please note that the convergence rate of vanilla ASGD is $O(1/T^{\frac{1}{2}})$.
    Hence, Theorem~\ref{thm:main_theorem} indicates that BASGD has a theoretical convergence rate as fast as vanilla ASGD, with an extra constant variance.
    The term $O(rD d\sigma/(q-r+1)^\frac{1}{2})$ is caused by the aggregation function, which can be deemed as a sacrifice for Byzantine resilience.
    The term $O(rDd\kappa/(q-r+1)^{\frac{1}{2}})$ is caused by the differences of training instances among different workers. In independent and identically distributed~(i.i.d.) cases, $\kappa=0$ and the term vanishes.
    The term $O( r^{\frac{3}{2}}LD\tilde{D}d^{\frac{3}{2}}\tau_{max}/(q-r+1)^{\frac{3}{4}})$ is caused by the delay, and related to parameter $\tau_{max}$. The term is also related to the buffer size. When $N_b^t$ increases, $N^{(t)}$ may increase, and thus $\tilde{D}$ will decrease. Namely, larger buffer size will result in smaller $\tilde{D}$. 
    Besides, the factor $(q-r+1)^{-\frac{1}{2}}$ or $(q-r+1)^{-\frac{3}{4}}$ decreases as $q$ increases, and increases as $r$ increases.


    Although general, the bound presented in Theorem~\ref{thm:main_theorem} is relatively loose in high-dimensional cases, since $d$ appears in all the three extra terms. To obtain a tighter bound, we introduce Theorem~\ref{thm:general_proper} for BASGD with $(A_1,A_2)$-effective aggregation function~(Definition~\ref{def:effc_aggr}).

\begin{theorem}\label{thm:general_proper}
 In BASGD, if the total number of heavily delayed workers and Byzantine workers is not larger than $r$, $B=O(r)$, and $Aggr(\cdot)$ is an $(A_1,A_2)$-effective aggregation function. With learning rate $\eta=O(\frac{1}{\sqrt{LT}})$ satisfying that $2\eta^2L^2\tau_{max}^2(B-r)<1$, in general asynchronous cases we have:
\begin{align*}
  \frac{\sum_{t=0}^{T-1}\EB[\|\nabla F(\w^t)\|^2]}{T}
  \leq O\left(\frac{L^\frac{1}{2}[F(\w^0)-F^*]}{T^\frac{1}{2}}\right)
  +O&\left(\frac{L^\frac{1}{2}\tau_{max}DA_2r^\frac{1}{2}}{T^\frac{1}{2}}\right) 
  +O\left(\frac{L^\frac{1}{2}(A_2)^2}{T^\frac{1}{2}}\right)\\
  &+O\left(\frac{L^\frac{5}{2}(A_2)^2\tau_{max}^2 r}{T^\frac{3}{2}}\right)
  +A_1.
\end{align*}
\end{theorem}
Theorem~\ref{thm:general_proper} indicates that 
if $Aggr(\cdot)$ makes a synchronous BL method converge~(i.e., satisfies Definition~\ref{def:effc_aggr}), BASGD converges when using $Aggr(\cdot)$ as aggregation function.
Hence, BASGD can also be seen as a technique of asynchronization.
That is to say, new asynchronous methods can be obtained from synchronous ones when using BASGD. The extra constant term $A_1$ is caused by gradient bias. When there is no Byzantine workers~($r=0$), and instances are i.i.d. across workers, letting $B=1$ and $Aggr(\h_1,\ldots,\h_B)=Aggr(\h_1)=\h_1$, BASGD degenerates to vanilla ASGD. In this case, there is no gradient bias ($A_1=0$), and BASGD has a convergence rate of $O(1/T^\frac{1}{2})$, which is the same as that of vanilla ASGD~\citep{liu2021distributed_DLwithFOM}. 

Meanwhile, it remains uncertain whether the dependence to the staleness parameter $\tau_{max}$ is tight enough. Theorem~\ref{thm:general_proper} illustrates that BASGD has a convergence rate of $O(\tau_{max}/T^{\frac{1}{2}})$, while the convergence rate of vanilla ASGD can reach $O(\tau_{max}/T)$. To the best of our knowledge, there exist few works revealing the tightness of $\tau_{max}$ in asynchronous BL, and we will leave this for future work.

Similarly, we have the following theoretical results for BASGDm.

\begin{theorem}\label{thm:general_proper_m}
  In BASGDm, if the total number of heavily delayed workers and Byzantine workers is not larger than $r$, $B=O(r)$, and $Aggr(\cdot)$ is an $(A_1,A_2)$-effective aggregation function. With learning rate $\eta=O(\frac{1}{\sqrt{LT}})$ satisfying that $2\eta^2L^2\tau_{max}^2(1-\mu)^2(B-r)<1$, in general asynchronous cases we have:
  \begin{align*}
    \frac{\sum_{t=0}^{T-1}\EB[\|\nabla F(\w^t)\|^2]}{T}
    \leq & O\left(\frac{L^\frac{1}{2}[F(\w^0)-F^*]}{T^\frac{1}{2}}\right)+O\left(\frac{L^\frac{1}{2}\tau_{max}DA_2r^\frac{1}{2}(1-\mu)}{T^\frac{1}{2}}\right)\\
    &+O\left(\frac{L^\frac{1}{2}(A_2)^2}{T^\frac{1}{2}}\right)+O\left(\frac{L^\frac{5}{2}(A_2)^2\tau_{max}^2 r(1-\mu)^2}{T^\frac{3}{2}}\right)+A_1.
    \end{align*}
 \end{theorem}

 Please note that when momentum hyper-parameter $\mu=0$, BASGDm degenerates to BASGD. In this case, $1-\mu=1$, and Theorem~\ref{thm:general_proper_m} is exactly the same as Theorem~\ref{thm:general_proper}. From this perspective, Theorem~\ref{thm:general_proper_m} can be deemed as a more general version of Theorem~\ref{thm:general_proper}. Besides, we would also like to point out that the factor $(1-\mu)$ in Theorem~\ref{thm:general_proper_m} does not mean that larger $\mu$ will lead to tighter upper bound, since constants $A_1$ and $A_2$ are dependent on momentum hyper-parameter $\mu$. In fact, the influence of momentum hyper-parameter is a complex problem, which has been studied for decades~\citep{Momentum}. Since it is not the focus of this work, we are not going to further discuss this problem here.

In general cases, Theorem~\ref{thm:general_proper} and Theorem~\ref{thm:general_proper_m} guarantee BASGD and BASGDm to find a point such that the squared $L_2$-norm of its gradient is not larger than $A_1$ in expectation, respectively. Please note that Assumption~\ref{ass:bounded_gradient} already guarantees that gradient's squared $L_2$-norm is not larger than $D^2$. We introduce Proposition~\ref{prop:A1_and_D2} to show that $A_1$ is guaranteed to be smaller than $D^2$ under a mild condition.

\begin{proposition}\label{prop:A1_and_D2}
Assume $Aggr(\cdot)$ is an $(A_1,A_2)$-effective aggregation function, and $\G_{syn}^t$ is aggregated by $Aggr(\cdot)$ in synchronous setting. If $\EB[\|\G_{syn}^t-\nabla F(\w^t)\|~|~\w^t]\leq D,$ $\forall \w^t\in\RB^d$, we have $A_1\leq D^2$.
\end{proposition}
$\G_{syn}^t$ is the aggregated result of $Aggr(\cdot)$, and is a robust estimator of $\nabla F(\w^t)$ used for updating. Since $\|\nabla F(\w^t)\|\leq D$, $\nabla F(\w^t)$ locates in a ball with radius $D$. $\EB[\|\G_{syn}^t-\nabla F(\w^t)\|~|~\w^t]\leq D$ means that the bias of $\G_{syn}^t$ is not larger than the radius $D$, which is a mild condition for $Aggr(\cdot)$. 


\section{Experiment}\label{sec:exp}
    In this section, we empirically evaluate the performance of BASGD (BASGDm) and baselines
    in both image classification~(IC) and natural language processing~(NLP) applications.
    Our experiments are conducted on a distributed platform with dockers.
    Each docker is bound to an NVIDIA Tesla V100~(32G) GPU. 
    We choose $30$ dockers as workers and an extra docker as server\footnote{In the conference version~\citep{yang2020_basgd}, we set $8$ workers in NLP experiment. To make the settings more consistent with that of IC experiment, we also set worker number to $30$ for NLP experiment in this journal version.}. All algorithms are implemented with PyTorch 1.3.
    

\subsection{Experimental Setting}\label{subsec:exp_method}
    Because BASGD (BASGDm) is ABL methods, SBL methods cannot be directly compared with BASGD (BASGDm). The ABL method Zeno++ either cannot be directly compared with BASGD (BASGDm), because Zeno++ needs to store some instances on server. The number of instances stored on server will highly affect the performance of Zeno++~\citep{xie2020zeno++}. Hence, we compare BASGD (BASGDm) with ASGD (ASGDm) and Kardam in our experiments. 
    We set dampening function $\Lambda(\tau)=\frac{1}{1+\tau}$ for Kardam as suggested in~\citep{damaskinos2018asynchronous_Kardam}, and set momentum hyper-parameter $\mu=0.9$ for BASGDm and ASGDm in each experiment. \\

    \textbf{Byzantine attacks.} We will compare BASGD (BASGDm) with baselines under the following different attack settings.
    \begin{itemize}
      \item No attack: In this setting, each worker will strictly follow the method, compute and send the gradient (or momentum) without error.

      \item Random disturbance attack (RD-attack): Byzantine workers with RD-attack will replace the true gradient $\g$ with $\tilde{\g}_{RD}=\g + \g_{rnd}$, where $\g_{rnd}$ is a random vector sampled from normal distribution $\NM(\0,\|\sigma_{atk}\g\|^2\cdot \I)$. Here, $\sigma_{atk}$ is a parameter and $\I$ is an identity matrix. We set $\sigma_{atk}=0.2$ in our experiments. RD-attack can be seen as an accidental failure with expectation $\0$.

      \item Negative gradient attack (NG-attack): Byzantine workers with NG-attack will replace the true gradient $\g$ with $\tilde{\g}_{NG}=-k_{atk}\cdot\g$, where $k_{atk}\in\RB_+$ is a parameter. We set $k_{atk}=10$ in our experiments. NG-attack is a typical kind of malicious attack. In some previous works, this type of attack is also called bit-flipping attack~\citep{xie2020zeno++,karimireddy2020_learning_history}.

      \item `Fall of Empires'~(FoE) attack~\citep{xie2020_FoE}: Byzantine workers with FoE attack will replace the gradient $\g$ with $\tilde{\g}_{FoE}=- \frac{\epsilon}{|\LM|}\sum_{i\in\LM}\g_i$, where $\LM$ is the index set of loyal workers and $\g_i$ is the gradient computed by the $i$-th worker at the same iteration. We set hyper-parameter $\epsilon=6$ for FoE attack in the experiments of this work. FoE is a type of omniscient attack originally proposed in synchronous settings, which require the gradients computed by loyal workers at the same iteration as omniscient knowledge. Thus, FoE cannot be directly adopted in asynchronous settings. To deal with this problem, we use the last sent gradient (or momentum) from each loyal worker as the omniscient knowledge for FoE. 

      \item `A Little is Enough'~(ALIE) attack~\citep{baruch2019little_ByztAttack}: Byzantine workers with ALIE attack will replace the gradient $\g$ with $\tilde{\g}_{ALIE}$, where $(\tilde{\g}_{ALIE})_j={mean}_j-z^{max}\cdot{std}_j$. The sub-index $(\cdot)_j$ denotes the $j$-th coordinate of the vector. The scalars ${mean}_j$ and ${std}_j$ are the mean and standard error of the $j$-th coordinate of loyal workers' gradients at the same iteration, respectively. $z^{max}=\Phi^{-1}(\frac{m-\lfloor m/2+1\rfloor}{m-r})$, where $\Phi^{-1}(\cdot)$ is the inverse of the standard normal cumulative distribution function, $m$ is the number of workers, and $r$ is the number of Byzantine workers. ALIE is also a type of omniscient attack originally proposed in synchronous settings. Similarly, to make it compatible with asynchronous settings, we use the last sent gradient (or momentum) from each loyal worker as the omniscient knowledge for ALIE.
    \end{itemize}

    In real world applications, it is usually hard to adopt the two types of omniscient attacks~(FoE and ALIE) due to the lack of omniscient knowledge. However, we still compare the performance of different methods under these two attacks to test resilience ability.

    \textbf{Aggregation rules.} In the experiments, BASGDm is tested with each of the following aggregation rules.
    \begin{itemize}
      \item Coordinate-wise $q$-trimmed-mean~(trmean): Please refer to Definition~\ref{def:cwTrmean}.
      \item Coordinate-wise median~(median): Please refer to Definition~\ref{def:cwMedian}. Since median can be deemed as a special case of trmean, we only report the results of BASGD (BASGDm) with median in the no attack case\footnote{In the conference version~\citep{yang2020_basgd}, we report the results of BASGD with median in all cases. In this journal version, we test BASGD (BASGDm) with two more aggregation rules (geometric median and centered clipping). Due to limited space in each single figure, we do not report the results of BASGD (BASGDm) with median for better readability in this journal version. The performance of median is similar to that of other aggregation rules.}.
      \item Geometric median~(geoMed)~\citep{chen2017distributed_geoMed}: The geometric median of $B$ vectors $\h_1,\ldots,\h_B\in\RB^d$ is defined as: 
      \begin{equation}\label{eq:geoMed}
        \text{geoMed}([\h_1,\ldots,\h_B])=\mathop{\arg\min}_{\h\in\RB^d}\left\{ \sum_{b=1}^B \|\h-\h_b\|_2\right\}.
      \end{equation}
      The optimization problem defined in the right-hand side of (\ref{eq:geoMed}) has a unique solution when vectors $\{\h_1,\ldots,\h_B\}$ do not lie in a line. However, geoMed usually does not have a closed-form solution. We use Weiszfeld's algorithm~\citep{pillutla2019robust_weiszfeld} to compute it and set the iteration number in Weiszfeld's algorithm to be $5$. 
      \item Centered clipping~(CC)~\citep{karimireddy2020_learning_history}: The CC aggregation result of vectors $\{\h_1,\ldots,\h_B\}$ is given by the following iteration formula: 
      \begin{equation}\label{eq:CC}
        \h^{l+1}=\h^l+\frac{1}{B}\sum_{b=1}^B (\h_b - \h^l)\min\left(1,\frac{R}{\|\h_b-\h^l\|_2}\right).
      \end{equation}
      We set initial point $\h^0$ to be the last aggregation result for quicker convergence as suggested in~\citep{karimireddy2020_learning_history}. The iteration number is set to be $5$ in IC task and $50$ in NLP task. Clipping size $R$ is set to be $0.5$. 
    \end{itemize}

    In addition, to simulate an unstable network environment where asynchronous methods are usually preferred, each worker is manually set to have a delay, which is $k_{del}$ times the computing time. Training set is randomly and equally distributed to different workers.
    We use the average top-$1$ test accuracy~(in IC) or average perplexity~(in NLP) on all workers w.r.t. epochs as final metrics. Average training loss w.r.t. epochs in IC experiment can be found in Appendix~\ref{appendix:more_exp_results}, which is consistent with the average top-$1$ test accuracy results presented in this section. For BASGD (BASGDm), reassignment interval is set to be $1$ second in the IC experiment and $5$ seconds in the NLP experiment. 

    
\subsection{Image Classification Experiment}\label{subsec:IC_exp}

\begin{figure}[t]
  \begin{center}
  \subfigure[BASGD with median]{
  \includegraphics[width=0.460\linewidth]{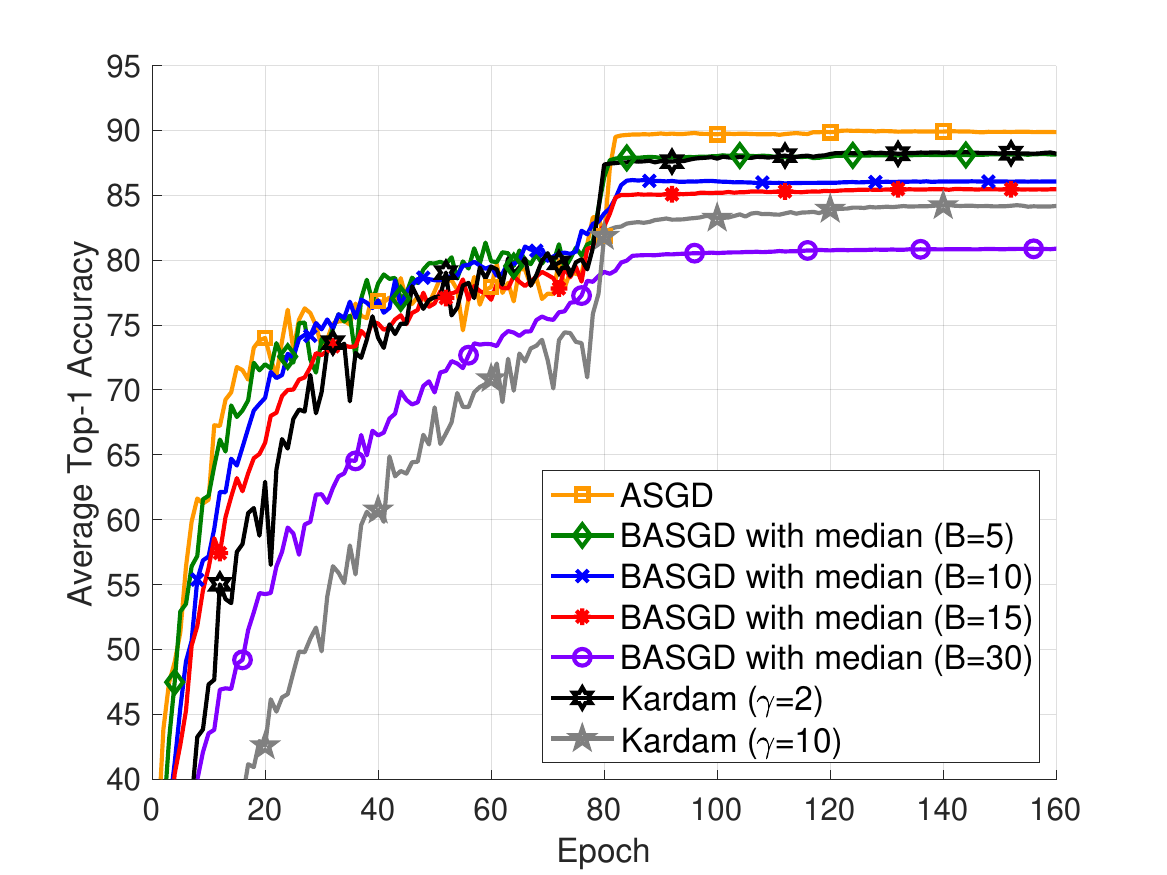}
  \label{fig:noAtk_med}
  }
  \subfigure[BASGD with trmean]{
  \includegraphics[width=0.460\linewidth]{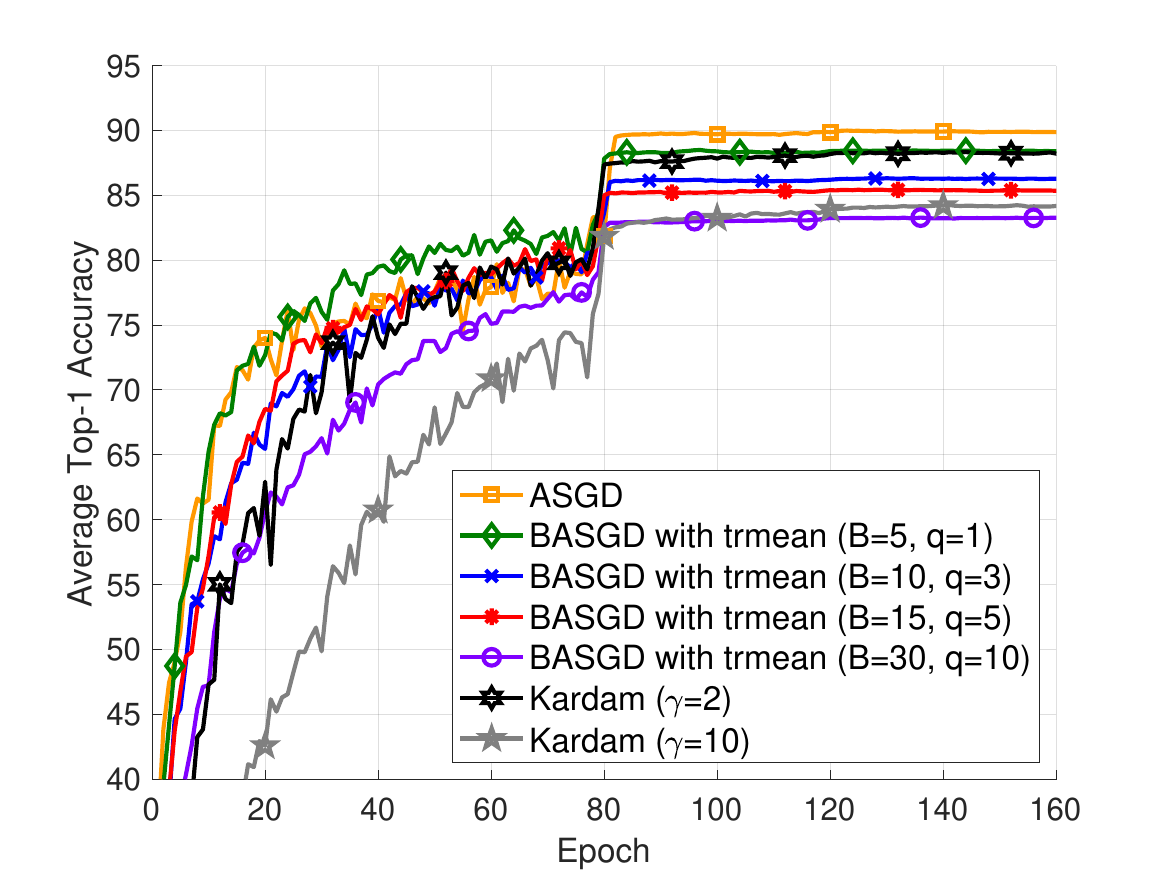}
  \label{fig:noAtk_trm}
  }
  \subfigure[BASGD with geoMed]{
  \includegraphics[width=0.460\linewidth]{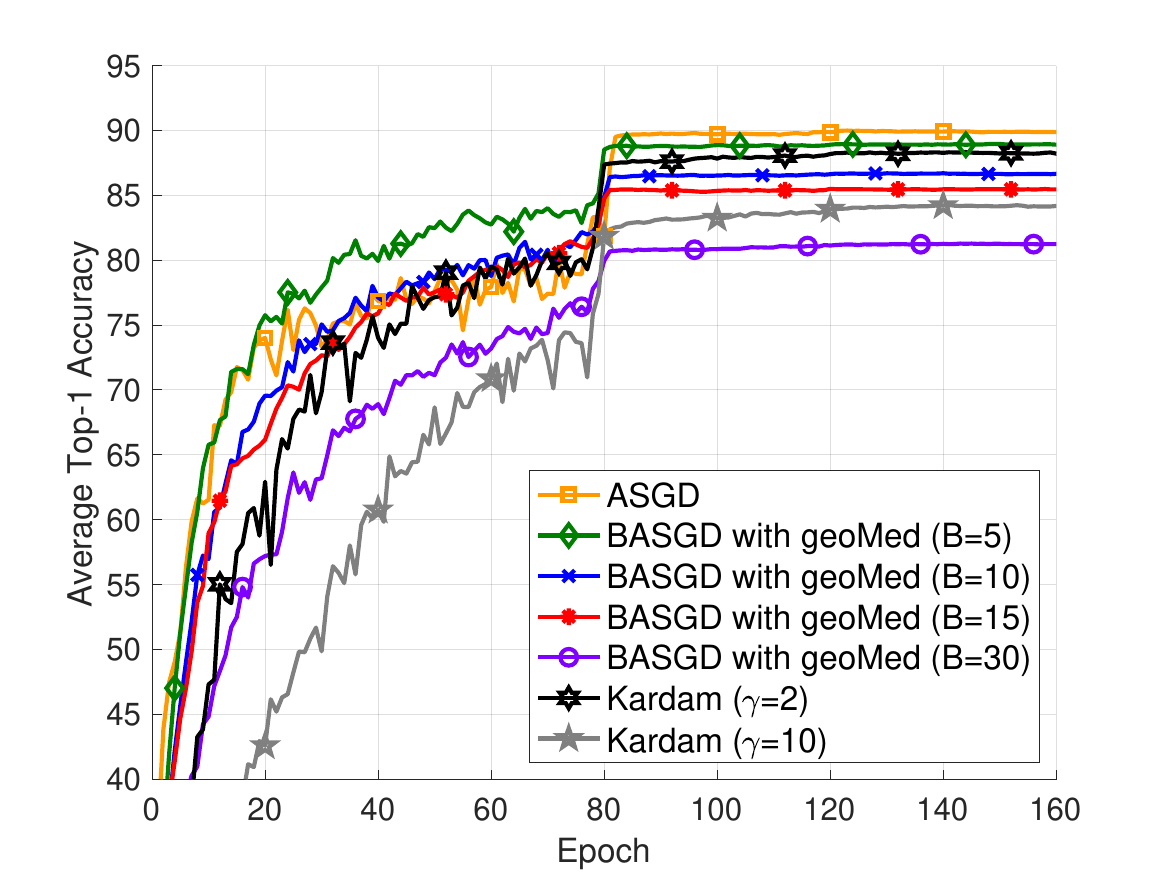}
  \label{fig:noAtk_geoMed}
  }
  \subfigure[BASGD with CC]{
  \includegraphics[width=0.460\linewidth]{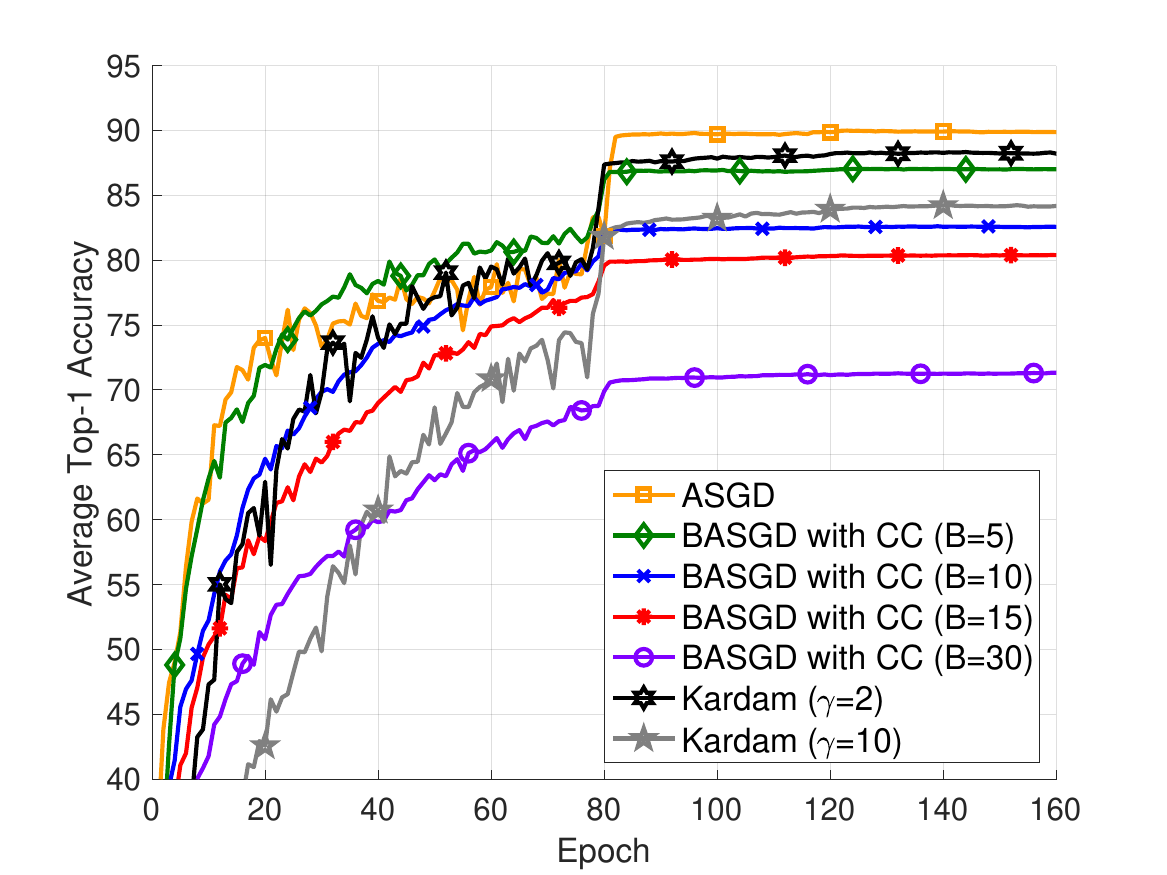}
  \label{fig:noAtk_CC}
  }
  \caption{Average top-$1$ test accuracy w.r.t. epochs of methods BASGD, ASGD, and Kardam when there are no Byzantine workers.
  }
  \label{fig:noAtk_1}
  \end{center}
  \end{figure}

\begin{figure}[t]
  \begin{center}
  \subfigure[BASGDm with median]{
    \includegraphics[width=0.460\linewidth]{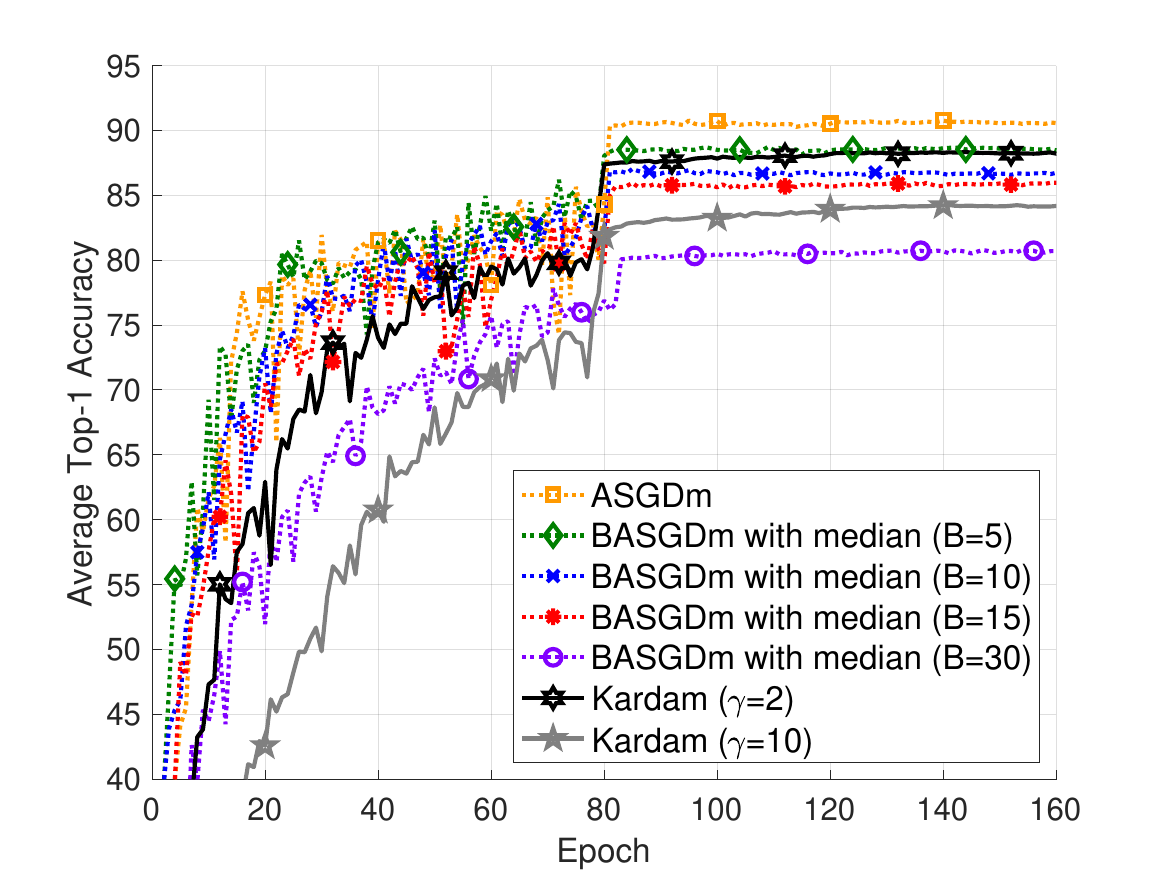}
    \label{fig:noAtk_med_M}
  }
  \subfigure[BASGDm with trmean]{
    \includegraphics[width=0.460\linewidth]{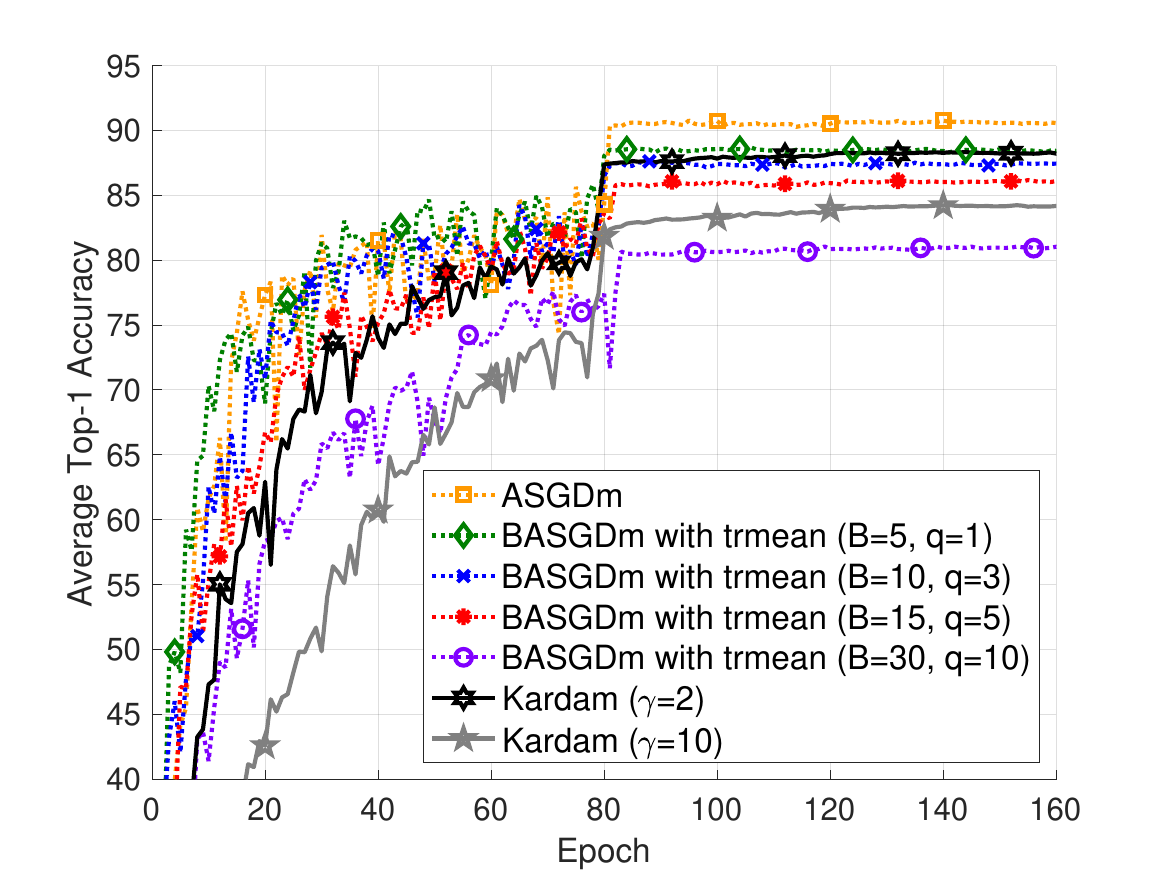}
    \label{fig:noAtk_trm_M}
  }
  \subfigure[BASGDm with geoMed]{
    \includegraphics[width=0.460\linewidth]{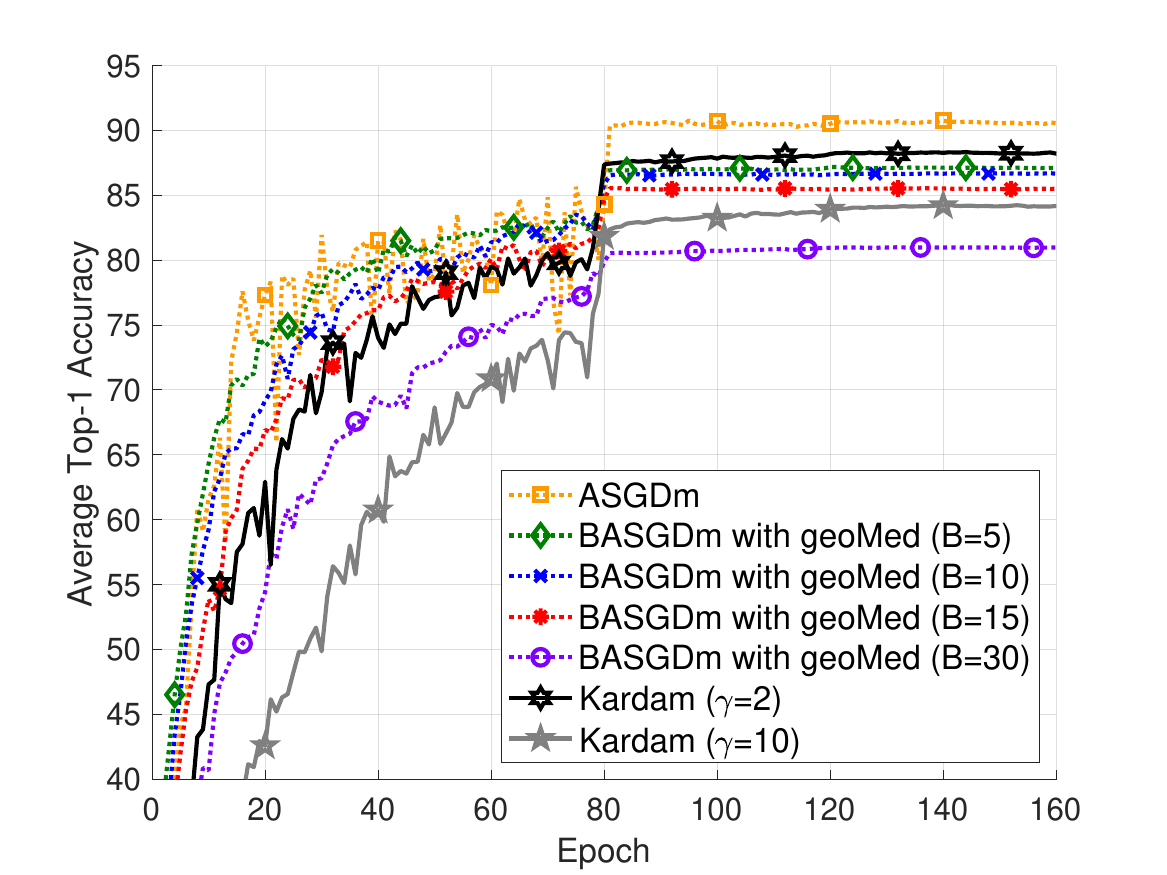}
    \label{fig:noAtk_geoMed_M}
  }
  \subfigure[BASGDm with CC]{
    \includegraphics[width=0.460\linewidth]{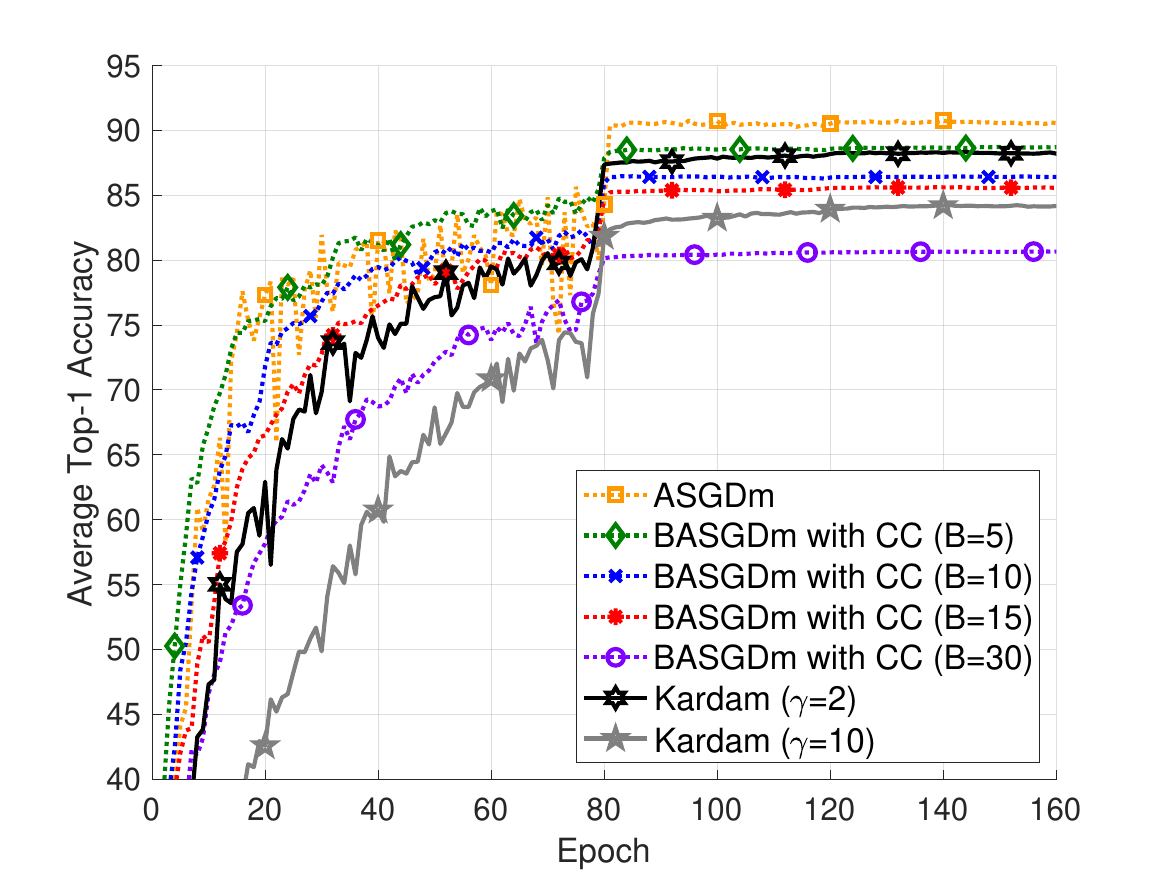}
    \label{fig:noAtk_CC_M}
  }
  \caption{Average top-$1$ test accuracy w.r.t. epochs of methods BASGDm, ASGDm, and Kardam when there are no Byzantine workers.
  }
  \label{fig:noAtk_2}
  \end{center}
  \end{figure}

  In this part, we will empirically compare the performance of BASGD (BASGDm) and existing asynchronous methods ASGD and Kardam in image classification tasks.

    In the experiment, algorithms are evaluated on CIFAR-10
    \citep{krizhevsky2009learning_cifar10}
    with deep learning model \mbox{ResNet-20}~\citep{he2016deep_resnet}.
    Cross-entropy is used as the loss function.
    $k_{del}$ is randomly sampled from truncated standard normal distribution within $[0,+\infty)$.
    As suggested in~\citep{he2016deep_resnet}, learning rate $\eta$ is set to $0.1$ initially for each algorithm,
    and multiplied by 0.1 at the $80$-th epoch and the $120$-th epoch respectively.
    The weight decay is set to $10^{-4}$. We run each algorithm for $160$ epochs.
    Batch size is set to $25$. 
  
    Firstly, we compare the performance of different methods when there are no Byzantine workers.
    Experimental results of BASGD and BASGDm are illustrated in Figure~\ref{fig:noAtk_1} and Figure~\ref{fig:noAtk_2}, respectively. 
    The solid line represents that the method does not use momentum while the dotted line represents that the method utilizes local momentum. 
    ASGD (ASGDm) achieves the best performance. \mbox{BASGD (BASGDm)}~($B>1$) and Kardam have similar convergence rate to ASGD (ASGDm), but both sacrifice a little accuracy. 
    Furthermore, the performance of BASGD (BASGDm) gets worse when the buffer number $B$ increases, which is consistent with the theoretical results.
    Please note that ASGD (ASGDm) is a degenerated case of BASGD (BASGDm) when $B=1$ and $Aggr(\h_1)=\h_1$. 
    Hence, BASGD (BASGDm) can achieve the same performance as ASGD (ASGDm) when there is no failure or attack. The wall-clock-time of running $160$ epochs is reported in Table~\ref{table:running_time}. The time cost of BASGDm is slightly larger than that of ASGD, while Kardam takes the most time.

    \begin{table}[t]
    \caption{Wall-clock-time of running $160$ epochs for different methods (in seconds)}
    \label{table:running_time}
    \begin{center}
    \begin{small}
    \begin{sc}
    \begin{tabular}{c|c|ccc|cc}
      
        \toprule
        \rule{0pt}{10pt}\multirow{2}{*}{Method}  & \multirow{2}{*}{ASGD} & \multicolumn{3}{c|}{BASGDm~($B=10$)} & \multicolumn{2}{c}{Kardam} \\ \cline{3-7} 
        \rule{0pt}{10pt} &  & w/ trmean  & w/ geoMed   & w/ CC   & $\gamma=2$   & $\gamma=10$  \\ 
        \midrule
        Wall-clock-time & 1172.30               & 1191.01       & 1287.07       & 1289.32   & 1522.05      & 1535.22      \\ 
        \bottomrule
    \end{tabular}
    \end{sc}
    \end{small}
    \end{center}
    \end{table}



  \begin{figure}[t]
    \begin{center}
    \subfigure[$3$ Byzantine workers with RD-attack]{
    \includegraphics[width=0.460\linewidth]{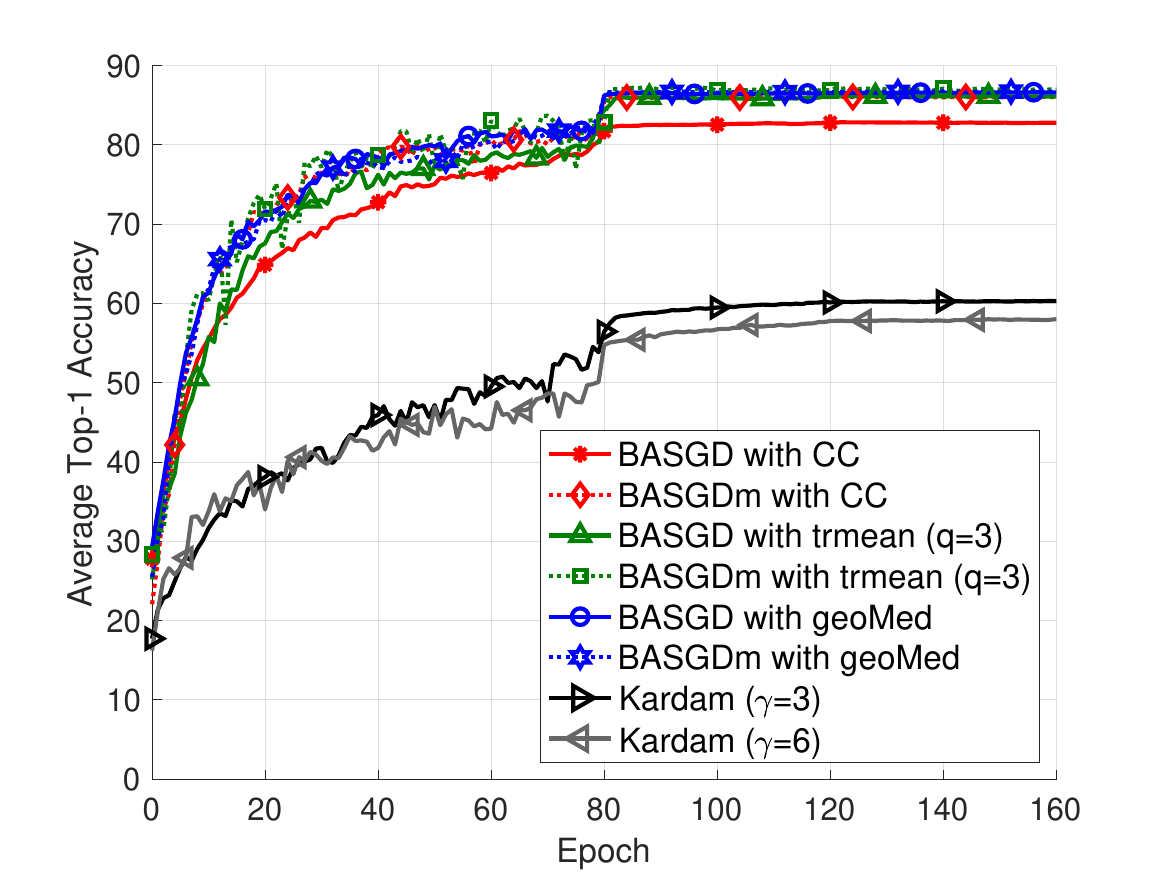}
    }
    \subfigure[$3$ Byzantine workers with NG-attack]{
    \includegraphics[width=0.460\linewidth]{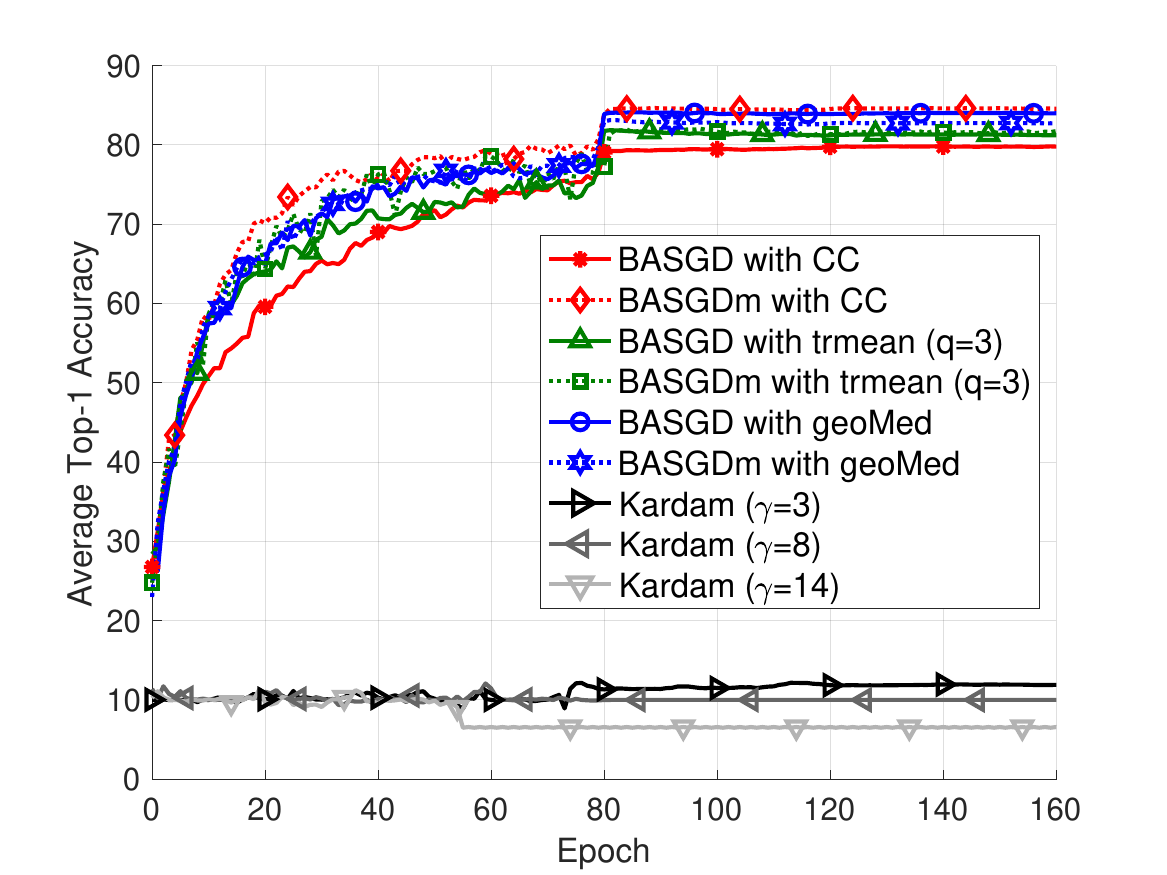}
    }
    \subfigure[$6$ Byzantine workers with RD-attack]{
    \includegraphics[width=0.460\linewidth]{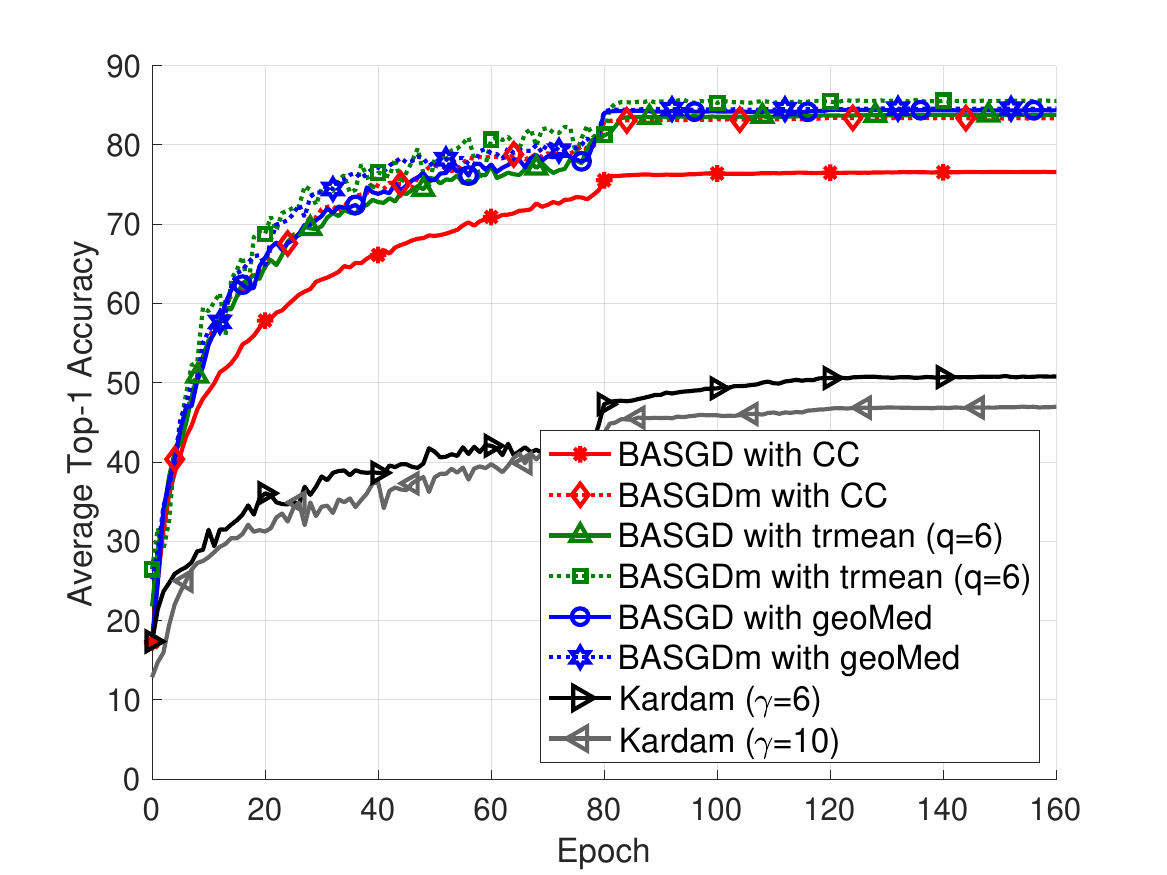}
    }
    \subfigure[$6$ Byzantine workers with NG-attack]{
    \includegraphics[width=0.460\linewidth]{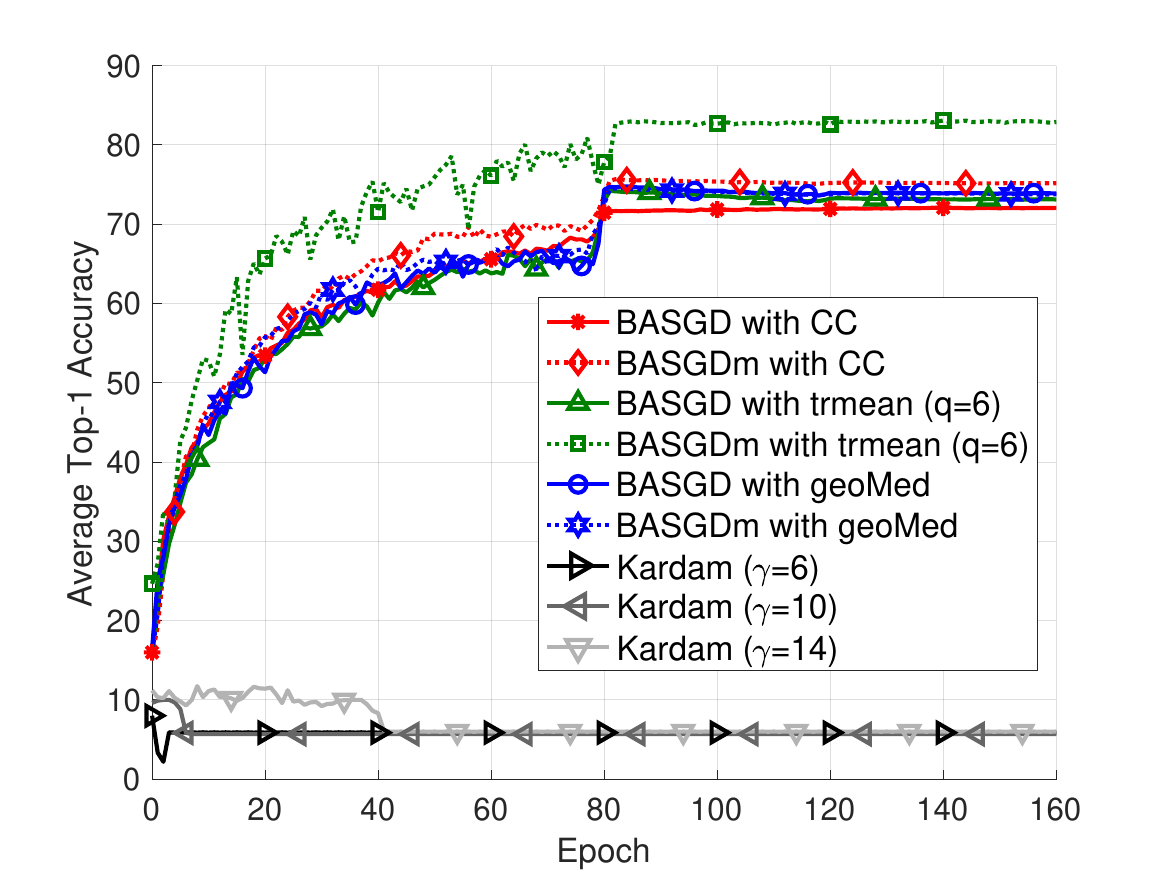}
    }
    \caption{Average top-$1$ test accuracy w.r.t. epochs under non-omniscient attacks. $B=10$ for BASGD (BASGDm) when there are $3$ Byzantine workers and $B=15$ for BASGD (BASGDm) when there are $6$ Byzantine workers.
    }\label{fig:IC_non_omniscient}
    \end{center}
  \end{figure}

    Then, for each type of attack, we compare the performance of BASGD (BASGDm) and Kardam by conducting two experiments in which there are $3$ and $6$ Byzantine workers, respectively\footnote{In the conference version~\citep{yang2020_basgd}, we also report the experimental results of ASGD under attacks. However, due to limit space in figures, we do not report the results of ASGD (ASGDm) in this journal version for better readability since ASGD (ASGDm) is not Byzantine-resilient.}.
    We respectively set $10$ and $15$ buffers for BASGD (BASGDm) in these two experiments.
    The experimental results of the methods under two types of non-omniscient attacks~(RD-attack and NG-attack) are presented in Figure~\ref{fig:IC_non_omniscient}.
    We can find that BASGD (BASGDm) significantly outperform Kardam under these two types of non-omniscient attacks.

    Under the less harmful RD-attack, although Kardam still converge, it suffers a significant loss on accuracy.
    Under NG-attack, Kardam cannot converge even if we have tried different values of \emph{assumed Byzantine worker number} for Kardam, which is denoted by the hyper-parameter $\gamma$ in this paper. 
    Hence, \mbox{Kardam} cannot resist these two types of attacks.
    On the contrary, BASGD still has a relatively good performance under both types of non-omniscient attacks.

    Moreover, we count the ratio of filtered gradients in Kardam, which is shown in Table~\ref{table:rejected count}.
    We can find that in order to filter Byzantine gradients, Kardam also filters approximately equal ratio of loyal gradients. 
    It explains why Kardam performs poorly under the attack.

  \begin{table}[t]
    \caption{Filtered ratio in Kardam under NG-attack in IC task~($3$ Byzantine workers)}
    \label{table:rejected count}
    \begin{center}
    \begin{small}
    \begin{sc}
    \begin{tabular}{c|ccc}
      \toprule
      Term & By Frequency Filter & By Lipschitz Filter & In total \\
      \midrule
      Loyal Grads~($\gamma=3$) &$10.15\%~(31202/307530)$ &$40.97\%~(126000/307530)$ &$51.12\%$ \\
      Byzt Grads~($\gamma=3$) &$10.77\%~(3681/34170)$ &$40.31\%~(13773/34170)$ &$51.08\%$ \\
      \midrule
      Loyal Grads~($\gamma=8$) &$28.28\%~(86957/307530)$ &$28.26\%~(86893/307530)$ &$56.53\%$ \\
      Byzt Grads~($\gamma=8$) &$28.38\%~(9699/34170)$ &$28.06\%~(9588/34170)$ &$56.44\%$ \\
      \midrule
      Loyal Grads~($\gamma=14$) &$85.13\%~(261789/307530)$ &$3.94\%~(12117/307530)$ &$89.07\%$ \\
      Byzt Grads~($\gamma=14$) &$84.83\%~(28985/34170)$ &$4.26\%~(1455/34170)$ &$89.08\%$ \\
      \bottomrule
    \end{tabular}
    \end{sc}
    \end{small}
    \end{center}
    \end{table}

    \begin{figure}[t]
      \begin{center}
      \subfigure[$3$ Byzantine workers with FoE attack]{
      \includegraphics[width=0.460\linewidth]{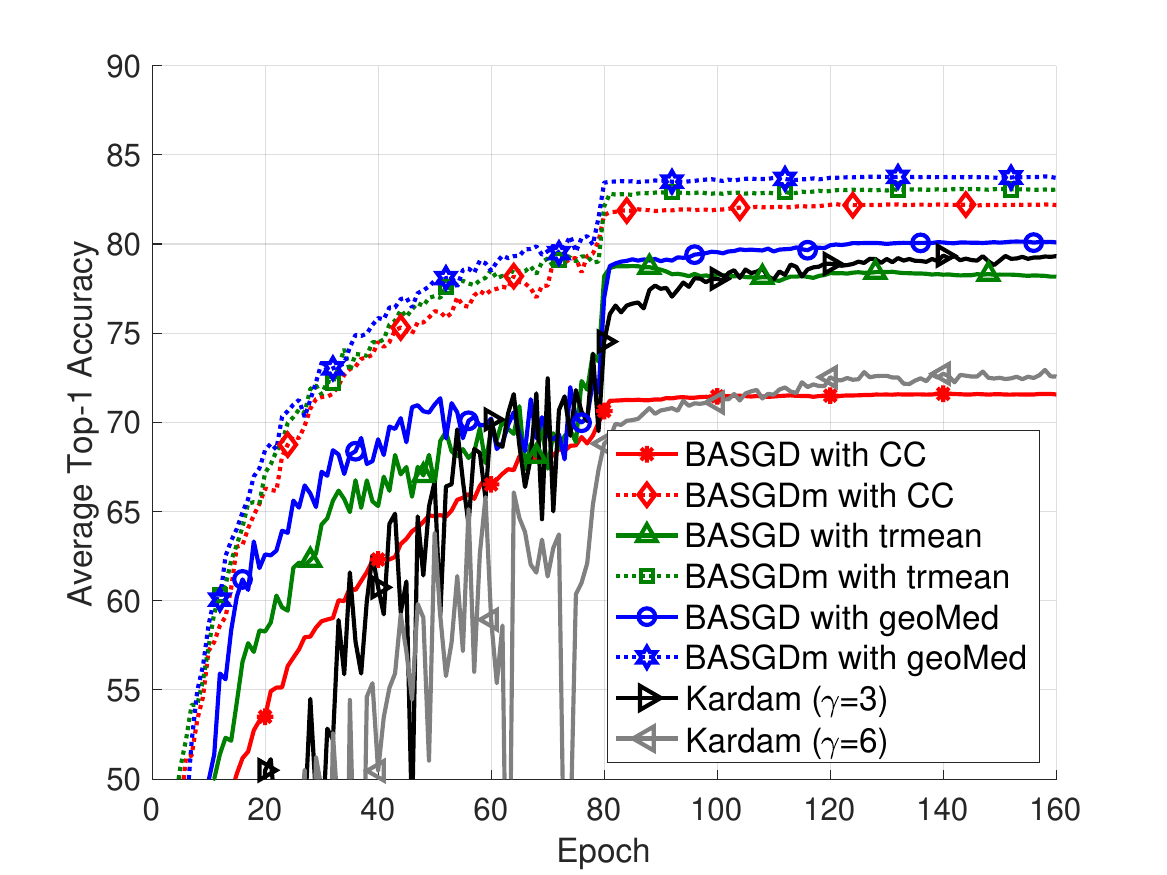}
      }
      \subfigure[$3$ Byzantine workers with ALIE attack]{
      \includegraphics[width=0.460\linewidth]{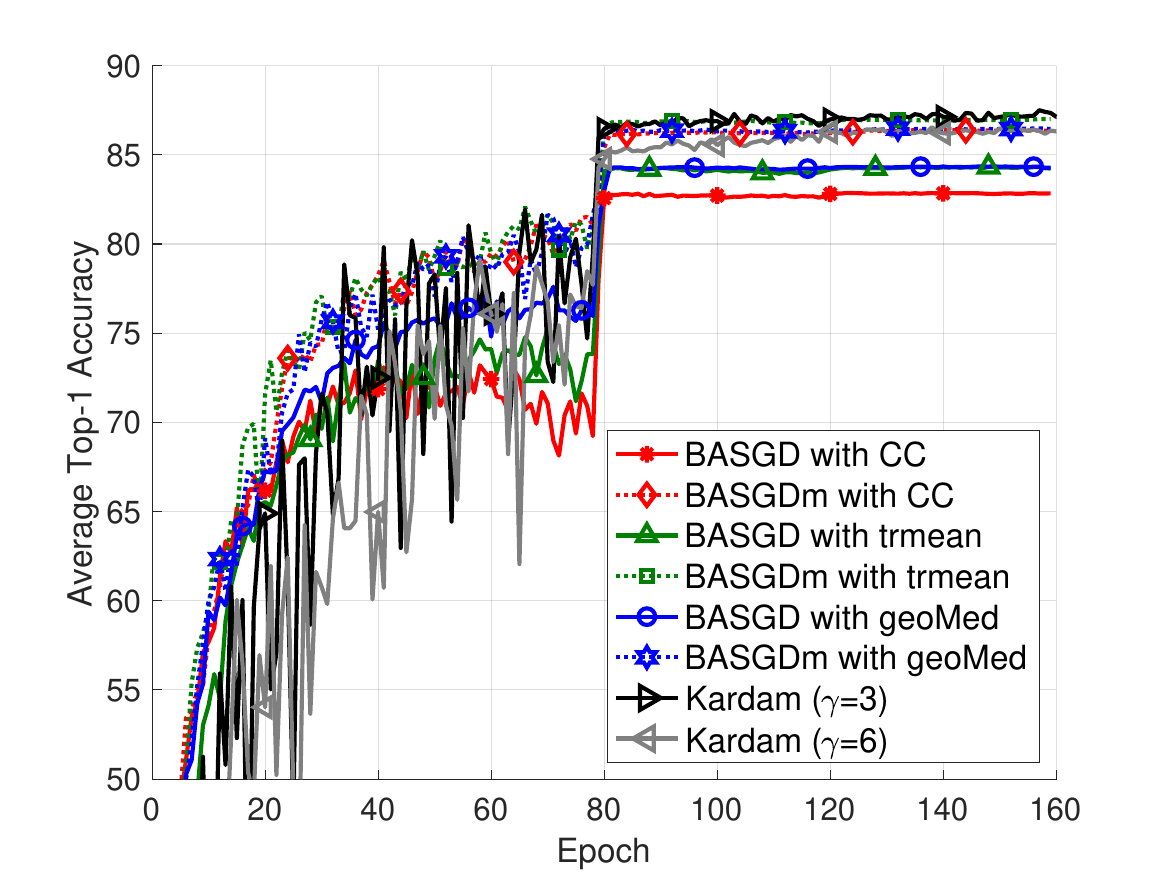}
      }
      \subfigure[$6$ Byzantine workers with FoE attack]{
      \includegraphics[width=0.460\linewidth]{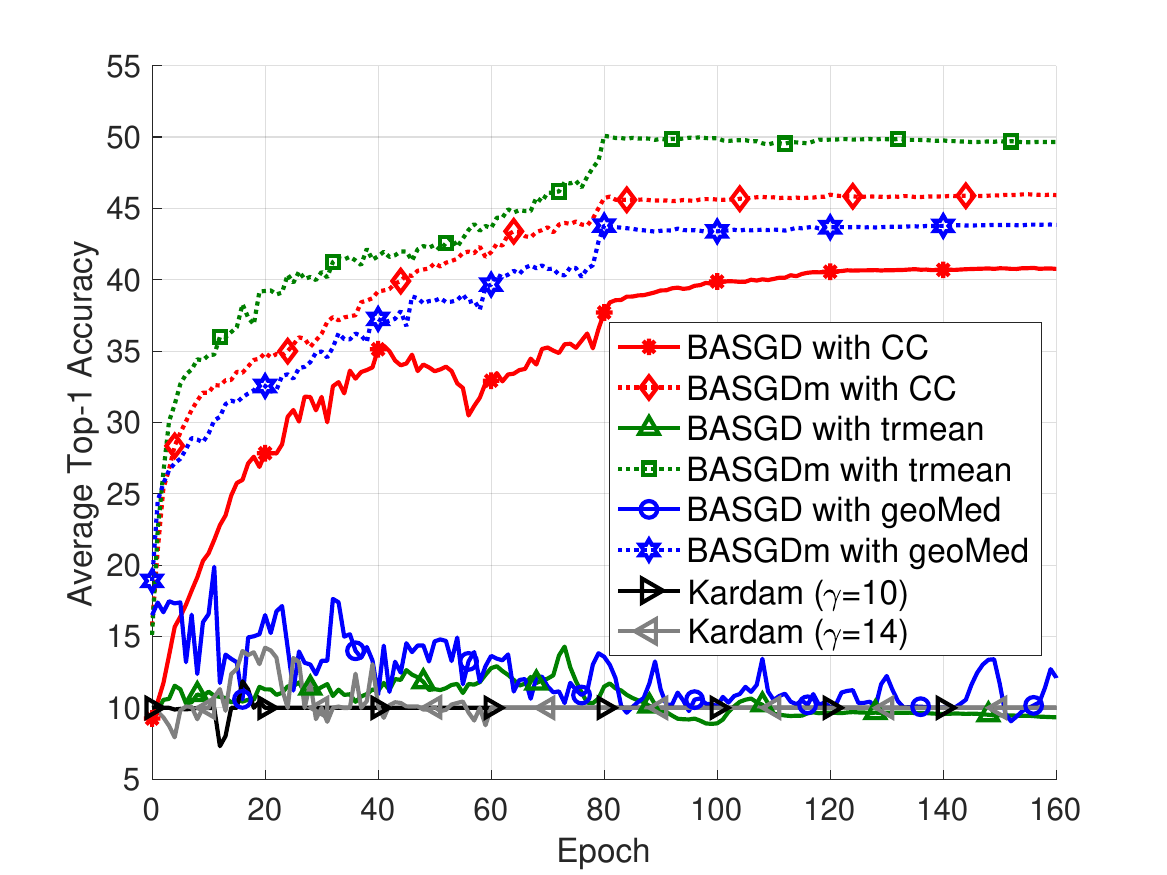}
      }
      \subfigure[$6$ Byzantine workers with ALIE attack]{
      \includegraphics[width=0.460\linewidth]{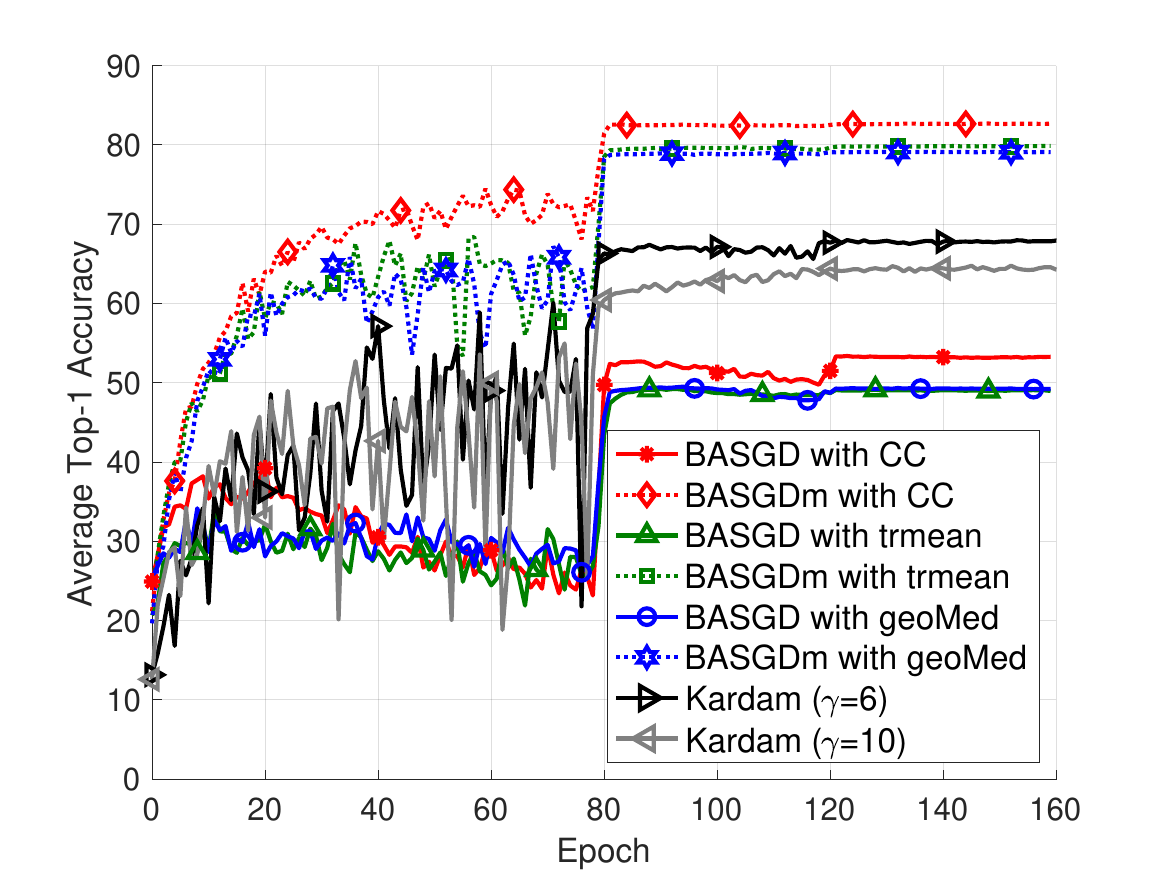}
      }
      \caption{Average top-$1$ test accuracy w.r.t. epochs under omniscient attacks. $B=10$ for BASGD (BASGDm) when there are $3$ Byzantine workers and $B=15$ for BASGD (BASGDm) when there are $6$ Byzantine workers.
      }\label{fig:IC_omniscient}
      \end{center}
    \end{figure}

    We also compare the performance of different methods under omniscient attacks~(FoE attack and ALIE attack), the results of which are shown in Figure~\ref{fig:IC_omniscient}. BASGDm can significantly outperform other methods in each case, except for the case of $3$ Byzantine workers with ALIE attack. When there are $3$ Byzantine workers with ALIE attack, all the methods have a comparable performance to each other. The main reason is that the Byzantine attack is not strong enough in this case. In addition, the performance of BASGDm is considerably better than BASGD. This reveals that using history information~(such as momentum) can strengthen the resilience and improve the performance in Byzantine-resilient machine learning, which is consistent with previous works~\citep{allen2020byzantine,el2020distributed_byzMomentum,karimireddy2020_learning_history}. Moreover, although the performance of BASGDm with different aggregation rules~(trmean, geoMed, and CC) slightly differ, all of them can outperform BASGD and Kardam.

\subsection{Natural Language Processing Experiment}\label{subsec:NLP_exp}

\begin{figure}[t]
  \begin{center}
  \subfigure[Under RD-attack]{
    \includegraphics[width=0.460\linewidth]{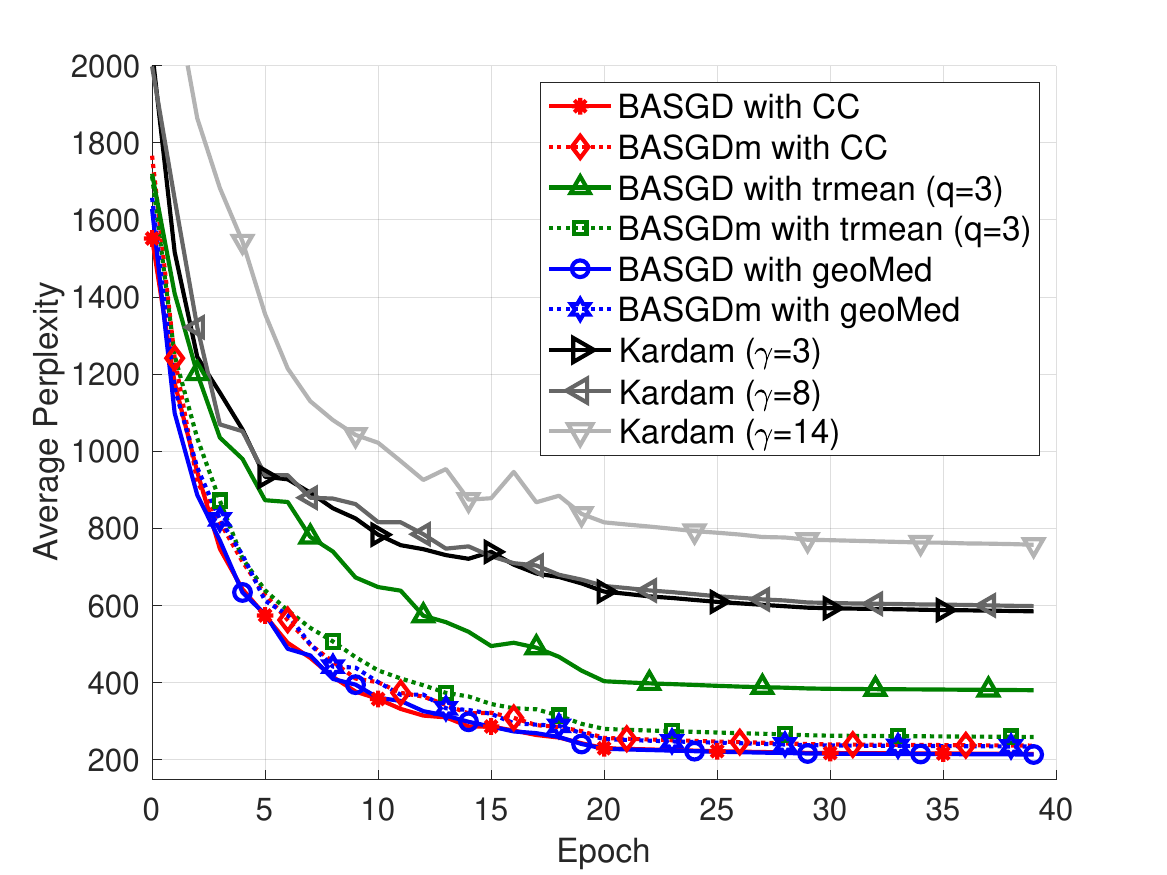}
    \label{fig:NLP_RD}
  }
  \subfigure[Under NG-attack]{
    \includegraphics[width=0.460\linewidth]{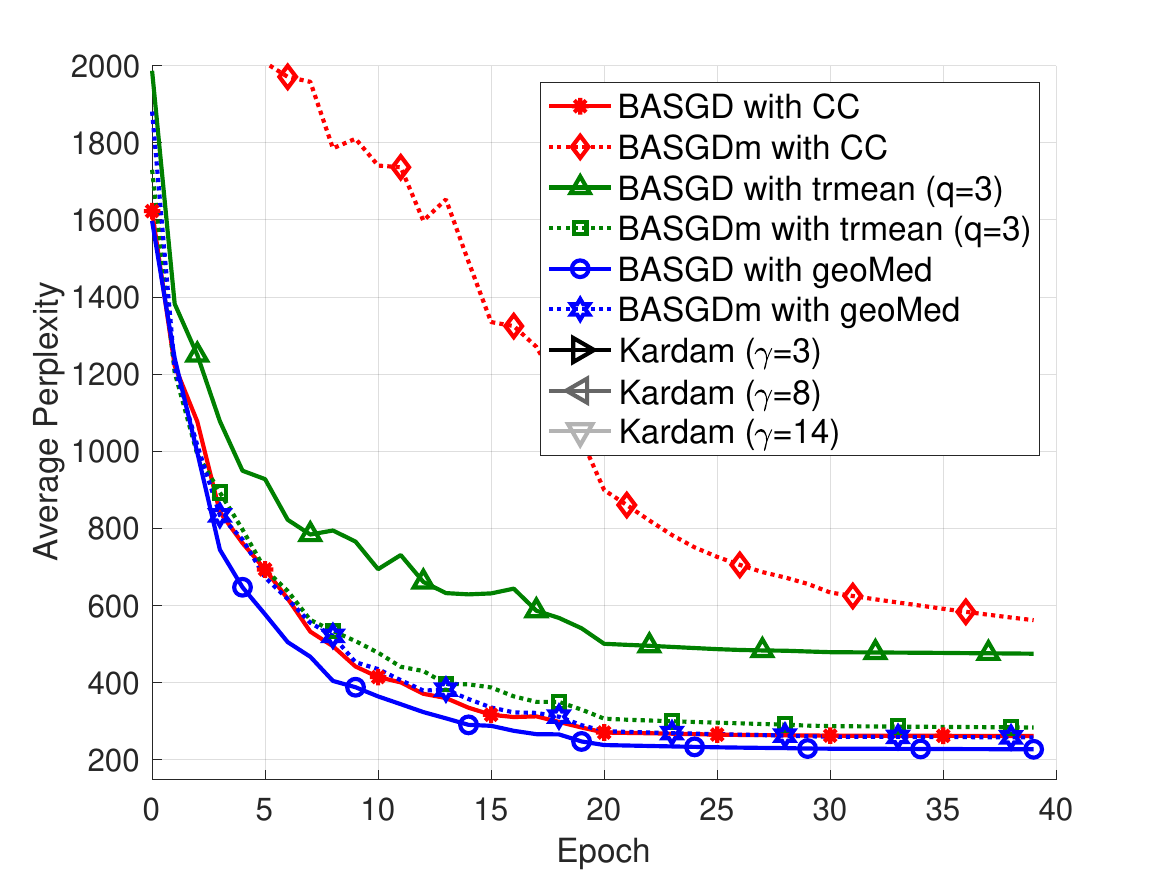}
    \label{fig:NLP_NG}
  }
  \subfigure[Under FoE attack]{
    \includegraphics[width=0.460\linewidth]{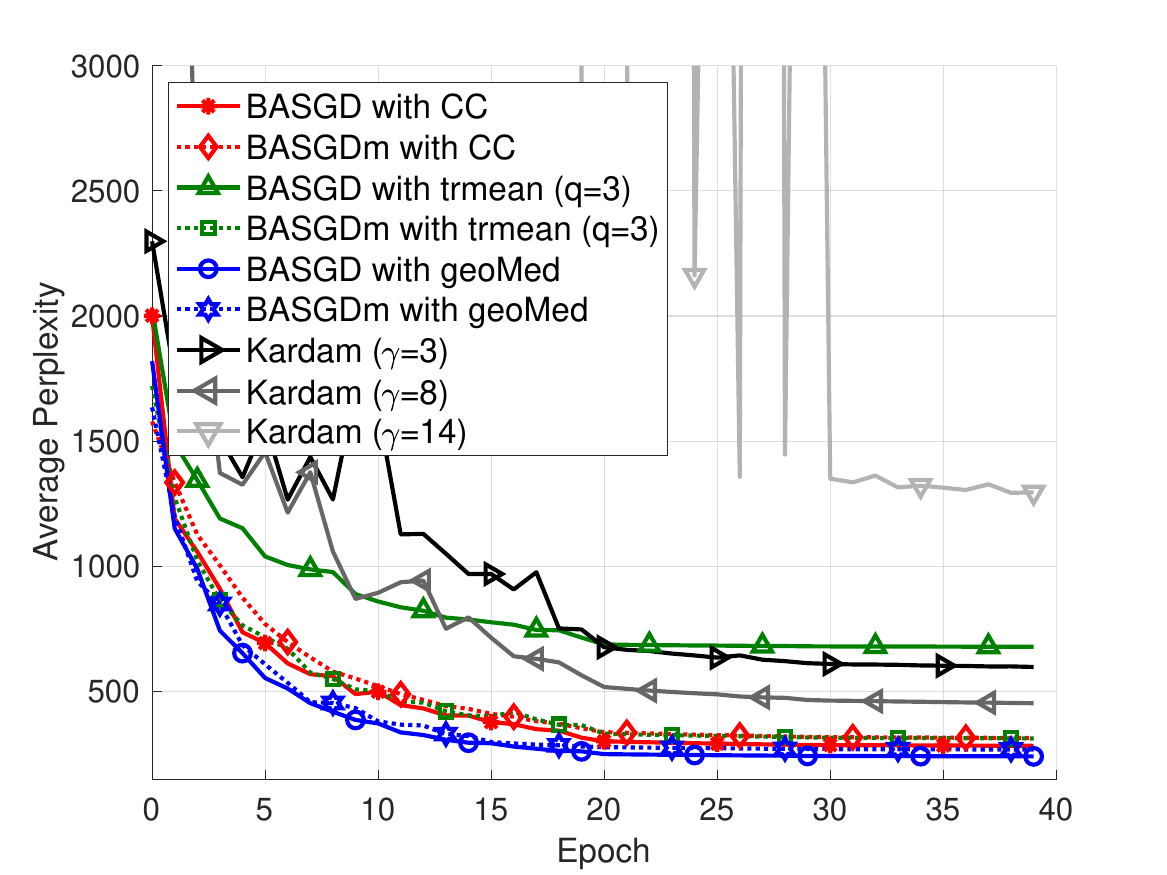}
    \label{fig:NLP_FoE}
  }
  \subfigure[Under ALIE attack]{
    \includegraphics[width=0.460\linewidth]{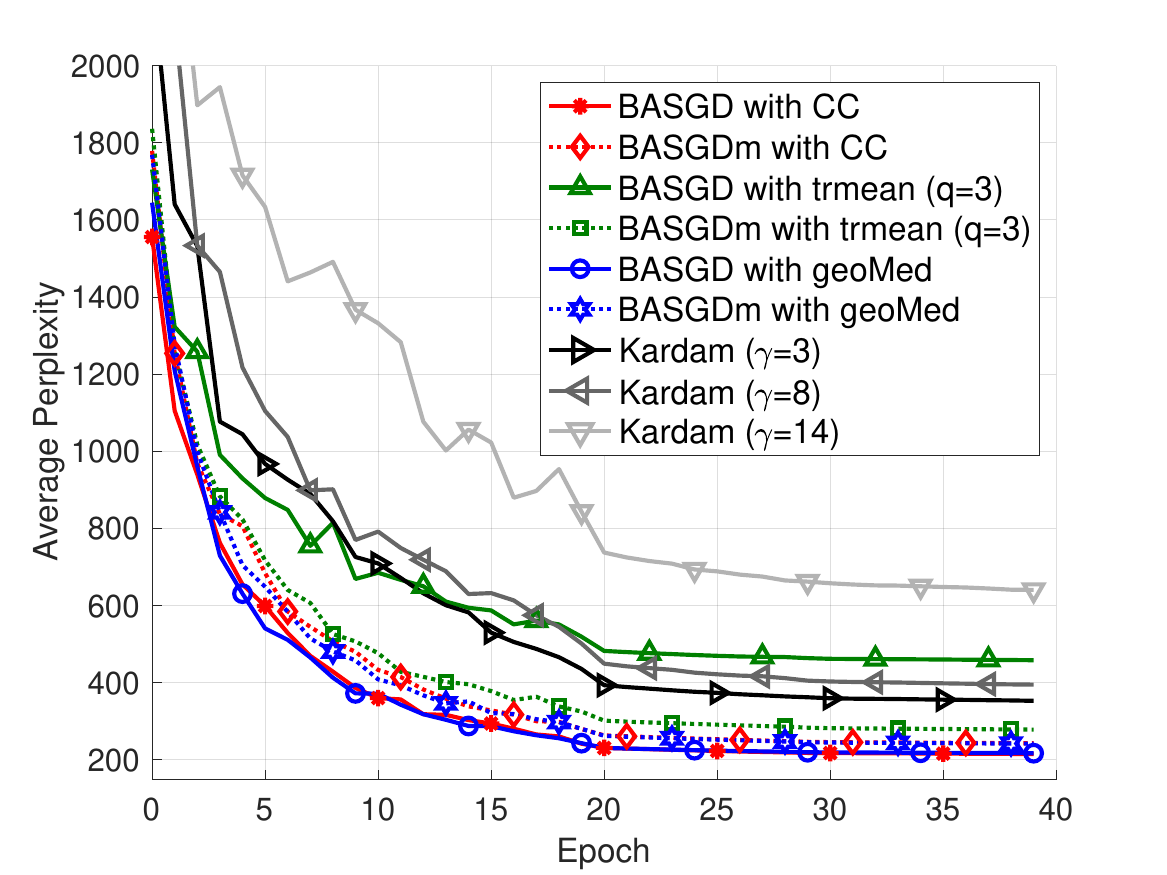}
    \label{fig:NLP_ALIE}
  }
  
  \caption{Average perplexity w.r.t. epochs ($3$ Byzantine workers, $B=15$ for BASGD and BASGDm).
  In subfigure~\ref{sub@fig:NLP_NG}, the curves representing Kardam do not appear because Kardam diverges in this case and the average perplexity explodes.}
  \label{fig:BASGDm_NLP}
  \end{center}
\end{figure}
    In this part, we will empirically compare the methods on natural language processing~(NLP) tasks.
    In our NLP experiment, the methods are evaluated on the \mbox{WikiText-2}
    dataset with an \mbox{LSTM}~\citep{hochreiter1997_LSTM} network. 
    We only use the training set and test set, 
    while the validation set is not used in our experiment.
    For LSTM, we adopt $2$ layers with $100$ units in each layer. Word embedding size is set to $100$, and sequence length is set to $35$.
    Gradient clipping size is set to $0.25$.
    Cross-entropy is used as the loss function.
    We run each algorithm for $40$ epochs.
    Initial learning rate $\eta$ is chosen from $\{1,2,5,10,20\}$ and is divided by $4$ at the $21$-st epoch and the $31$-st epoch. 
    The best result is adopted as the final one. 
    $k_{del}$ is randomly sampled from a standard exponential distribution.
    Similarly, each method is tested under RD-attack, NG-attack, FoE attack, and ALIE attack.
    The average perplexity is reported in Figure~\ref{fig:BASGDm_NLP}.

    As illustrated in Figure~\ref{fig:NLP_RD} and Figure~\ref{fig:NLP_NG}, under the two types of non-omniscient attacks~(RD-attack and NG-attack), BASGD (BASGDm) can outperform Kardam, no matter which of the three aggregation rules is used. Moreover, the curves representing Kardam do not appear in Figure~\ref{fig:NLP_NG} because Kardam diverges under NG-attack and the perplexity explodes. We would also like to clarify that the performance of CC can get further improved by tuning the clipping size hyper-parameter more finely in different settings. However, it requires much computing power and is beyond the scope of this work. Therefore, we fix clipping size $R=0.5$, and this can already make BASGDm with CC outperform Kardam. Theoretically, the best performance of CC can not be worse than geoMed since CC is equivalent to geoMed when the clipping size $R$ is small enough~(please see Appendix~\ref{subsec:proof_CC} for the proof).

  As illustrated in Figure~\ref{fig:NLP_FoE} and Figure~\ref{fig:NLP_ALIE}, under FoE attack and ALIE attack, BASGD can outperform Kardam except for the case of using trmean as aggregation rule. BASGD with trmean performs slightly worse than Kardam. A possible reason is that trmean is sensitive to model dimensions. On the contrary, by using momentum, BASGDm with any aggregation rule can always outperform Kardam.

  The experimental results in this section have shown that BASGD (BASGDm) can outperform asynchronous Byzantine learning baselines under different settings. Moreover, BASGD (BASGDm) is compatible with various aggregation rules, such as trmean, geoMed, and CC. With the benefit of local momentum, BASGDm gets even stronger Byzantine resilience than BASGD, especially under the omniscient attacks FoE and ALIE.

\section{Conclusion}

  In this paper, we propose a novel method called \mbox{BASGD} and an improved variant of \mbox{BASGD} called \mbox{BASGDm}. 
  To the best of our knowledge, BASGD is the first ABL method that can resist non-omniscient attacks without storing any instances on server. With the benefit of local momentum, BASGDm can resist both non-omniscient and omniscient attacks. Compared with those methods which need to store instances on server, \mbox{BASGD} and \mbox{BASGDm} have a wider scope of application.  
  Both \mbox{BASGD} and \mbox{BASGDm} are compatible with various aggregation rules. Moreover, both \mbox{BASGD} and \mbox{BASGDm} are proved to be convergent and be able to resist failure or attack.
  Empirical results show that our methods significantly outperform existing ABL baselines when there exists failure or attack on workers.

  \acks{This work is supported by National Key R\&D Program of China (No. 2020YFA0713900), NSFC-NRF Joint Research Project (No. 61861146001) and NSFC Project (No. 61921006).}



\clearpage
\onecolumn
\appendix
\section{Asynchronous SGD~(ASGD)}\label{appendix:alg_details}
One popular asynchronous method to solve the problem in~(\ref{eq:obj})
under the PS framework is ASGD~\citep{dean2012large_ASGD}, which is presented in Algorithm~{\ref{alg:asyn_sgd}}.

\begin{algorithm}[ht]
  \caption{Asynchronous SGD~(ASGD)}
  \label{alg:asyn_sgd}
\begin{algorithmic}
  \vskip 0.05in
  \STATE {\bfseries  Server:}
  \STATE {\bfseries Initialization:} initial parameter $\w^0$, learning rate $\eta$;
  \STATE Send initial $\w^0$ to all workers;
  \FOR {$t=0$ {\bfseries to} $t_{max}-1$}
      \STATE Wait until a new gradient $\g_{k}^t$ is received from arbitrary worker\_$k$;
      \STATE Execute SGD step: $\w^{t+1}\leftarrow\w^t-\eta\cdot\g_{k}^t$;
      \STATE Send $\w^{t+1}$ back to worker\_$k$;
  \ENDFOR
  \STATE Notify all workers to stop;

  \vskip 0.2in

  \STATE{\bfseries  Worker\_$k$:} ~~$(k=0,1,...,m-1)$
  \REPEAT
      \STATE{Wait until receiving the latest parameter $\w$ from server};
      \STATE{Randomly sample an index $i$ from $\DM_k$;}
      \STATE{Compute $\nabla f(\w;z_i)$;}
      \STATE{Send $\nabla f(\w;z_i)$ to server};
  \UNTIL{receive server's notification to stop}
  \vskip 0.1in
\end{algorithmic}
\end{algorithm}



    

\section{Proof Details}\label{appendix:proof_details}
\subsection{Proof of Proposition~\ref{prop:qbr}}
\begin{proof}
    Firstly, we prove coordinate-wise $q$-trimmed-mean is $q$-BR.
    It is not hard to check that trmean
    satisfies the property (a) in the definition of $q$-BR,
    then we prove that it also satisfies property (b).

    Without loss of generality, we assume $h_{1j},\ldots,h_{Bj}$ are already in descending order.
    By definition, $Trm(h_{\cdot j})$ is the average value of $\MM_j$,
    which is obtained by removing $q$ largest values and $q$ smallest values of $\{h_{ij}\}_{i=1}^B$.
    Therefore,
    $$h_{(q+1)j}=\max_{x\in \MM_j}\{x\}\geq Trm(h_{\cdot j})\geq\min_{x\in \MM_j}\{x\}=h_{(n-q)j}$$
    For any $\SM \subset [B]$ with $|\SM|=B-q$, by Pigeonhole Principle,
    $\SM$ includes at least one of $h_{1j},\ldots,h_{(q+1)j}$,
    and includes at least one of $h_{(n-q)j},\ldots,h_{Bj}$. Therefore,
    $$\max_{s\in\SM} \{h_{sj}\}\geq h_{(q+1)j}; \qquad \min_{s\in\SM} \{h_{sj}\}\leq h_{(n-q)j}.$$
    Combining these two inequalities, we have:
    $$\max_{s\in\SM}\{h_{sj}\}\geq Trm(h_{\cdot j})\geq\min_{s\in\SM} \{h_{sj}\}.$$
    Thus, coordinate-wise $q$-trimmed-mean is $q$-BR.\\
    By definition, coordinate-wise median can be seen as $\lfloor\frac{B-1}{2}\rfloor$-trimmed-mean,
    and thus is $\lfloor\frac{B-1}{2}\rfloor$-BR.
\end{proof}

\subsection{Proof of Lemma~\ref{lemma:variance_bound}}
To begin with, we will introduce a lemma to estimate the ordered statistics.
\begin{lemma}\label{lemma:BOOSV}
    $X_1,\ldots,X_M$ are non-negative, independent and identically distributed~(i.i.d.) random variables sampled from distribution $\DM$,
    and have limited expectation $\EB[X]$.
    Denote the $K$-th largest value in $\{X_1,\ldots,X_M\}$ as $X_{(K)}$,
    then $\EB[X_{(K)}]\leq C_{M,K}\cdot\EB[X]$, where
    \begin{equation*}
        C_{M,K}=\left.
        \begin{cases}
            M, &K=1;\\
            \frac{M!(K-1)^{K-1}(M-K)^{M-K}}{(K-1)!(M-K)!(M-1)^{M-1}},&1<K<\frac{M}{2}.
        \end{cases}
        \right.
    \end{equation*}
    \begin{proof}
        Denote the Probability Density Function~(PDF) and
        Cumulative Density Function~(CDF) of $\DM$ as
        $p(x)$ and $P(x)$, respectively. Then the PDF of $X_{(K)}$ is:
        $$p_{(K)}(x) = \frac{M!}{(K-1)!(M-K)!}[1-P(x)]^{K-1}P(x)^{M-K}p(x).$$
        Thus,
        \begin{align*}
            \EB[X_{(K)}]
            =&\int_{0}^{+\infty}x\cdot p_{(K)}(x)dx\\
            =&\int_{0}^{+\infty}[\frac{M!}{(K-1)!(M-K)!} \cdot[1-P(x)]^{K-1}P(x)^{M-K}]\cdot xp(x)dx&\\
            \overset{(a)}{\leq}&\int_{0}^{+\infty}[\frac{M!}{(K-1)!(M-K)!} \cdot\frac{(K-1)^{K-1}(M-K)^{M-K}}{(M-1)^{M-1}}]\cdot xp(x)dx\\
            =&\frac{M!(K-1)^{K-1}(M-K)^{M-K}}{(K-1)!(M-K)!(M-1)^{M-1}}\cdot\EB[X].
        \end{align*}
        Inequality (a) is derived based on $[1-P(x)]^{K-1}P(x)^{M-K}\leq\frac{(K-1)^{K-1}(M-K)^{M-K}}{(M-1)^{M-1}}$, which is obtained by the following process:\\
        Let $\theta(x)=(1-x)^{K-1}x^{M-K},~x\in[0,1]$.\\
        Then $\theta'(x)=(1-x)^{K-2}x^{M-K-1}[(M-K)(1-x)-(K-1)x].$\\
        Let $\theta'(x)=0$. Solving the equation, we obtain $x=\frac{M-K}{M-1}$, $0$ or $1$.\\
        Also, we have $\theta(0)=\theta(1)=0$, and $\theta(\frac{M-K}{M-1})=\frac{(K-1)^{K-1}(M-K)^{M-K}}{(M-1)^{M-1}}.$\\
        Then we have $\max_{x\in[0,1]}\theta(x) = \theta(\frac{M-K}{M-1})=\frac{(K-1)^{K-1}(M-K)^{M-K}}{(M-1)^{M-1}}.$\\
        Thus, 
        $[1-P(x)]^{K-1}P(x)^{M-K}=\theta(P(x))\leq\frac{(K-1)^{K-1}(M-K)^{M-K}}{(M-1)^{M-1}}$.
    \end{proof}
\end{lemma}

\begin{proposition}\label{prop:estimation_cnk}
  $\forall B,q,r\in\ZB_+, 0\leq r\leq q<\frac{B}{2}$,
  \begin{equation*}
      C_{B-r,q-r+1}\leq \frac{(B-r)\sqrt{B-r+1}}{\sqrt{(B-q-1)(q-r+1)}}.
  \end{equation*}
\end{proposition}
\begin{proof}
    By Stirling's approximation, we have:
    $$\sqrt{2\pi n}\cdot n^n e^{-n} \leq n! \leq e \sqrt{n}\cdot n^n e^{-n},~\forall n\in\ZB_+.$$
    Therefore,
    \begin{equation}
        \sqrt{2\pi n}\cdot e^{-n} \leq \frac{n!}{n^n} \leq e \sqrt{n}\cdot e^{-n},~\forall n\in\ZB_+.\label{ineq:stirling}
    \end{equation}
    By definition of $C_{M,k}$,
    \begin{align*}
        C_{M,K}=&\frac{M!(K-1)^{K-1}(M-K)^{M-K}}{(K-1)!(M-K)!(M-1)^{M-1}}\\
        = & M\cdot\frac{(M-1)!}{(M-1)^{M-1}}\cdot
            \frac{(K-1)^{K-1}}{(K-1)!}\cdot\frac{(M-K)^{M-K}}{(M-K)!}\\
        \leq & M\cdot [e\sqrt{M-1}\cdot e^{-(M-1)}]\cdot\frac{e^{K-1}}{\sqrt{2\pi(K-1)}}\cdot\frac{e^{M-K}}{\sqrt{2\pi(M-K)}}\\
        = & \frac{e}{2\pi}\cdot\frac{M\sqrt{M-1}}{\sqrt{(M-K)(K-1)}},
    \end{align*}
    where the inequality uses Inequality~(\ref{ineq:stirling}).\\
    Case (i). When $r<q$,
    \begin{align*}
        C_{B-r,q-r+1}\leq&\frac{e}{2\pi}\cdot\frac{(B-r)\sqrt{B-r-1}}{\sqrt{(B-q-1)(q-r)}}\\
        \leq&\frac{(B-r)\sqrt{B-r+1}}{\sqrt{(B-q-1)(q-r+1)}}.
    \end{align*}
    Case (ii). When $r=q$, by definition of $C_{M,K}$, we have:
    $$C_{B-r,q-r+1}=C_{B-q,1}=B-q=\frac{(B-r)\sqrt{B-r+1}}{\sqrt{(B-q-1)(q-r+1)}}.$$
    In conclusion, when $r\leq q$, we have:
    $$C_{B-r,q-r+1}\leq\frac{(B-r)\sqrt{B-r+1}}{\sqrt{(B-q-1)(q-r+1)}}.$$
\end{proof}


When $B$ and $q$ are fixed, the upper bound of $C_{B-r,q-r+1}$ will increase when $r$~(number of Byzantine workers) increases.
  Namely, the upper bound will be larger if there are more Byzantine workers.
  When $B$ and $r$ are fixed, $q$ measures the Byzantine Robust degree of
  aggregation function $Aggr(\cdot)$. The factor $[{(B-q-1)(q-r)}]^{-\frac{1}{2}}$
  is monotonically decreasing with respect to $q$, when $q<\frac{B-1+r}{2}$.
  Since $r\leq q<\frac{B}{2}$, the upper bound will decrease when $q$ increases.
  Also, $B-q$ decreases when $q$ increases.
  Namely, the upper bound will be smaller if
  $Aggr(\cdot)$ has a stronger $q$-BR property.

  In the worst case~($q=r$), the upper bound of $C_{B-r,q-r+1}$ is linear to $B$.
  Even in the best case~($r=0, q=\lfloor\frac{B-1}{2}\rfloor$),
  the denominator is about $\frac{B}{2}$ and the upper bound
  of $C_{B-r,q-r+1}$ is linear to $\sqrt{B}$. 
  Thus, larger $B$ might result in larger error. 
  Hence, buffer number is not supposed to be set too large.

Now we prove Lemma~\ref{lemma:variance_bound}.
\begin{proof}
    \begin{align*}
        &\EB[||\G^t||^2~|~\w^t]\\
        =&\EB[||Aggr([\h_1,\ldots,\h_B])||^2~|~\w^t]\\
        =&\sum_{j=1}^d\EB[Aggr([\h_1,\ldots,\h_B])_j^2~|~\w^t],
    \end{align*}
    where $Aggr([\h_1,\ldots,\h_B])_j$ represents the $j$-th coordinate of the aggregated gradient.

    We use $\HM^t$ to denote the credible buffer index set, which is composed by the index of buffers, where the stored gradients are all from loyal workers.
    
    For each $b\in\HM^t$, $\h_b$ has stored $N_b^t$ gradients
    at iteration $t$: $\g_1,\ldots,\g_{N_b^t}$, and we have:
    $$\h_b=\frac{1}{N_b^t}\sum_{i=1}^{N_b^t}\g_i.$$

    Then,
    \begin{align*}
      \EB[\|\h_b\|^2~|~\w^t]
      =&\EB[\|\h_b-\EB[\h_b~|~\w^t]\|^2~|~\w^t]+\|\EB[\h_b~|~\w^t]\|^2\\
      =&\EB[\|\frac{1}{N_b^t}\sum_{i=1}^{N_b^t}(\g_i-\EB[\g_i~|~\w^t])\|^2~|~\w^t]
      +\|\EB[\frac{1}{N_b^t}\sum_{i=1}^{N_b^t}\g_i~|~\w^t]\|^2\\
      \overset{(a)}{\leq}&\frac{\sigma^2}{N_b^t}
      +\|\EB[\frac{1}{N_b^t}\sum_{i=1}^{N_b^t}\g_i~|~\w^t]\|^2\\
      =&\frac{\sigma^2}{N_b^t}
      +\frac{1}{(N_b^t)^2}\|\sum_{i=1}^{N_b^t}\EB[\g_i~|~\w^t]\|^2\\
      \overset{(b)}{\leq}&\frac{\sigma^2}{N_b^t}
      +\frac{1}{(N_b^t)^2}\cdot N_b^t\cdot\sum_{i=1}^{N_b^t}\|\EB[\g_i~|~\w^t]\|^2\\
      \overset{(c)}{\leq}&\frac{\sigma^2}{N_b^t} + D^2.
    \end{align*}
    Inequality (a) is derived based on Assumption~\ref{ass:lim_variance} and the fact that $\g_i$ is mutually uncorrelated.
    Inequality (b) is derived by the following process:
    \begin{align*}
      \|\sum_{i=1}^{N_b^t}\EB[\g_i~|~\w^t]\|^2
      =&\sum_{i=1}^{N_b^t}\|\EB[\g_i~|~\w^t]\|^2 
        + \sum_{1\leq i < i'\leq N_b^t} 2 \cdot \EB[\g_i~|~\w^t]^T \EB[\g_i'~|~\w^t]\\
      \leq& \sum_{i=1}^{N_b^t}\|\EB[\g_i~|~\w^t]\|^2 
        + \sum_{1\leq i < i'\leq N_b^t} (\|\EB[\g_i~|~\w^t]\|^2+\|\EB[\g_i'~|~\w^t]\|^2\\
      =& \sum_{i=1}^{N_b^t}\|\EB[\g_i~|~\w^t]\|^2 + (N_b^t-1)\cdot\sum_{i=1}^{N_b^t}\|\EB[\g_i~|~\w^t]\|^2 \\
      =& N_b^t\cdot\sum_{i=1}^{N_b^t}\|\EB[\g_i~|~\w^t]\|^2.
    \end{align*}
    Inequality (c) is derived based on Assumption~\ref{ass:bounded_gradient}.

    Because there are no more than $r$ Byzantine workers at iteration $t$,
    no more than $r$ buffers contain Byzantine gradient. Thus, the credible buffer index set $\HM^t$
    has at least $(B-r)$ elements.
    In case that $\HM^t$ has more than $(B-r)$ elements,
    we take the indices of the smallest $(B-q)$ elements in $\{h_{bj}\}_{b\in\HM^t}$
    to compose $\HM^t_j$,
    and we have $|\HM^t_j|=B-q$.

    Note that $Aggr(\cdot)$ is $q$-BR, and by definition we have:
    $$\min_{b \in\HM^t_j}\{h_{bj}\}\leq Aggr([\h_1,\ldots,\h_B])_j \leq\max_{b\in\HM^t_j}\{h_{bj}\}.$$
    Therefore,
    $$\sum_{j=1}^d\EB[Aggr([\h_1,\ldots,\h_B])_j^2|\w^t]
    \leq\sum_{j=1}^d\EB[\max_{b\in\HM^t_j}\{h_{bj}^2\}|\w^t].$$

    There are $(B-r)$ credible buffers,
    and we choose the smallest $(B-q)$ buffers to compose $\HM^t_j$.
    Therefore, for all $b\in\HM^t_j$, $h_{bj}$ is not larger than the $(q-r+1)$-th largest one in $\{h_{bj}\}_{b\in\HM^t}$.
    Let $N^{(t)}$ be the $(q+1)$-th smallest value in $\{N_b^t\}_{b\in[B]}$.
    Using Lemma~\ref{lemma:BOOSV}, we have:
    \begin{align*}
      \EB[\max_{b\in\HM^t_j}\{h_{bj}^2\}|\w^t]
      \leq &\EB[\max_{b\in\HM^t_j}\{\|\h_b\|^2\}|\w^t] \\
      \leq&\EB[\max_{b\in\HM^t_j}\{D^2+\frac{\sigma^2}{N_b^t}\}|\w^t]\\
      =&C_{B-r,q-r+1}\cdot(D^2+\frac{\sigma^2}{N^{(t)}}).
    \end{align*}
    Thus,
    \begin{align*}
        \EB[||\G^t||^2~|~\w^t]
        \ \leq\ \sum_{j=1}^d\EB[\max_{b\in\HM^t_j}\{h_{bj}^2\}|\w^t]
        \ \leq C_{B-r,q-r+1}d\cdot(D^2+\frac{\sigma^2}{N^{(t)}}).
    \end{align*}
    By Proposition~\ref{prop:estimation_cnk}, we have:
    \begin{equation*}
      \EB[||\G^t||^2~|~\w^t]\leq d\cdot\frac{(B-r)\sqrt{B-r+1}}{\sqrt{(B-q-1)(q-r+1)}}\cdot(D^2+\frac{\sigma^2}{N^{(t)}}).
    \end{equation*}
\end{proof}

\subsection{Proof of Lemma~\ref{lemma:expectation_bound}}
\begin{proof}
    \begin{align}
        &\EB[\G^t-\nabla F(\w^t)~|~\w^t] \nonumber\\
        =&\EB[Aggr([\h_1,\ldots,\h_B])-\nabla F(\w^t)~|~\w^t] \nonumber\\
        =&\EB[Aggr([\h_1-\nabla F(\w^t),\ldots,\h_B-\nabla F(\w^t)])~|~\w^t], \label{eq:lm_exp_1}
    \end{align}
    where the second equation is derived based on the Property (b) in the definition of $q$-BR.

    For each $b\in\HM^t$, $\h_b$ has stored $N_b^t$ gradients
    at iteration $t$: $\g_1,\ldots,\g_{N_b^t}$, and we have:
    \begin{align*}
        \h_b-\nabla F(\w^t)=\frac{1}{N_b^t}\sum_{k=1}^{N_b^t}\g_i-\nabla F(\w^t)
        =\frac{1}{N_b^t}\sum_{k=1}^{N_b^t}[\nabla f(\w^{t_k};z_{i_k})-\nabla F(\w^t)],
    \end{align*}
    where $0\leq t-t_k\leq\tau_{max}$, $\forall k=1,2,\ldots,N_b^t$.

    Taking expectation on both sides, we have:
    \begin{align*}
        &\EB[||\h_b-\nabla F(\w^t)||~|\w^{t}]\\
        =&\EB[||\frac{1}{N_b^t}\sum_{k=1}^{N_b^t}(\nabla f(\w^{t_k};z_{i_k})-\nabla F(\w^t))||~|\w^{t}]\\
        \leq&\frac{1}{N_b^t}\sum_{k=1}^{N_b^t}\EB[||\nabla f(\w^{t_k};z_{i_k})-\nabla F(\w^t)||~|\w^{t}]\\
        \overset{(a)}{\leq}&\frac{1}{N_b^t}\sum_{k=1}^{N_b^t}\{\EB[||\nabla F(\w^{t_k})-\nabla F(\w^t)||~|\w^{t}]\\
        &\qquad +\EB[||\nabla f(\w^{t_k};z_{i_k})-\EB[\nabla f(\w^{t_k};z_{i_k})]||~|\w^{t}] \\
        &\qquad +\EB[||\EB[\nabla f(\w^{t_k};z_{i_k})]-\nabla F(\w^{t_k})||~|\w^{t}]\},
    \end{align*}
    where (a) is derived based on Triangle Inequality.

    The first part:
    \begin{align*}
      &\EB[||\nabla F(\w^{t_k})-\nabla F(\w^t)||~|\w^{t}]\\
      \overset{(b)}{\leq}&L\cdot\EB[||\w^{t_k}-\w^t||~|\w^{t}]\\
      =&L\cdot\EB[||\sum_{t'=t_k}^{t-1}\G^{t'}||~|\w^{t}]\\
      \leq&\sum_{t'=t_k}^{t-1}L\cdot\EB[||\G^{t'}||~|\w^{t}]\\
      =&\sum_{t'=t_k}^{t-1}L\cdot\sqrt{\EB[||\G^{t'}||~|\w^{t}]^2}\\
      \leq&\sum_{t'=t_k}^{t-1}L\cdot\sqrt{\EB[||\G^{t'}||^2~|\w^{t}]}\\
      \overset{(c)}{\leq}&\sum_{t'=t_k}^{t-1}L\cdot\sqrt{C_{B-r,q-r+1}d\cdot(D^2+\sigma^2/N^{(t)})}\\
      \overset{(d)}{\leq}&\tau_{max}L\cdot\sqrt{C_{B-r,q-r+1}d\cdot(D^2+\sigma^2/N^{(t)})},
    \end{align*}
    where (b) is derived based on Assumption~\ref{ass:L_smooth},
    (c) is derived based on Lemma~\ref{lemma:variance_bound} and
    (d) is derived based on $t-t_k \leq \tau_{max}$.

    The second part:
    \begin{align*}
            & \EB[||\nabla f(\w^{t_k};z_{i_k})-\EB[\nabla f(\w^{t_k};z_{i_k})]||~|\w^{t}] \\
        =& \sqrt{\EB[||\nabla f(\w^{t_k};z_{i_k})-\EB[\nabla f(\w^{t_k};z_{i_k})]||~|\w^{t}]^2} \\
        \leq&\sqrt{\EB[||\nabla f(\w^{t_k};z_{i_k})-\EB[\nabla f(\w^{t_k};z_{i_k})]||^2~|\w^{t}]} \\
        \overset{(e)}{\leq}& \sigma,
    \end{align*}
    where (e) is derived based on Assumption~\ref{ass:lim_variance}.

    By Assumption~\ref{ass:limit_bias}, we have the following estimation for the third part:
    $$\EB[||\EB[\nabla f(\w^{t_k};z_{i_k})]-\nabla F(\w^{t_k})||~|\w^{t}]\leq \kappa.\nonumber$$
    Therefore,
    \begin{align}
        &\EB[||\h_b-\nabla F(\w^t)||~|\w^{t}]  \nonumber\\ 
        \leq&\frac{1}{N_b^t}\sum_{k=1}^{N_b^t}(\tau_{max}L \sqrt{C_{B-r,q-r+1}d\cdot(D^2+\sigma^2/N^{(t)})} + \sigma +\kappa)  \nonumber\\ 
        =&\tau_{max}L\sqrt{C_{B-r,q-r+1}d\cdot(D^2+\sigma^2/N^{(t)})} + \sigma +\kappa.  \label{ineq:xxx}
    \end{align}
    Similar to the proof of Lemma~\ref{lemma:variance_bound}, $\forall j\in[d]$, we have:
    \begin{align*}
        &\min_{b\in\HM^t_j}\{h_{bj}-\nabla F(\w^t)_j\}\\
        \leq&Aggr([\h_1-\nabla F(\w^t),\ldots,\h_B-\nabla F(\w^t)])_j \\
        \leq&\max_{b\in\HM^t_j}\{h_{bj}-\nabla F(\w^t)_j\},
    \end{align*}
    where $\HM^t_j$ is composed by the indices of the smallest $(B-q)$
    elements in $\{h_{bj}-\nabla F(\w^t)_j\}_{b\in\HM^t}$.
    Therefore,
    \begin{align}
        &||\EB[Aggr([\h_1-\nabla F(\w^t),\ldots,\h_B-\nabla F(\w^t)])~|~\w^t]|| \nonumber\\
        \leq&\sum_{j=1}^d||\EB[Aggr([\h_1-\nabla F(\w^t),\ldots,\h_B-\nabla F(\w^t)])_j~|~\w^t]|| \nonumber\\
        \leq&\sum_{j=1}^d\EB[||Aggr([\h_1-\nabla F(\w^t),\ldots,\h_B-\nabla F(\w^t)])_j||~|~\w^t] \nonumber\\
        \overset{(f)}{\leq}&\sum_{j=1}^d \EB[\max_{b\in\HM^t_j}||h_{bj}-\nabla F(\w^t)_j||~|~\w^t] \nonumber\\
        \overset{(g)}{\leq}&\sum_{j=1}^d C_{B-r,q-r+1}\EB[||h_{bj}-\nabla F(\w^t)_j||~|\w^{t}] \nonumber\\
        \leq&\sum_{j=1}^d C_{B-r,q-r+1}\EB[||\h_b-\nabla F(\w^t)||~|\w^{t}] \nonumber\\
        \overset{(h)}{\leq}&\sum_{j=1}^d C_{B-r,q-r+1}\cdot (\tau_{max}L\sqrt{C_{B-r,q-r+1}d\cdot(D^2+\sigma^2/N^{(t)})} + \sigma +\kappa) \nonumber\\
        =&C_{B-r,q-r+1}d\cdot (\tau_{max}L\sqrt{C_{B-r,q-r+1}d\cdot(D^2+\sigma^2/N^{(t)})} + \sigma +\kappa), \label{eq:lm_exp_2}
    \end{align}
    where (f) is derived based on definition of $q$-BR, (g) is derived based on Lemma~\ref{lemma:BOOSV}, and (h) is derived based on Inequality~(\ref{ineq:xxx}).


    Combining Equation~(\ref{eq:lm_exp_1}) and Inequality~(\ref{eq:lm_exp_2}), 
    we obtain:
    \begin{equation*}
      ||\EB[\G^t-\nabla F(\w^t)~|~\w^t]||
      \leq C_{B-r,q-r+1}d\cdot (\tau_{max}L\sqrt{C_{B-r,q-r+1}d\cdot(D^2+\sigma^2/N^{(t)})} + \sigma +\kappa) .
    \end{equation*}

    By Proposition~(\ref{prop:estimation_cnk}), we have:   
    \begin{align*}
      ||\EB[\G^t-\nabla F(\w^t)~|~\w^t]|| \leq &\frac{d(B-r)\sqrt{B-r+1}}{\sqrt{(B-q-1)(q-r+1)}}\\
      \cdot &(\tau_{max}L\sqrt{d\frac{(B-r)\sqrt{B-r+1}}{\sqrt{(B-q-1)(q-r+1)}}\cdot(D^2+\sigma^2/N^{(t)})} + \sigma +\kappa) .
    \end{align*}
\end{proof}

\subsection{Proof of Theorem~\ref{thm:main_theorem}}
\begin{proof}
    \begin{align*}
        \EB[F(\w^{t+1})~|~\w^t]
        =&\EB[F(\w^t-\eta\cdot\G^t)~|~\w^t]\\
        \overset{(a)}{\leq}&\EB[F(\w^t)-\eta\cdot\nabla F(\w^t)^T\G^t+\frac{L}{2}\eta^2||\G^t||^2~|~\w^t]\\
        =&F(\w^t)-\eta\cdot\EB[\nabla F(\w^t)^T\G^t~|~\w^t] + \frac{\eta^2 L}{2}\EB[||\G^t||^2~|~\w^t]\\
        =&F(\w^t)-\eta\cdot\nabla F(\w^t)^T\EB[\G^t~|~\w^t] + \frac{\eta^2 L}{2}\EB[||\G^t||^2~|~\w^t]\\
        =&F(\w^t)-\eta\cdot\nabla F(\w^t)^T\nabla F(\w^t)+ \frac{\eta^2 L}{2}\EB[||\G^t||^2~|~\w^t]\\
        &-\eta\cdot\nabla F(\w^t)^T \EB[\G^t-\nabla F(\w^t)~|~\w^t]\\
        \leq &F(\w^t)-\eta\cdot||\nabla F(\w^t)||^2+ \frac{\eta^2 L}{2}\EB[||\G^t||^2~|~\w^t]\\
        &+\eta\cdot||\nabla F(\w^t)||\cdot||\EB[\G^t-\nabla F(\w^t)~|~\w^t]||,
    \end{align*}
    where (a) is derived based on Assumption~\ref{ass:L_smooth}.

    Using Lemma~\ref{lemma:variance_bound} and Lemma~\ref{lemma:expectation_bound}, we have:
    \begin{align*}
        &\EB[F(\w^{t+1})~|~\w^t]\\
        \leq &F(\w^t)-\eta\cdot||\nabla F(\w^t)||^2+ \frac{\eta^2 L}{2}C_{B-r,q-r+1}d \cdot (D^2+\sigma^2/N^{(t)}) \\
        &+\eta \cdot C_{B-r,q-r+1}d\cdot (\tau_{max}L\sqrt{C_{B-r,q-r+1}d\cdot(D^2+\sigma^2/N^{(t)})} + \sigma +\kappa) \cdot||\nabla F(\w^t)||.
    \end{align*}

    Also, by Assumption~\ref{ass:bounded_gradient}, 
        $||\nabla F(\w^t)||\leq D$.

    Taking total expectation and combining $||\nabla F(\w^t)||\leq D$, we have:
    \begin{align*}
        \EB[F(\w^{t+1})]
        \leq &\EB[F(\w^t)]-\eta\cdot\EB[||\nabla F(\w^t)||^2]+ \frac{\eta^2 L}{2}C_{B-r,q-r+1}d \cdot (D^2+\sigma^2/N^{(t)})\\
        &+\eta \cdot C_{B-r,q-r+1}D d(\tau_{max}L\sqrt{C_{B-r,q-r+1}d\cdot(D^2+\sigma^2/N^{(t)})} + \sigma +\kappa).
    \end{align*}
    Let $\tilde{D}=\frac{1}{T}\sum_{t=0}^{T-1}\sqrt{D^2 +\sigma^2/N^{(t)}}$.
    By telescoping, we have:
    \begin{align*}
        \eta\cdot\sum_{t=0}^{T-1} \EB[||\nabla F(\w^t)||^2]
        \leq&\{F(\w^0)-\EB[F(\w^T)]\}\\
        &+\eta^2 T \cdot \frac{ L}{2}C_{B-r,q-r+1}d \cdot \frac{1}{T}\sum_{t=0}^{T-1}(D^2 +\sigma^2/N^{(t)})\\
        &+\eta T \cdot C_{B-r,q-r+1}D d(\tau_{max}L\tilde{D}\sqrt{C_{B-r,q-r+1}d} + \sigma +\kappa).
    \end{align*}
    Note that $\EB[F(\w^T)]\geq F^*$, and let $\eta=O\left(\frac{1}{L\sqrt{T}}\right)$:
    \begin{align*}
        \frac{\sum_{t=0}^{T-1} \EB[||\nabla F(\w^t)||^2]}{T}
        \leq&O\left(\frac{L[F(\w^0)-F^*]}{\sqrt{T}}\right)
        +O\left(\frac{C_{B-r,q-r+1}\tilde{D}d}{\sqrt{T}}\right)\\
        &+O\left(C_{B-r,q-r+1}D d \cdot (\tau_{max}L\tilde{D}\sqrt{C_{B-r,q-r+1}d}+\sigma+\kappa)\right).
    \end{align*}
    When $q=r$ and $B=O(r)$, we have $C_{B-r,q-r+1}\leq\frac{(B-r)\sqrt{B-r+1}}{\sqrt{(B-q-1)(q-r+1)}}=O\left(\frac{r}{(q-r+1)^\frac{1}{2}}\right)$. Thus,

    \begin{align*}
      \frac{\sum_{t=0}^{T-1} \EB[||\nabla F(\w^t)||^2]}{T}
      \leq&O\left(\frac{L[F(\w^0)-F^*]}{T^\frac{1}{2}}
      \right)
      +O\left(\frac{rd\tilde{D}}{T^\frac{1}{2}(q-r+1)^\frac{1}{2}}\right)
      +O\left(\frac{rD d\sigma}{(q-r+1)^\frac{1}{2}}\right)\\
      &\qquad \qquad \qquad \qquad +O\left(\frac{rD d\kappa}{(q-r+1)^\frac{1}{2}}\right)
      +O\left(\frac{r^\frac{3}{2}LD\tilde{D}d^\frac{3}{2}\tau_{max}}{(q-r+1)^\frac{3}{4}}\right).
    \end{align*}
\end{proof}

\subsection{Proof of Theorem~\ref{thm:general_proper}}
\begin{proof}
Let $\h_b'$ be the value of the $b$-th buffer, 
if all received loyal gradients were computed based on $\w^t$.
Note $\G^t=Aggr(\h_1,\ldots,\h_B)$. 
\begin{align}
   &\EB[F(\w^{t+1})~|~\w^t]\nonumber\\
   =&\EB[F(\w^t-\eta\cdot\G^t)~|~\w^t]\nonumber\\
   \overset{(a)}{\leq}&\EB[F(\w^t)-\eta\cdot\nabla F(\w^t)^T\G^t+\frac{L}{2}\eta^2||\G^t||^2~|~\w^t]\nonumber\\
   =&F(\w^t)-\eta\cdot\EB[\nabla F(\w^t)^T\G^t~|~\w^t] + \frac{\eta^2 L}{2}\EB[||\G^t||^2~|~\w^t],\label{ineq:key_taylor}
\end{align}
where (a) is derived based on Assumption~\ref{ass:L_smooth}.

Firstly, we estimate the value of $\EB[\nabla F(\w^t)^T\G^t~|~\w^t]$.

Since there are at most $r$ Byzantine workers, at most $r$ buffers may contain Byzantine gradients.
Without loss of generality, suppose only the first $r$ buffers may contain Byzantine gradients.

Let $\G_{syn}^t=Aggr(\h_1,\ldots,\h_r,\h_{r+1}',\ldots,\h_B')$,
where $\h_1,\ldots,\h_r$ may contain Byzantine gradients and be arbitrary value, 
and $\h_{r+1}',\ldots,\h_B'$ each stores loyal gradients computed based on $\w^t$. 
Thus,
\begin{equation}\label{ineq:exp_gsyn}
   \EB[\nabla F(\w^t)^T\G_{syn}^t~|~\w^t] \geq \|\nabla F(\w^t)\|^2 - A_1,
\end{equation}
\begin{equation}\label{ineq:var_gsyn}
   \EB[\|\G_{syn}^t\|^2~|~\w^t]\leq (A_2)^2.
\end{equation} 
Let $\alpha = 2\eta^2L^2\tau_{max}^2(B-r)<1$.

We claim that
\begin{equation*}
   \EB[\|\G^t-\G_{syn}^t\|^2~|~\w^t]\leq (\frac{1}{2}\alpha^{t+1}+\frac{\alpha}{1-\alpha})\cdot (A_2)^2,
\end{equation*}
and
\begin{equation*}
   \EB[\|\G^t\|^2~|~\w^t]\leq (\alpha^{t+1}+\frac{2}{1-\alpha})\cdot (A_2)^2.
\end{equation*} 
Now we prove it by induction on $t$.

Step 1. When $t=0$, all gradients are computed according to $\w^0$, and we have $\G^0=\G_{syn}^0$. Thus,
\begin{equation*}
  \EB[\|\G^0-\G_{syn}^0\|^2~|~\w^0]=0\leq (\frac{1}{2}\alpha^1+\frac{\alpha}{1-\alpha})\cdot (A_2)^2,
\end{equation*}
\begin{equation*}
   \EB[\|\G^0\|^2~|~\w^0]=\EB[\|\G_{syn}^0\|^2~|~\w^0]\leq (A_2)^2\leq (\alpha^1+\frac{2}{1-\alpha})\cdot(A_2)^2.
\end{equation*}

Step 2. If
\begin{equation*}
   \EB[\|\G^{t'}-\G_{syn}^{t'}\|^2~|~\w^{t'}]\leq (\frac{1}{2}\alpha^{t'+1}+\frac{\alpha}{1-\alpha})\cdot (A_2)^2,
\end{equation*} 
\begin{equation*}
  \EB[\|\G^{t'}\|^2~|~\w^{t'}]\leq (\alpha^{t'+1}+\frac{2}{1-\alpha})\cdot (A_2)^2,
\end{equation*} 
holds for all $t'=0,1,\ldots,t-1$~(induction hypothesis), then:
\begin{align}
   &\EB[\|\G^t-\G_{syn}^t\|^2~|~\w^t]\nonumber\\
   =&\EB[\|Aggr(\h_1,\ldots,\h_r,\h_{r+1},\ldots,\h_B)-Aggr(\h_1,\ldots,\h_r,\h_{r+1}',\ldots,\h_B')\|^2~|~\w^t]\nonumber\\
   \overset{(b)}{\leq}&\EB[\sum_{b=r+1}^B\|\h_b-\h_b'\|^2~|~\w^t] \nonumber\\
   =&\sum_{b=r+1}^B\EB[\| \frac{1}{N_b^t}\sum_{i=1}^{N_b^t}(\nabla f(\w^{t_k};z_{i_k})-\nabla f(\w^t;z_{i_k}))\|^2~|~\w^t]\nonumber\\
   \overset{(c)}{\leq}&\sum_{b=r+1}^B\EB[\frac{1}{N_b^t}\sum_{i=1}^{N_b^t}\|\nabla f(\w^{t_k};z_{i_k})-\nabla f(\w^t;z_{i_k})\|^2~|~\w^t]\nonumber\\
   \overset{(d)}{\leq}&\sum_{b=r+1}^B\EB[\frac{1}{N_b^t}\sum_{i=1}^{N_b^t}L^2\|\w^{t_k}-\w^t\|^2~|~\w^t]\nonumber\\
   =&\sum_{b=r+1}^B\frac{L^2}{N_b^t}\sum_{i=1}^{N_b^t}\EB[\|\w^{t_k}-\w^t\|^2~|~\w^t]\nonumber\\
   =&\sum_{b=r+1}^B\frac{L^2}{N_b^t}\sum_{i=1}^{N_b^t}\EB[\|\sum_{t'=t_k}^{t-1}\eta\cdot\G^{t'}\|^2~|~\w^t]\nonumber\\
   \overset{(e)}{\leq}&\sum_{b=r+1}^B\frac{\eta^2L^2}{N_b^t}\sum_{i=1}^{N_b^t}\EB[(t-t_k)\sum_{t'=t_k}^{t-1}\|\G^{t'}\|^2~|~\w^t]\nonumber\\
   \overset{(f)}{\leq}&\sum_{b=r+1}^B\frac{\eta^2L^2}{N_b^t}\sum_{i=1}^{N_b^t}[(t-t_k)\sum_{t'=t_k}^{t-1} (\alpha^{t'+1}+\frac{2}{1-\alpha})\cdot (A_2)^2]\nonumber\\
   \leq&\sum_{b=r+1}^B\frac{\eta^2L^2}{N_b^t}\sum_{i=1}^{N_b^t}[(t-t_k)\sum_{t'=t_k}^{t-1} (\alpha^{t}+\frac{2}{1-\alpha})\cdot (A_2)^2]\nonumber\\
   \overset{(g)}{\leq}&\sum_{b=r+1}^B(\eta^2L^2\tau_{max}^2)\cdot (\alpha^{t}+\frac{2}{1-\alpha})\cdot (A_2)^2\nonumber\\
   =&(\eta^2L^2(B-r)\tau_{max}^2)\cdot (\alpha^{t}+\frac{2}{1-\alpha})\cdot (A_2)^2\nonumber\\
   \overset{(h)}{\leq}&\frac{1}{2}\alpha\cdot (\alpha^{t}+\frac{2}{1-\alpha})\cdot (A_2)^2\nonumber\\
   =&(\frac{1}{2}\alpha^{t+1}+\frac{\alpha}{1-\alpha})\cdot (A_2)^2, \label{ineq:induction_1}
\end{align}
where (b) is derived based on the definition of stable aggregation function,
(c) is derived based on Cauchy's Inequality,
(d) is derived based on Assumption~\ref{ass:L_smooth},
(e) is also derived based on Cauchy's Inequality,
(f) is derived based on induction hypothesis,
(g) is derived based on that $t-t_k\leq \tau_{max}$, and 
(h) is derived based on that $\alpha = 2\eta^2L^2\tau_{max}^2(B-r)$.

Therefore,
\begin{align}
   \EB[\|\G^t\|^2~|~\w^t]
   =&\EB[||\G_{syn}^t+(\G^t-\G_{syn}^t)||^2~|~\w^t]\nonumber\\
   \overset{(i)}{\leq}& 2\cdot \EB[\|\G_{syn}^t\|^2~|~\w^t] + 2\cdot \EB[||\G^t-\G_{syn}^t||^2~|~\w^t]\nonumber\\
   \overset{(j)}{\leq}& 2\cdot (A_2)^2 + 2\cdot \EB[||\G^t-\G_{syn}^t||^2~|~\w^t]\nonumber\\
   \overset{(k)}{\leq}& 2\cdot (A_2)^2 +2\cdot(\frac{1}{2}\alpha^{t+1}+\frac{\alpha}{1-\alpha})\cdot (A_2)^2\nonumber\\
   =& (\alpha^{t+1}+\frac{2}{1-\alpha})\cdot (A_2)^2,\label{ineq:induction_2}
\end{align}
where (i) is derived based on that $\|\x+\y\|^2\leq 2\|\x\|^2+2\|\y\|^2,\ \forall \x,\y\in\RB^d$,
(j) is derived by the definition of $(A_1,A_2)$-effective aggregation function, and
(k) is derived based on Inequality~(\ref{ineq:induction_1}).

By Inequality~(\ref{ineq:induction_1}) and (\ref{ineq:induction_2}), the claimed property also holds for $t'=t$.

In conclusion, for all $t=0,1,\ldots,T-1$, we have:
\begin{equation}\label{ineq:key_lemma1_1}
  \EB[\|\G^t-\G_{syn}^t\|^2~|~\w^t]\leq (\frac{1}{2}\alpha^{t+1}+\frac{\alpha}{1-\alpha})\cdot (A_2)^2,
\end{equation}
and
\begin{equation}\label{ineq:key_lemma1}
  \EB[\|\G^t\|^2~|~\w^t]\leq (\alpha^{t+1}+\frac{2}{1-\alpha})\cdot (A_2)^2.
\end{equation} 

Also, $\EB[\|\G^t\|~|~\w^t]^2 + Var[\|\G^t\|~|~\w^t]= \EB[\|\G^t\|^2~|~\w^t].$
Therefore,
\begin{equation}
   \EB[\|\G^t\|~|~\w^t]=\sqrt{\EB[\|\G^t\|~|~\w^t]^2}\leq \sqrt{\alpha^{t+1}+\frac{2}{1-\alpha}}\cdot A_2.
\end{equation}

We have:
\begin{align}
   &\eta\cdot\EB[\nabla F(\w^t)^T\G^t~|~\w^t] \nonumber\\
   =&\eta\cdot\EB[\nabla F(\w^t)^T\G_{syn}^t~|~\w^t] 
    + \eta\cdot\EB[\nabla F(\w^t)^T(\G^t-\G_{syn}^t)~|~\w^t] \nonumber\\
    \overset{(l)}{\geq} & \eta\cdot(\|\nabla F(\w^t)\|^2 - A_1) 
    + \eta\cdot\EB[\nabla F(\w^t)^T(\G^t-\G_{syn}^t)~|~\w^t] \nonumber\\
    \geq & \eta\cdot\|\nabla F(\w^t)\|^2 - \eta\cdot A_1
    - \eta\cdot\|\nabla F(\w^t)\|\cdot\|\EB[(\G^t-\G_{syn}^t)~|~\w^t]\| \nonumber\\
    \overset{(m)}{\geq} & \eta\cdot\|\nabla F(\w^t)\|^2 - \eta\cdot A_1
    - \eta\cdot D\cdot\|\EB[(\G^t-\G_{syn}^t)~|~\w^t]\| \nonumber\\
    \overset{(n)}{\geq} & \eta\cdot\|\nabla F(\w^t)\|^2 - \eta\cdot A_1
    - \eta\cdot D \cdot \sqrt{\frac{1}{2}\alpha^{t+1}+\frac{\alpha}{1-\alpha}}\cdot A_2,\label{ineq:key_lemma2}
\end{align}
where (l) is derived based on the definition of $(A_1,A_2)$-effective aggregation function,
(m) is derived by Assumption~\ref{ass:bounded_gradient}, and
(n) is derived based on Inequality~(\ref{ineq:key_lemma1_1}).

Combining Inequalities~(\ref{ineq:key_taylor}), (\ref{ineq:key_lemma1}), (\ref{ineq:key_lemma2}) and taking total expectation, we have:
\begin{align*}
   \EB[F(\w^{t+1})]
   \leq &\EB[F(\w^t)]-\eta\cdot\EB[\|\nabla F(\w^t)\|^2]\\
   &+\eta\cdot A_1+\eta\cdot D\sqrt{\frac{1}{2}\alpha^{t+1}+\frac{\alpha}{1-\alpha}}\cdot A_2
   +\frac{1}{2} \eta^2 L (\alpha^{t+1}+\frac{2}{1-\alpha})\cdot (A_2)^2.
\end{align*}

By telescoping, we have:
\begin{align*}
   \eta\cdot\sum_{t=0}^{T-1}\EB[\|\nabla F(\w^t)\|^2]
   \leq & \{F(\w^0)-\EB[F(\w^T)]\}+\frac{1}{2} \eta^2 T L (\alpha+\frac{2}{1-\alpha})\cdot (A_2)^2\\
   &+\eta T A_1+\eta T  D\cdot\sqrt{\frac{1}{2}\alpha+\frac{\alpha}{1-\alpha}}\cdot A_2.
\end{align*}
Divide both sides of the equation by $\eta T$, and let $\eta=O(\frac{1}{\sqrt{LT}})$:
\begin{align*}
  &\frac{\sum_{t=0}^{T-1}\EB[\|\nabla F(\w^t)\|^2]}{T}\\
  \leq & \frac{\{F(\w^0)-\EB[F(\w^T)]\}}{\eta T}+\frac{1}{2} \eta L (\alpha+\frac{2}{1-\alpha})\cdot (A_2)^2
  +A_1+D\cdot\sqrt{\frac{1}{2}\alpha+\frac{\alpha}{1-\alpha}}\cdot A_2\\
  \leq & \frac{\sqrt{L}[F(\w^0)-F^*]}{\sqrt{T}}+\frac{\sqrt{L}(\frac{1}{2}\alpha+\frac{1}{1-\alpha})\cdot (A_2)^2}{\sqrt{T}} 
  +A_1+\alpha^\frac{1}{2}[\frac{3-\alpha}{2(1-\alpha)}]^\frac{1}{2}\cdot D A_2.
\end{align*}
Note that $\alpha = 2\eta^2L^2\tau_{max}^2(B-r)=O\left( \frac{L\tau_{max}^2(B-r)}{T}\right)$, finally we have:
\begin{align*}
 \frac{\sum_{t=0}^{T-1}\EB[\|\nabla F(\w^t)\|^2]}{T}
 \leq & O\left(\frac{\sqrt{L}\cdot[F(\w^0)-F^*]}{\sqrt{T}}\right)+O\left(\frac{\sqrt{L} (A_2)^2(1+\alpha)}{\sqrt{T}}\right) \\
 &+O\left(\alpha^\frac{1}{2} D A_2\right)+A_1\\
 =& O\left(\frac{L^\frac{1}{2}[F(\w^0)-F^*]}{T^\frac{1}{2}}\right)+O\left(\frac{L^\frac{1}{2}\tau_{max}(B-r)^\frac{1}{2}DA_2}{T^\frac{1}{2}}\right)\\
 &+O\left(\frac{L^\frac{1}{2}(A_2)^2}{T^\frac{1}{2}}\right)+O\left(\frac{L^\frac{5}{2}(A_2)^2\tau_{max}^2(B-r)}{T^\frac{3}{2}}\right)+A_1.
\end{align*}
Specailly, when $B=O(r)$, we have: 
\begin{align*}
\frac{\sum_{t=0}^{T-1}\EB[\|\nabla F(\w^t)\|^2]}{T}
\leq & O\left(\frac{L^\frac{1}{2}[F(\w^0)-F^*]}{T^\frac{1}{2}}\right)+O\left(\frac{L^\frac{1}{2}\tau_{max}DA_2r^\frac{1}{2}}{T^\frac{1}{2}}\right)\\
&+O\left(\frac{L^\frac{1}{2}(A_2)^2}{T^\frac{1}{2}}\right)+O\left(\frac{L^\frac{5}{2}(A_2)^2\tau_{max}^2 r}{T^\frac{3}{2}}\right)+A_1.
\end{align*}
\end{proof}

\subsection{Proof of Theorem~\ref{thm:general_proper_m}}

\begin{proof}
 The proof of this theorem is similar to that of Theorem~\ref{thm:general_proper}. The main differences are the choices of the values $\alpha$~(in Theorem~\ref{thm:general_proper}) and $\tilde{\alpha}$~(here in Theorem~\ref{thm:general_proper_m}). For more readability, we still present the detailed proof processes here.

 Let $\h_b'$ be the value of the $b$-th buffer, 
 if all received loyal gradients were computed based on $\w^t$.
 Note $\G^t=Aggr(\h_1,\ldots,\h_B)$. 
 \begin{align}
    &\EB[F(\w^{t+1})~|~\w^t]\nonumber\\
    =&\EB[F(\w^t-\eta\cdot\G^t)~|~\w^t]\nonumber\\
    \overset{(a)}{\leq}&\EB[F(\w^t)-\eta\cdot\nabla F(\w^t)^T\G^t+\frac{L}{2}\eta^2||\G^t||^2~|~\w^t]\nonumber\\
    =&F(\w^t)-\eta\cdot\EB[\nabla F(\w^t)^T\G^t~|~\w^t] + \frac{\eta^2 L}{2}\EB[||\G^t||^2~|~\w^t],\label{ineq:key_taylor_m}
 \end{align}
 where (a) is derived based on Assumption~\ref{ass:L_smooth}.
 
 Firstly, we estimate the value of $\EB[\nabla F(\w^t)^T\G^t~|~\w^t]$.
 
 Since there are at most $r$ Byzantine workers, at most $r$ buffers may contain Byzantine gradients.
 Without loss of generality, suppose only the first $r$ buffers may contain Byzantine gradients.
 
 Let $\G_{syn}^t=Aggr(\h_1,\ldots,\h_r,\h_{r+1}',\ldots,\h_B')$,
 where $\h_1,\ldots,\h_r$ may contain Byzantine gradients and be arbitrary value, 
 and $\h_{r+1}',\ldots,\h_B'$ each stores loyal gradients computed based on $\w^t$. 
 Thus,
 \begin{equation}\label{ineq:exp_gsyn_m}
    \EB[\nabla F(\w^t)^T\G_{syn}^t~|~\w^t] \geq \|\nabla F(\w^t)\|^2 - A_1,
 \end{equation}
 \begin{equation}\label{ineq:var_gsyn_m}
    \EB[\|\G_{syn}^t\|^2~|~\w^t]\leq (A_2)^2.
 \end{equation} 
 Let $\tilde{\alpha} = 2\eta^2L^2\tau_{max}^2(1-\mu)^2(B-r)<1$.
 
 We claim that
 \begin{equation*}
    \EB[\|\G^t-\G_{syn}^t\|^2~|~\w^t]\leq (\frac{1}{2}\tilde{\alpha}^{t+1}+\frac{\tilde{\alpha}}{1-\tilde{\alpha}})\cdot (A_2)^2,
 \end{equation*}
 and
 \begin{equation*}
    \EB[\|\G^t\|^2~|~\w^t]\leq (\tilde{\alpha}^{t+1}+\frac{2}{1-\tilde{\alpha}})\cdot (A_2)^2.
 \end{equation*} 
 Now we prove it by induction on $t$.
 
 Step 1. When $t=0$, all gradients are computed according to $\w^0$, and we have $\G^0=\G_{syn}^0$. Thus,
 \begin{equation*}
   \EB[\|\G^0-\G_{syn}^0\|^2~|~\w^0]=0\leq (\frac{1}{2}\tilde{\alpha}^1+\frac{\tilde{\alpha}}{1-\tilde{\alpha}})\cdot (A_2)^2,
 \end{equation*}
 \begin{equation*}
    \EB[\|\G^0\|^2~|~\w^0]=\EB[\|\G_{syn}^0\|^2~|~\w^0]\leq (A_2)^2\leq (\tilde{\alpha}^1+\frac{2}{1-\tilde{\alpha}})\cdot(A_2)^2.
 \end{equation*}
 
 Step 2. If
 \begin{equation*}
    \EB[\|\G^{t'}-\G_{syn}^{t'}\|^2~|~\w^{t'}]\leq (\frac{1}{2}\tilde{\alpha}^{t'+1}+\frac{\tilde{\alpha}}{1-\tilde{\alpha}})\cdot (A_2)^2,
 \end{equation*} 
 \begin{equation*}
   \EB[\|\G^{t'}\|^2~|~\w^{t'}]\leq (\tilde{\alpha}^{t'+1}+\frac{2}{1-\tilde{\alpha}})\cdot (A_2)^2,
 \end{equation*} 
 holds for all $t'=0,1,\ldots,t-1$~(induction hypothesis), then:
 \begin{align}
   &\EB[\|\G^t-\G_{syn}^t\|^2~|~\w^t]\nonumber\\
   =&\EB[\|Aggr(\h_1,\ldots,\h_r,\h_{r+1},\ldots,\h_B)-Aggr(\h_1,\ldots,\h_r,\h_{r+1}',\ldots,\h_B')\|^2~|~\w^t]\nonumber\\
   \overset{(b)}{\leq}&\EB[\sum_{b=r+1}^B\|\h_b-\h_b'\|^2~|~\w^t] \nonumber\\
   \overset{(c)}{=}&\sum_{b=r+1}^B\EB[\| \frac{1}{N_b^t}\sum_{i=1}^{N_b^t}(1-\mu)(\nabla f(\w^{t_k};z_{i_k})-\nabla f(\w^t;z_{i_k}))\|^2~|~\w^t]\nonumber\\
   \overset{(d)}{\leq}&\sum_{b=r+1}^B\EB[\frac{1}{N_b^t}\sum_{i=1}^{N_b^t}\|(1-\mu)(\nabla f(\w^{t_k};z_{i_k})-\nabla f(\w^t;z_{i_k}))\|^2~|~\w^t]\nonumber\\
   \overset{(e)}{\leq}&\sum_{b=r+1}^B\EB[\frac{1}{N_b^t}\sum_{i=1}^{N_b^t}(1-\mu)^2 L^2\|\w^{t_k}-\w^t\|^2~|~\w^t]\nonumber\\
   =&\sum_{b=r+1}^B\frac{L^2(1-\mu)^2}{N_b^t}\sum_{i=1}^{N_b^t}\EB[\|\w^{t_k}-\w^t\|^2~|~\w^t]\nonumber\\
   =&\sum_{b=r+1}^B\frac{L^2(1-\mu)^2}{N_b^t}\sum_{i=1}^{N_b^t}\EB[\|\sum_{t'=t_k}^{t-1}\eta\cdot\G^{t'}\|^2~|~\w^t]\nonumber\\
   \overset{(f)}{\leq}&\sum_{b=r+1}^B\frac{\eta^2L^2(1-\mu)^2}{N_b^t}\sum_{i=1}^{N_b^t}\EB[(t-t_k)\sum_{t'=t_k}^{t-1}\|\G^{t'}\|^2~|~\w^t]\nonumber\\
   \overset{(g)}{\leq}&\sum_{b=r+1}^B\frac{\eta^2L^2(1-\mu)^2}{N_b^t}\sum_{i=1}^{N_b^t}[(t-t_k)\sum_{t'=t_k}^{t-1} (\tilde{\alpha}^{t'+1}+\frac{2}{1-\tilde{\alpha}})\cdot (A_2)^2]\nonumber\\
   \leq&\sum_{b=r+1}^B\frac{\eta^2L^2(1-\mu)^2}{N_b^t}\sum_{i=1}^{N_b^t}[(t-t_k)\sum_{t'=t_k}^{t-1} (\tilde{\alpha}^{t}+\frac{2}{1-\tilde{\alpha}})\cdot (A_2)^2]\nonumber\\
   \overset{(h)}{\leq}&\sum_{b=r+1}^B(\eta^2L^2(1-\mu)^2\tau_{max}^2)\cdot (\tilde{\alpha}^{t}+\frac{2}{1-\tilde{\alpha}})\cdot (A_2)^2\nonumber\\
   =&((B-r)\eta^2L^2(1-\mu)^2\tau_{max}^2)\cdot (\tilde{\alpha}^{t}+\frac{2}{1-\tilde{\alpha}})\cdot (A_2)^2\nonumber\\
   \overset{(i)}{\leq}&\frac{1}{2}\tilde{\alpha}\cdot (\tilde{\alpha}^{t}+\frac{2}{1-\tilde{\alpha}})\cdot (A_2)^2\nonumber\\
    =&(\frac{1}{2}\tilde{\alpha}^{t+1}+\frac{\tilde{\alpha}}{1-\tilde{\alpha}})\cdot (A_2)^2, \label{ineq:induction_1_m}
 \end{align}
 where (b) is derived based on the definition of stable aggregation function,
 (c) is derived based on the worker momentum updating formula $\u\leftarrow \mu\cdot\u+(1-\mu)\cdot \nabla f(\w;z_i)$,
 (d) is derived based on Cauchy's Inequality,
 (e) is derived based on Assumption~\ref{ass:L_smooth},
 (f) is also derived based on Cauchy's Inequality,
 (g) is derived based on induction hypothesis,
 (h) is derived based on that $t-t_k\leq \tau_{max}$, and 
 (i) is derived based on that $\tilde{\alpha} = 2\eta^2L^2\tau_{max}^2(B-r)$.
 
 Therefore,
 \begin{align}
    \EB[\|\G^t\|^2~|~\w^t]
    =&\EB[||\G_{syn}^t+(\G^t-\G_{syn}^t)||^2~|~\w^t]\nonumber\\
    \overset{(i)}{\leq}& 2\cdot \EB[\|\G_{syn}^t\|^2~|~\w^t] + 2\cdot \EB[||\G^t-\G_{syn}^t||^2~|~\w^t]\nonumber\\
    \overset{(j)}{\leq}& 2\cdot (A_2)^2 + 2\cdot \EB[||\G^t-\G_{syn}^t||^2~|~\w^t]\nonumber\\
    \overset{(k)}{\leq}& 2\cdot (A_2)^2 +2\cdot(\frac{1}{2}\tilde{\alpha}^{t+1}+\frac{\tilde{\alpha}}{1-\tilde{\alpha}})\cdot (A_2)^2\nonumber\\
    =& (\tilde{\alpha}^{t+1}+\frac{2}{1-\tilde{\alpha}})\cdot (A_2)^2,\label{ineq:induction_2_m}
 \end{align}
 where (i) is derived based on that $\|\x+\y\|^2\leq 2\|\x\|^2+2\|\y\|^2,\ \forall \x,\y\in\RB^d$,
 (j) is derived by the definition of $(A_1,A_2)$-effective aggregation function, and
 (k) is derived based on Inequality~(\ref{ineq:induction_1_m}).
 
 By Inequality~(\ref{ineq:induction_1_m}) and (\ref{ineq:induction_2_m}), the claimed property also holds for $t'=t$.
 
 In conclusion, for all $t=0,1,\ldots,T-1$, we have:
 \begin{equation}\label{ineq:key_lemma1_1_m}
   \EB[\|\G^t-\G_{syn}^t\|^2~|~\w^t]\leq (\frac{1}{2}\tilde{\alpha}^{t+1}+\frac{\tilde{\alpha}}{1-\tilde{\alpha}})\cdot (A_2)^2,
 \end{equation}
 and
 \begin{equation}\label{ineq:key_lemma1_m}
   \EB[\|\G^t\|^2~|~\w^t]\leq (\tilde{\alpha}^{t+1}+\frac{2}{1-\tilde{\alpha}})\cdot (A_2)^2.
 \end{equation} 
 
 Also, $\EB[\|\G^t\|~|~\w^t]^2 + Var[\|\G^t\|~|~\w^t]= \EB[\|\G^t\|^2~|~\w^t].$
 Therefore,
 \begin{equation}
    \EB[\|\G^t\|~|~\w^t]=\sqrt{\EB[\|\G^t\|~|~\w^t]^2}\leq \sqrt{\tilde{\alpha}^{t+1}+\frac{2}{1-\tilde{\alpha}}}\cdot A_2.
 \end{equation}
 
 We have:
 \begin{align}
    &\eta\cdot\EB[\nabla F(\w^t)^T\G^t~|~\w^t] \nonumber\\
    =&\eta\cdot\EB[\nabla F(\w^t)^T\G_{syn}^t~|~\w^t] 
     + \eta\cdot\EB[\nabla F(\w^t)^T(\G^t-\G_{syn}^t)~|~\w^t] \nonumber\\
     \overset{(l)}{\geq} & \eta\cdot(\|\nabla F(\w^t)\|^2 - A_1) 
     + \eta\cdot\EB[\nabla F(\w^t)^T(\G^t-\G_{syn}^t)~|~\w^t] \nonumber\\
     \geq & \eta\cdot\|\nabla F(\w^t)\|^2 - \eta\cdot A_1
     - \eta\cdot\|\nabla F(\w^t)\|\cdot\|\EB[(\G^t-\G_{syn}^t)~|~\w^t]\| \nonumber\\
     \overset{(m)}{\geq} & \eta\cdot\|\nabla F(\w^t)\|^2 - \eta\cdot A_1
     - \eta\cdot D\cdot\|\EB[(\G^t-\G_{syn}^t)~|~\w^t]\| \nonumber\\
     \overset{(n)}{\geq} & \eta\cdot\|\nabla F(\w^t)\|^2 - \eta\cdot A_1
     - \eta\cdot D \cdot \sqrt{\frac{1}{2}\tilde{\alpha}^{t+1}+\frac{\tilde{\alpha}}{1-\tilde{\alpha}}}\cdot A_2,\label{ineq:key_lemma2_m}
 \end{align}
 where (l) is derived based on the definition of $(A_1,A_2)$-effective aggregation function,
 (m) is derived by Assumption~\ref{ass:bounded_gradient}, and
 (n) is derived based on Inequality~(\ref{ineq:key_lemma1_1_m}).
 
 Combining Inequalities~(\ref{ineq:key_taylor_m}), (\ref{ineq:key_lemma1_m}), (\ref{ineq:key_lemma2_m}) and taking total expectation, we have:
 \begin{align*}
    \EB[F(\w^{t+1})]
    \leq &\EB[F(\w^t)]-\eta\cdot\EB[\|\nabla F(\w^t)\|^2]\\
    &+\eta\cdot A_1+\eta\cdot D\sqrt{\frac{1}{2}\tilde{\alpha}^{t+1}+\frac{\tilde{\alpha}}{1-\tilde{\alpha}}}\cdot A_2
    +\frac{1}{2} \eta^2 L (\tilde{\alpha}^{t+1}+\frac{2}{1-\tilde{\alpha}})\cdot (A_2)^2.
 \end{align*}
 
 By telescoping, we have:
 \begin{align*}
    \eta\cdot\sum_{t=0}^{T-1}\EB[\|\nabla F(\w^t)\|^2]
    \leq & \{F(\w^0)-\EB[F(\w^T)]\}+\frac{1}{2} \eta^2 T L (\tilde{\alpha}+\frac{2}{1-\tilde{\alpha}})\cdot (A_2)^2\\
    &+\eta T A_1+\eta T  D\cdot\sqrt{\frac{1}{2}\tilde{\alpha}+\frac{\tilde{\alpha}}{1-\tilde{\alpha}}}\cdot A_2.
 \end{align*}
 Divide both sides of the equation by $\eta T$, and let $\eta=O(\frac{1}{\sqrt{LT}})$:
 \begin{align*}
   &\frac{\sum_{t=0}^{T-1}\EB[\|\nabla F(\w^t)\|^2]}{T}\\
   \leq & \frac{\{F(\w^0)-\EB[F(\w^T)]\}}{\eta T}+\frac{1}{2} \eta L (\tilde{\alpha}+\frac{2}{1-\tilde{\alpha}})\cdot (A_2)^2
   +A_1+D\cdot\sqrt{\frac{1}{2}\tilde{\alpha}+\frac{\tilde{\alpha}}{1-\tilde{\alpha}}}\cdot A_2\\
   \leq & \frac{\sqrt{L}[F(\w^0)-F^*]}{\sqrt{T}}+\frac{\sqrt{L}(\frac{1}{2}\tilde{\alpha}+\frac{1}{1-\tilde{\alpha}})\cdot (A_2)^2}{\sqrt{T}} 
   +A_1+\tilde{\alpha}^\frac{1}{2}[\frac{3-\tilde{\alpha}}{2(1-\tilde{\alpha})}]^\frac{1}{2}\cdot D A_2.
 \end{align*}
 Note that $\tilde{\alpha} = 2\eta^2L^2\tau_{max}^2(1-\mu)^2(B-r)=O\left( \frac{L\tau_{max}^2(1-\mu)^2(B-r)}{T}\right)$, finally we have:
 \begin{align*}
  \frac{\sum_{t=0}^{T-1}\EB[\|\nabla F(\w^t)\|^2]}{T}
  \leq & O\left(\frac{\sqrt{L}\cdot[F(\w^0)-F^*]}{\sqrt{T}}\right)+O\left(\frac{\sqrt{L} (A_2)^2(1+\tilde{\alpha})}{\sqrt{T}}\right) \\
  &+O\left(\tilde{\alpha}^\frac{1}{2} D A_2\right)+A_1\\
  =& O\left(\frac{L^\frac{1}{2}[F(\w^0)-F^*]}{T^\frac{1}{2}}\right)+O\left(\frac{L^\frac{1}{2}\tau_{max}(1-\mu)(B-r)^\frac{1}{2}DA_2}{T^\frac{1}{2}}\right)\\
  &+O\left(\frac{L^\frac{1}{2}(A_2)^2}{T^\frac{1}{2}}\right)+O\left(\frac{L^\frac{5}{2}(A_2)^2\tau_{max}^2(1-\mu)^2(B-r)}{T^\frac{3}{2}}\right)+A_1.
 \end{align*}
 Specailly, when $B=O(r)$, we have: 
 \begin{align*}
 \frac{\sum_{t=0}^{T-1}\EB[\|\nabla F(\w^t)\|^2]}{T}
 \leq & O\left(\frac{L^\frac{1}{2}[F(\w^0)-F^*]}{T^\frac{1}{2}}\right)+O\left(\frac{L^\frac{1}{2}\tau_{max}DA_2r^\frac{1}{2}(1-\mu)}{T^\frac{1}{2}}\right)\\
 &+O\left(\frac{L^\frac{1}{2}(A_2)^2}{T^\frac{1}{2}}\right)+O\left(\frac{L^\frac{5}{2}(A_2)^2\tau_{max}^2 r(1-\mu)^2}{T^\frac{3}{2}}\right)+A_1.
 \end{align*}
 \end{proof}

\subsection{Proof of Proposition~\ref{prop:A1_and_D2}}
\begin{proof}
Under the condition that $\forall \w^t\in\RB^d,\ \EB[\|\G_{syn}^t-\nabla F(\w^t)\|~|~\w^t]\leq D$, we have:
\begin{align*}
  &\EB[\nabla F(\w^t)^T\G_{syn}^t~|~\w^t] \\
  =&\ \EB[\nabla F(\w^t)^T\ [\nabla F(\w^t)+(\G_{syn}^t-\nabla F(\w^t))~|~\w^t]\\
  =&\ \|\nabla F(\w^t)\|^2 +\EB[\nabla F(\w^t)^T(\G_{syn}^t-\nabla F(\w^t))~|~\w^t]\ \\
  \geq&\ \|\nabla F(\w^t)\|^2 - \|\nabla F(\w^t)\|\cdot\EB[\|\G_{syn}^t-\nabla F(\w^t)\|~|~\w^t]\\
  \geq&\ \|\nabla F(\w^t)\|^2 - D\times D\\
  =&\ \|\nabla F(\w^t)\|^2 - D^2.
\end{align*}
Combining with the property (i) of $(A_1,A_2)$-effective aggregation function, we have $A_1\leq D^2.$

\end{proof}

\subsection{Relation between Geometric Median and Centered Clipping}\label{subsec:proof_CC}
\begin{corollary}
  Aggregation rule centered clipping~(CC) is equivalent to geometric median~(geoMed) when clipping size $R\rightarrow 0^+$.
\end{corollary}
\begin{proof}
  The definition of CC is given by:

  \begin{equation}
    \h^{l+1}=\h^l+\frac{1}{B}\sum_{b=1}^B (\h_b - \h^l)\min\left(1,\frac{R}{\|\h_b-\h^l\|_2}\right).
  \end{equation}
  When CC converges to $\h^*_{CC}$, it means that 
  \begin{equation}
    \h^*_{CC}=\h^*_{CC}+\frac{1}{B}\sum_{b=1}^B (\h_b - \h^*_{CC})\min\left(1,\frac{R}{\|\h_b-\h^*_{CC}\|_2}\right).
  \end{equation}
  Thus, we have:
  \begin{equation}
    \sum_{b=1}^B (\h_b - \h^*_{CC})\min\left(1,\frac{R}{\|\h_b-\h^*_{CC}\|_2}\right)=\0.
  \end{equation}
  When $\forall b\in[B],\ R\leq\|\h_b-\h^*_{CC}\|_2$~(since $R\rightarrow 0^+$), we have 
  \begin{equation}
    \min\left(1,\frac{R}{\|\h_b-\h^*_{CC}\|_2}\right)=\frac{R}{\|\h_b-\h^*_{CC}\|_2}.
  \end{equation}
  Therefore, 
  \begin{equation}
    R\cdot\sum_{b=1}^B \frac{(\h_b - \h^*_{CC})}{\|\h_b-\h^*_{CC}\|_2}=\0.
  \end{equation}
  Namely,
  \begin{equation}
    R\cdot \left[\nabla\left.\left(\sum_{b=1}^B \|\h-\h_b\|_2\right)\right]\right|_{\h=\h^*_{CC}}=\0.
  \end{equation}
  Considering that the function $\sum_{b=1}^B \|\h-\h_b\|_2$ is convex, we have:

  \begin{equation}
    \h^*_{CC}=\mathop{\arg\min}_{\h\in\RB^d}\left\{ \sum_{b=1}^B \|\h-\h_b\|_2\right\}=\text{geoMed}([\h_1,\ldots,\h_B]).
  \end{equation}
\end{proof}

Meanwhile, we have to point out that although CC is theoretically equivalent to geoMed when $R$ is small enough, $R$ is not supposed to be set too small in practical applications. Too small $R$ will slow the convergence rate of CC.

\section{More Experimental Results}\label{appendix:more_exp_results}
\begin{figure}[h]
  \begin{center}
  \subfigure[BASGD with median]{
  \includegraphics[width=0.460\linewidth]{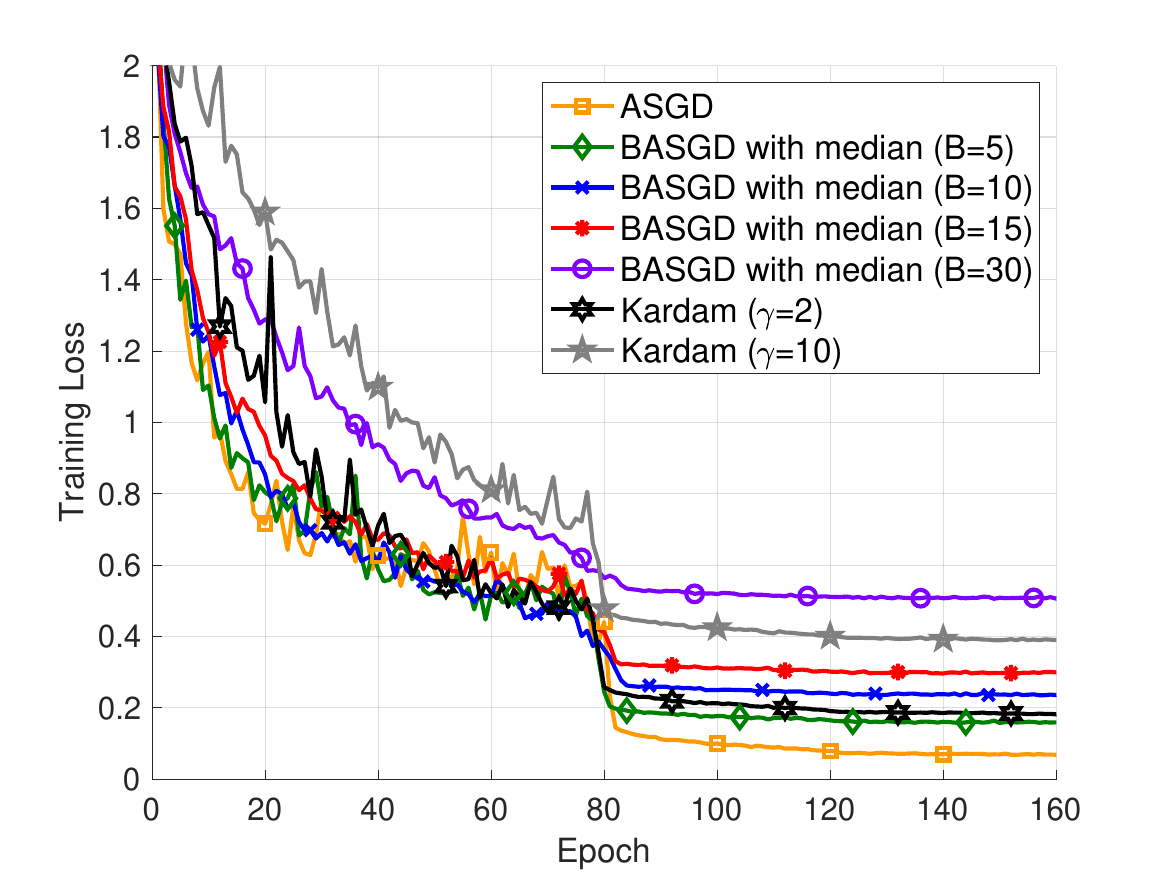}
  }
  \subfigure[BASGD with trmean]{
  \includegraphics[width=0.460\linewidth]{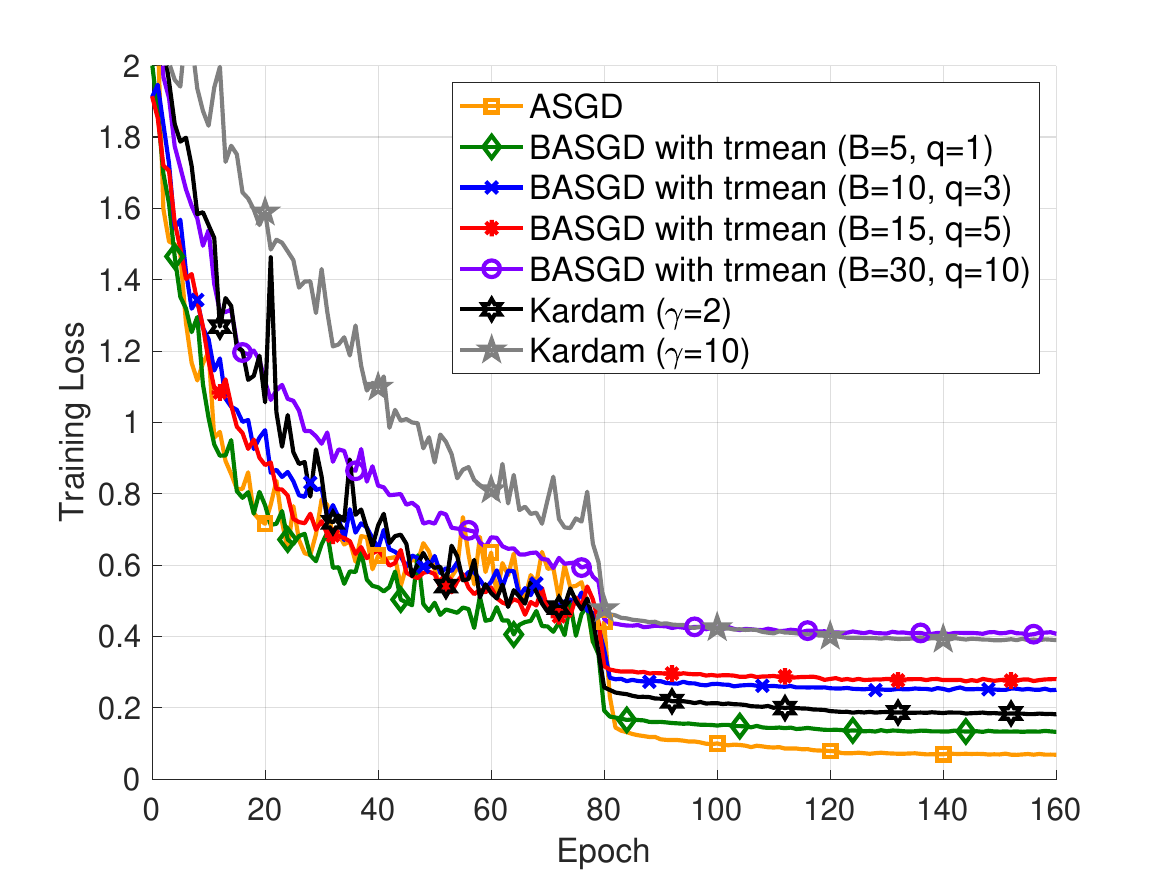}
  }
  \subfigure[BASGD with geoMed]{
  \includegraphics[width=0.460\linewidth]{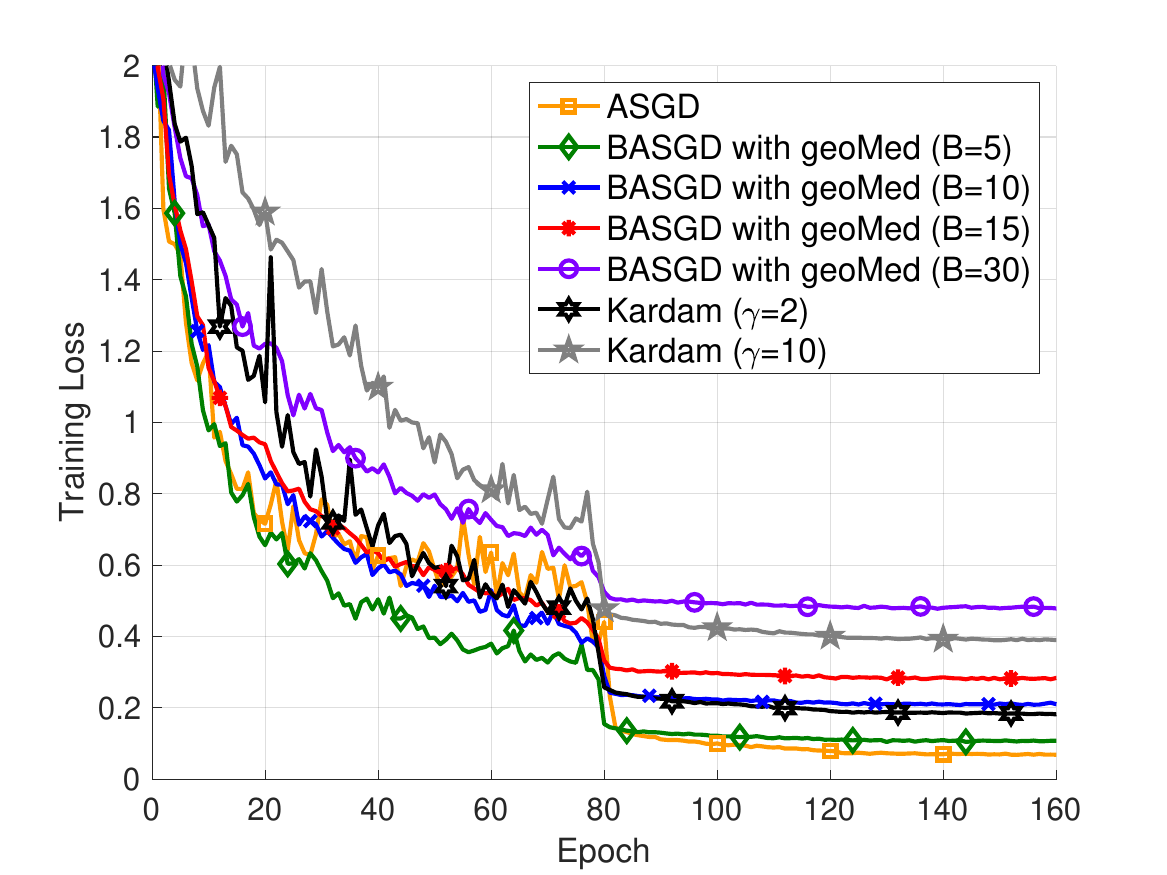}
  }
  \subfigure[BASGD with CC]{
  \includegraphics[width=0.460\linewidth]{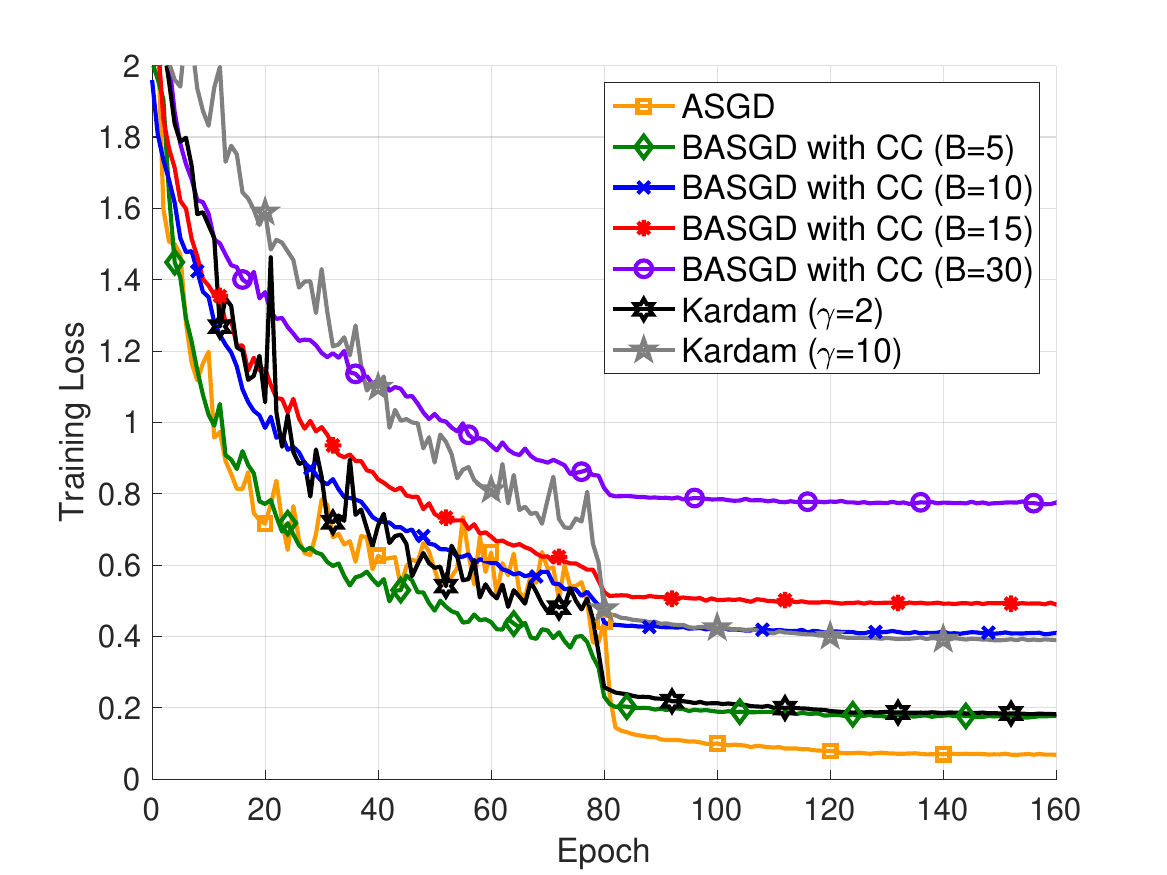}
  }
  \caption{Average training loss w.r.t. epochs of methods BASGD, ASGD, and Kardam when there are no Byzantine workers.
  }
  \label{fig:noAtk_loss}
  \end{center}
  \end{figure}

\begin{figure}[t]
  \begin{center}
  \subfigure[BASGDm with median]{
    \includegraphics[width=0.460\linewidth]{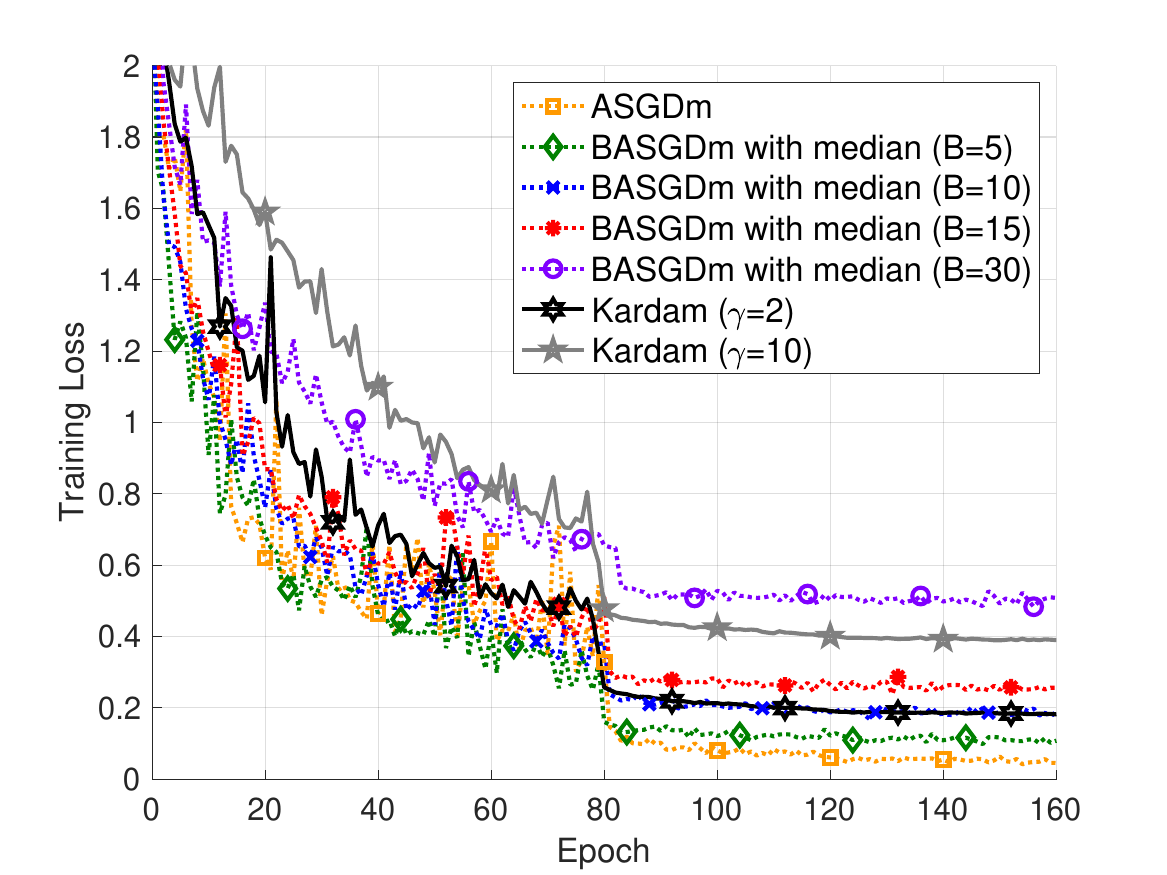}
  }
  \subfigure[BASGDm with trmean]{
    \includegraphics[width=0.460\linewidth]{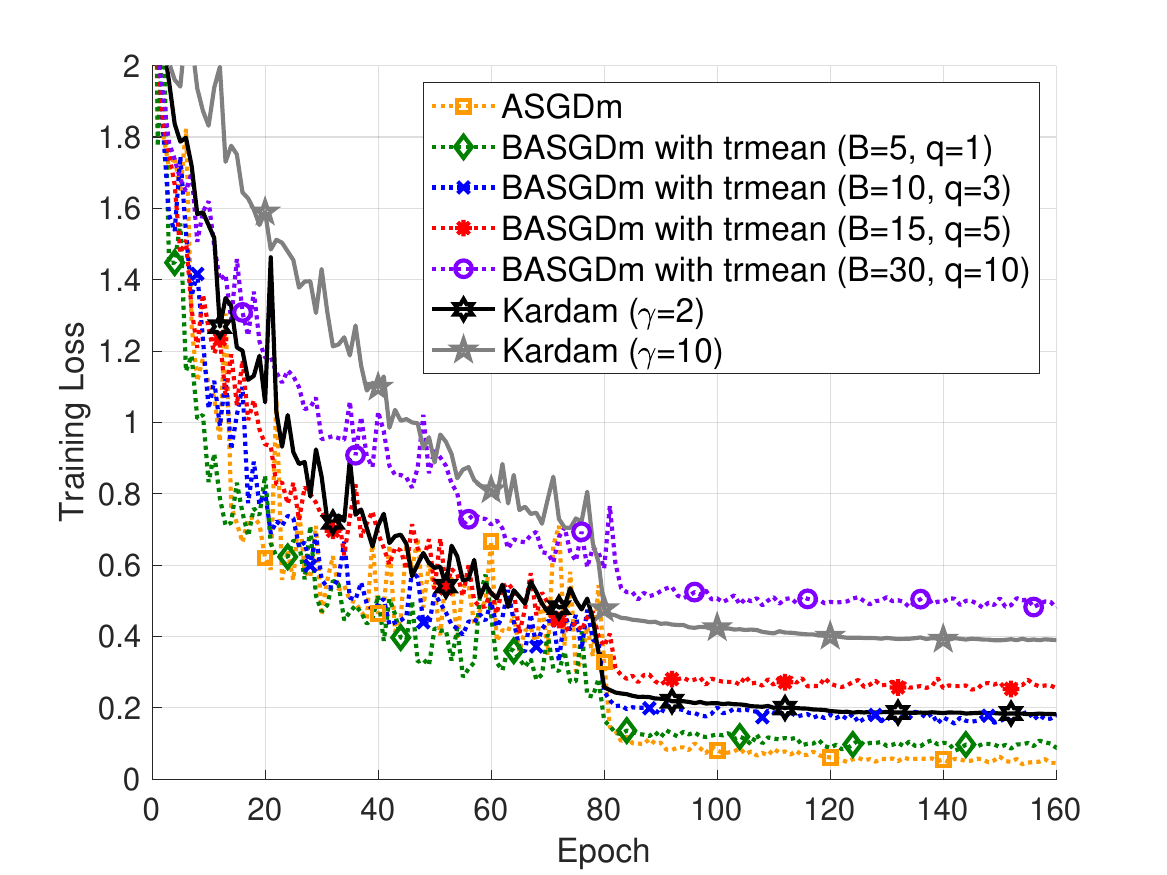}
  }
  \subfigure[BASGDm with geoMed]{
    \includegraphics[width=0.460\linewidth]{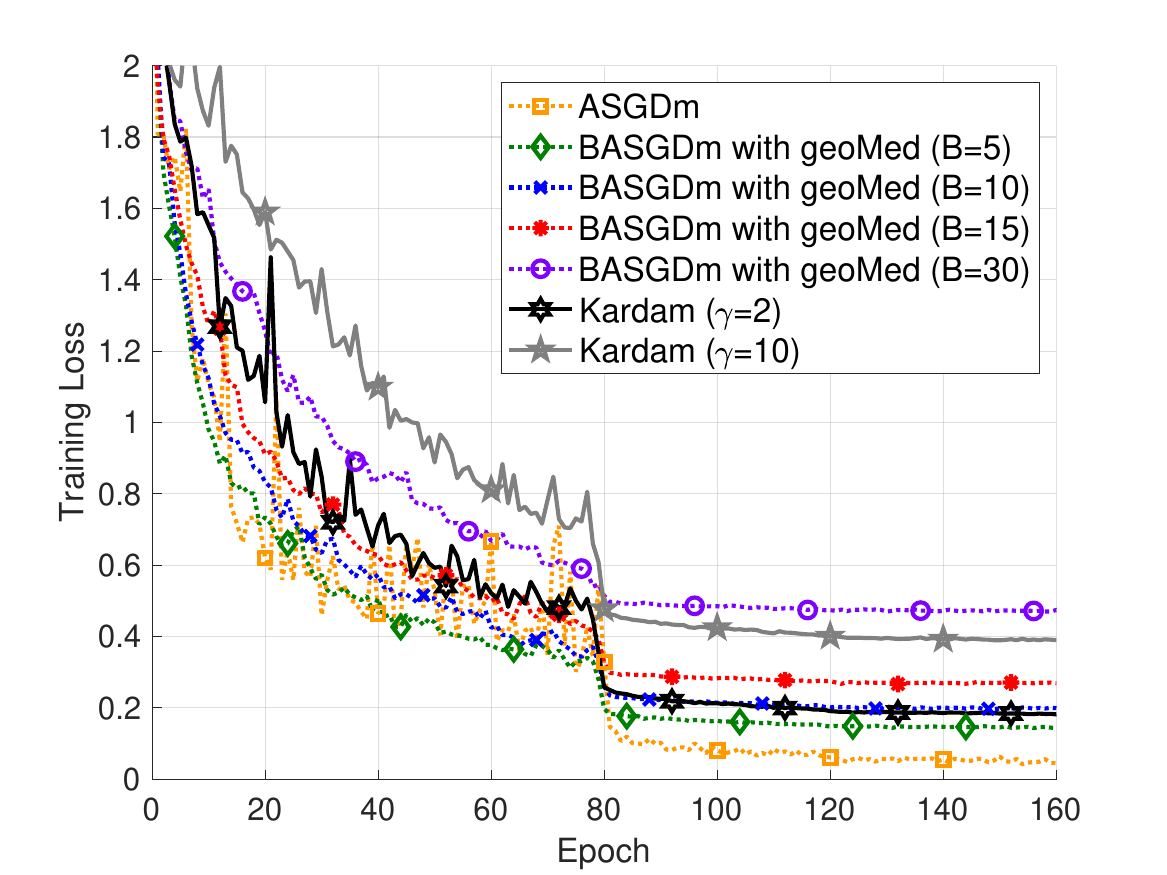}
  }
  \subfigure[BASGDm with CC]{
    \includegraphics[width=0.460\linewidth]{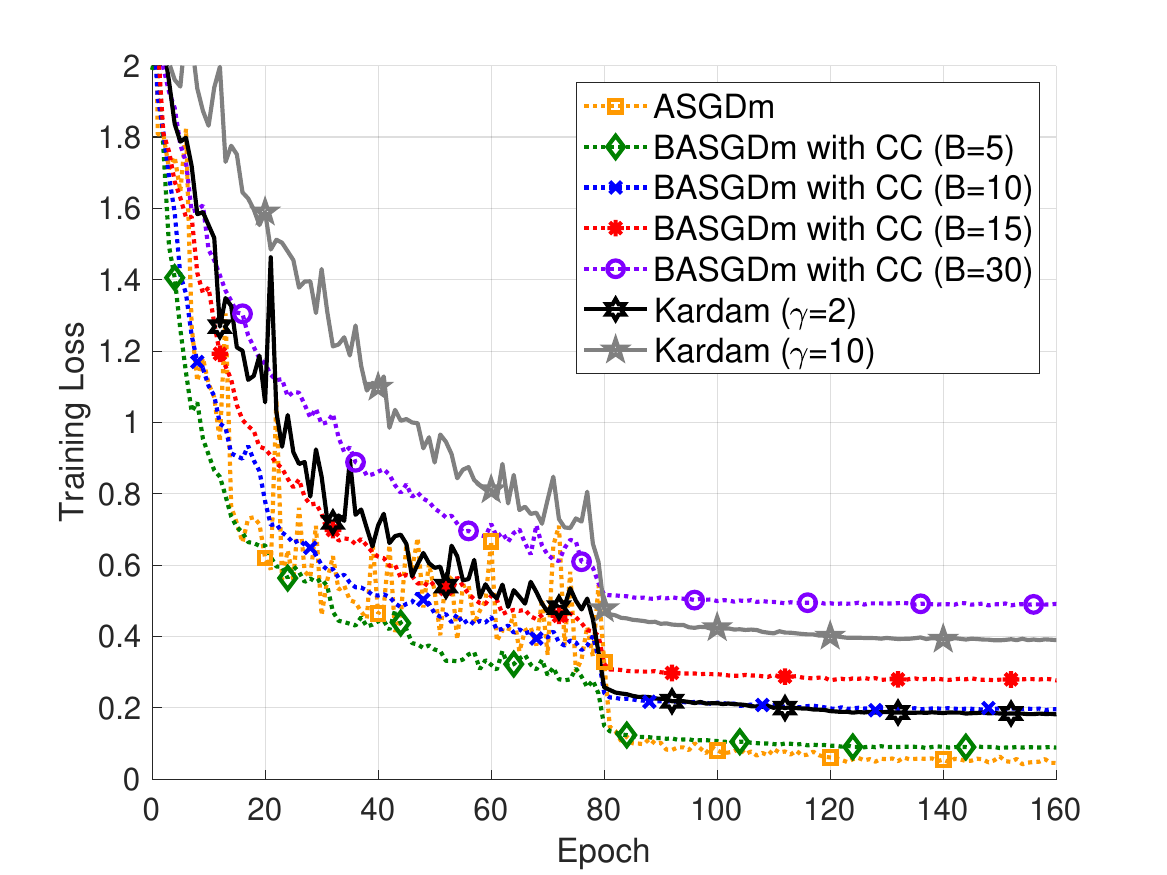}
  }
  \caption{Average training loss w.r.t. epochs of methods BASGDm, ASGDm, and Kardam when there are no Byzantine workers.
  }
  \label{fig:noAtk_loss_2}
  \end{center}
  \end{figure}

\begin{figure}[ht]
  \begin{center}
  \subfigure[$3$ Byzantine workers with RD-attack]{
  \includegraphics[width=0.460\linewidth]{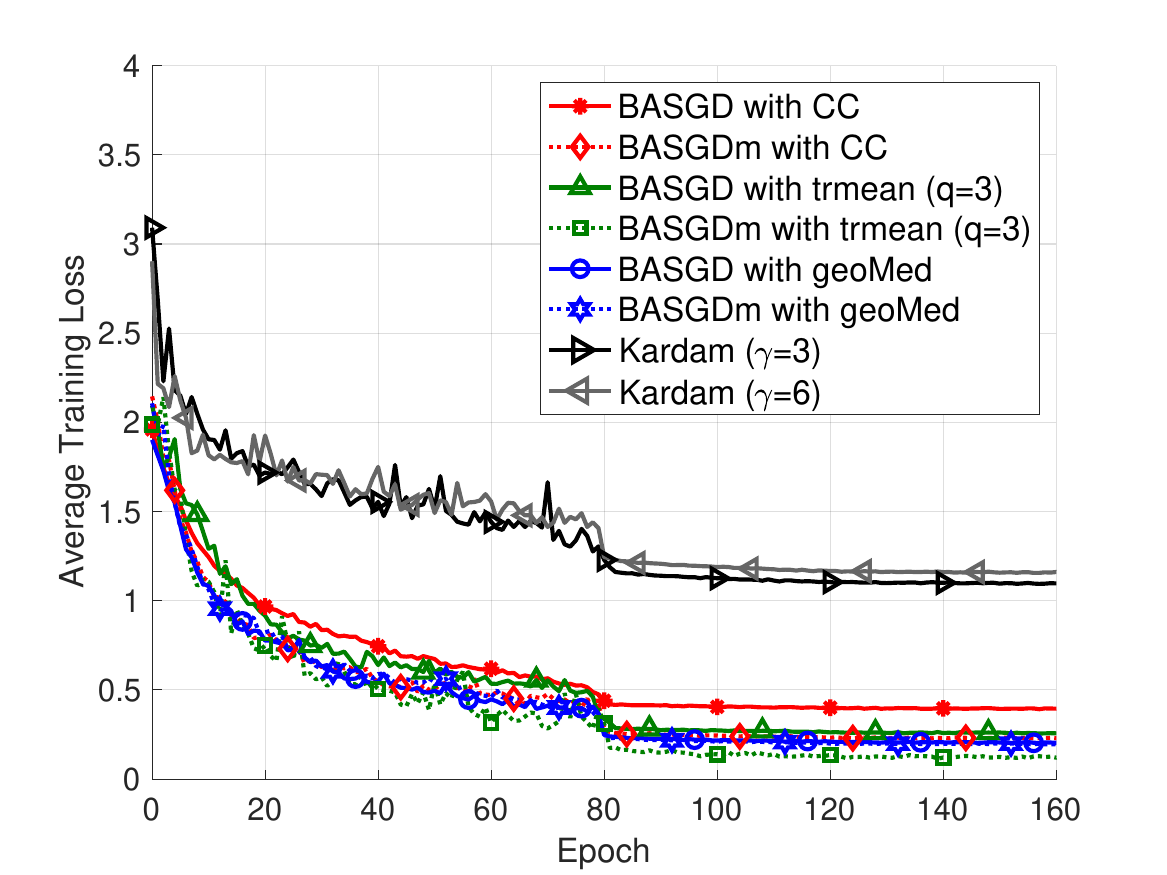}
  }
  \subfigure[$3$ Byzantine workers with NG-attack]{
  \includegraphics[width=0.460\linewidth]{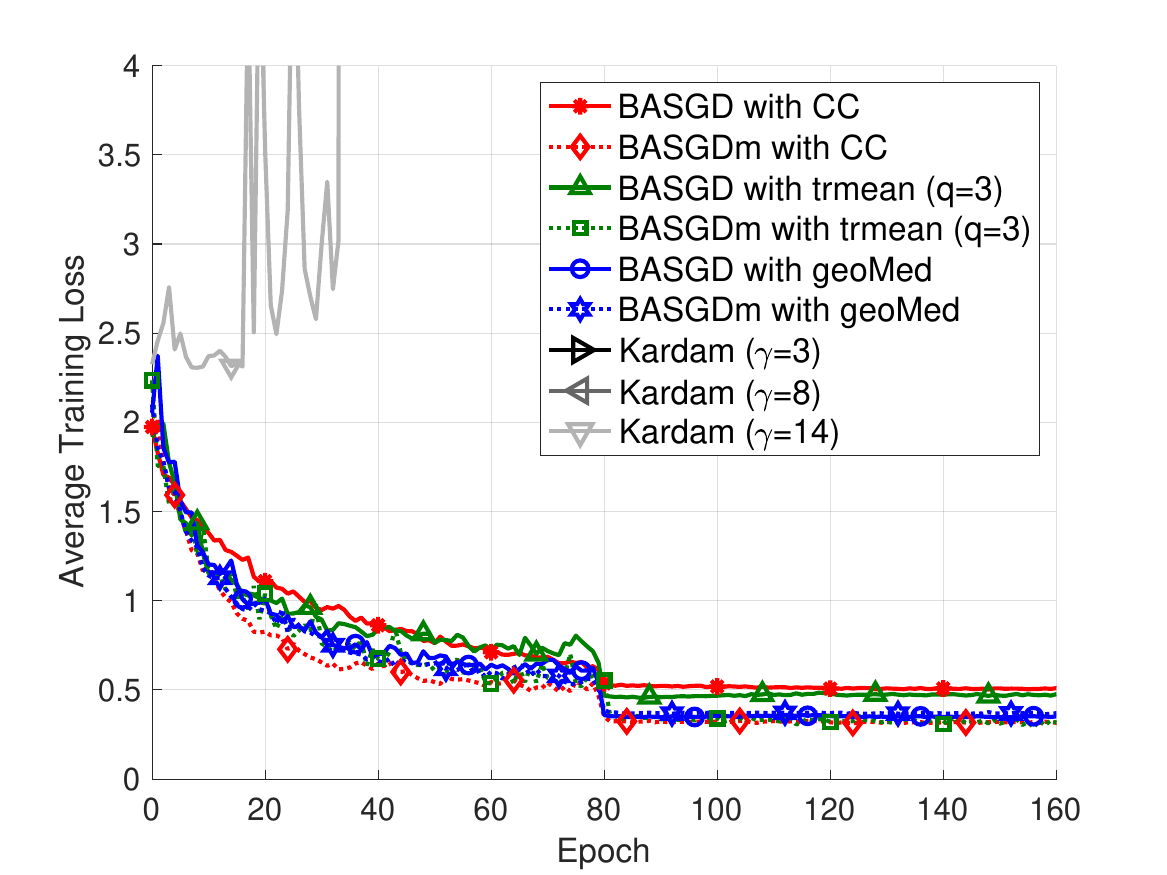}
  }
  \subfigure[$6$ Byzantine workers with RD-attack]{
  \includegraphics[width=0.460\linewidth]{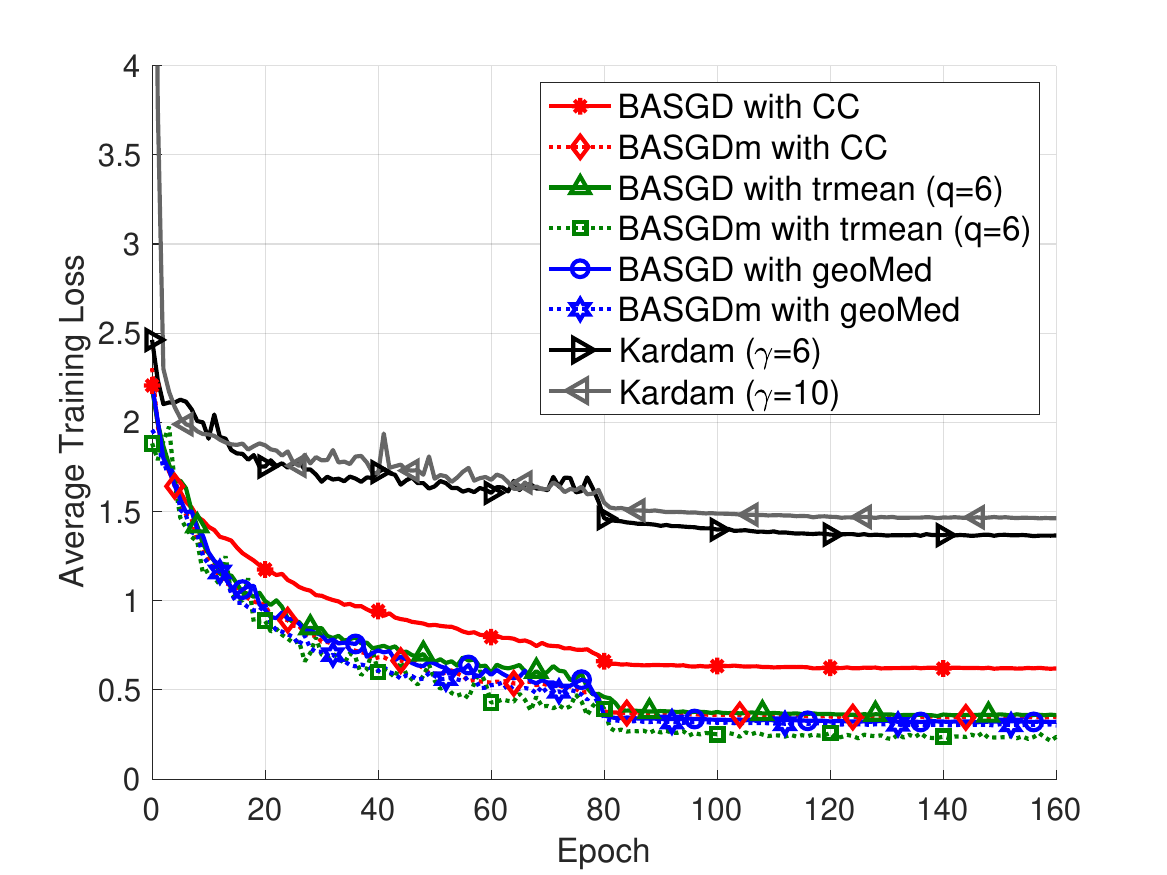}
  }
  \subfigure[$6$ Byzantine workers with NG-attack]{
  \includegraphics[width=0.460\linewidth]{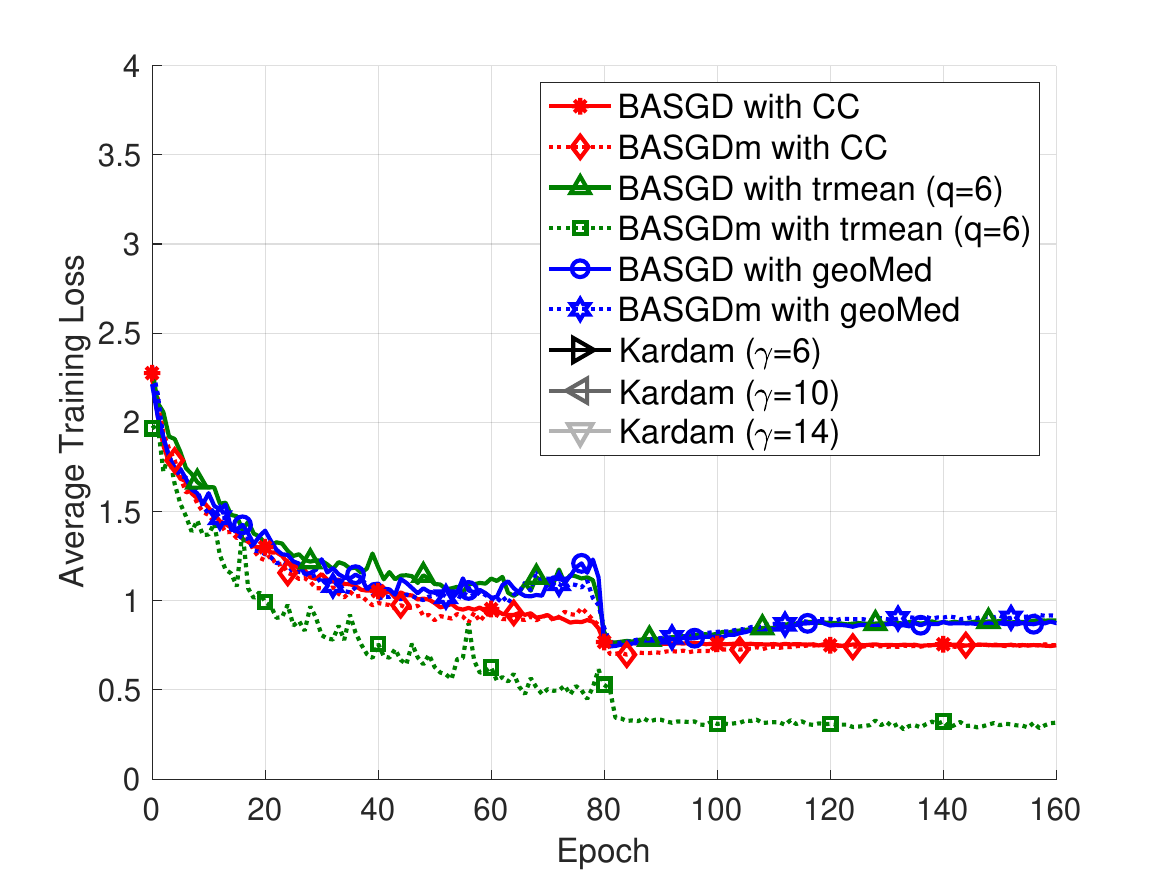}
  }
  \caption{Average training loss w.r.t. epochs under non-omniscient attacks. $B=10$ for BASGD (BASGDm) when there are $3$ Byzantine workers and $B=15$ for BASGD (BASGDm) when there are $6$ Byzantine workers. Some curves do not appear in the figure, because the value of loss function is extremely large.
  }\label{fig:IC_non_omniscient_loss}
  \end{center}
\end{figure}

\begin{figure}[t]
  \begin{center}
  \subfigure[$3$ Byzantine workers with FoE attack]{
  \includegraphics[width=0.460\linewidth]{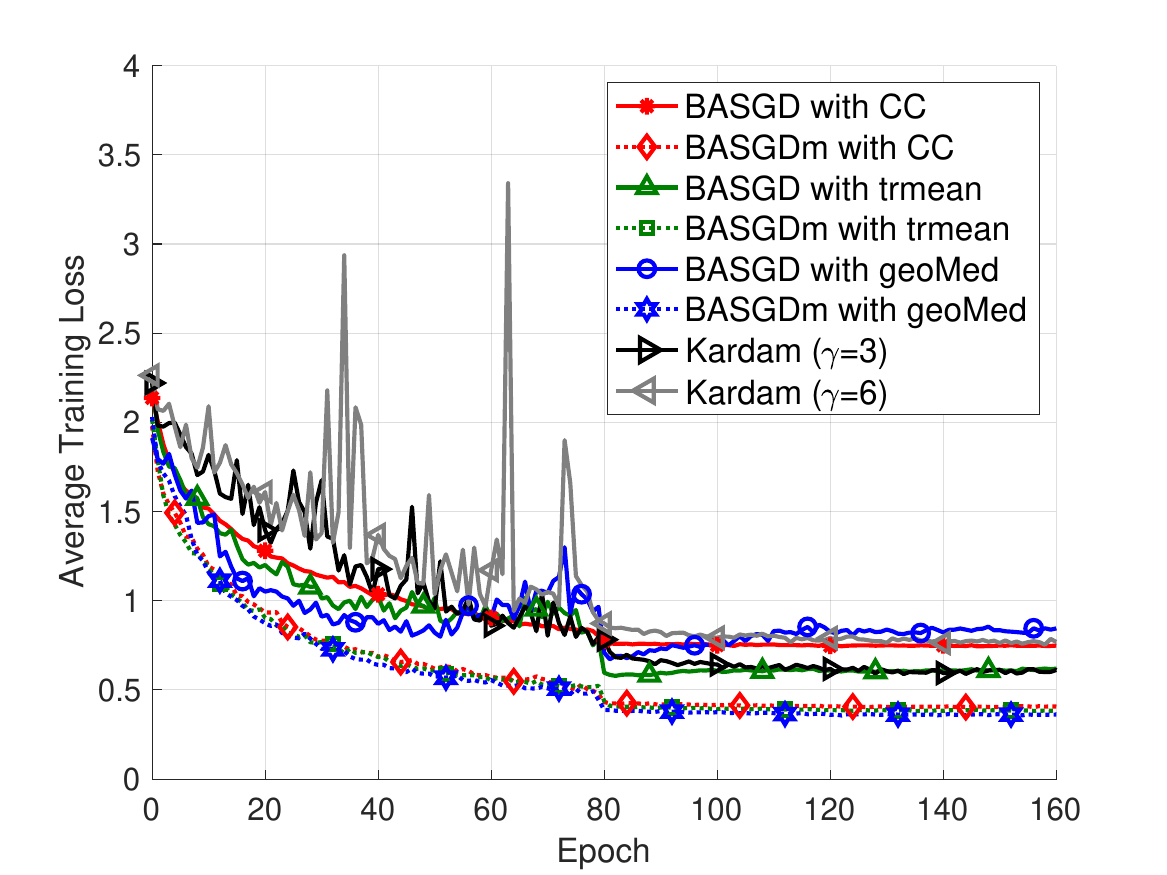}
  }
  \subfigure[$3$ Byzantine workers with ALIE attack]{
  \includegraphics[width=0.460\linewidth]{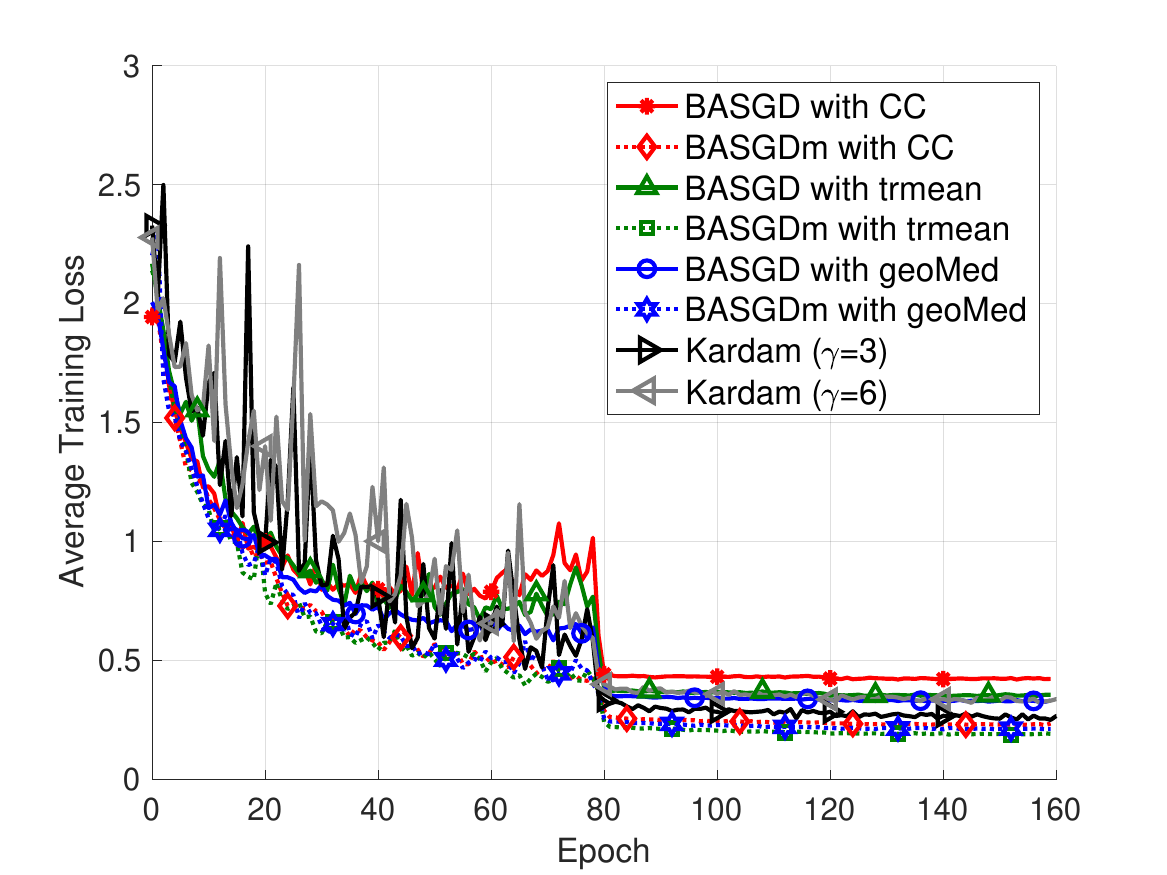}
  }
  \subfigure[$6$ Byzantine workers with FoE attack]{
  \includegraphics[width=0.460\linewidth]{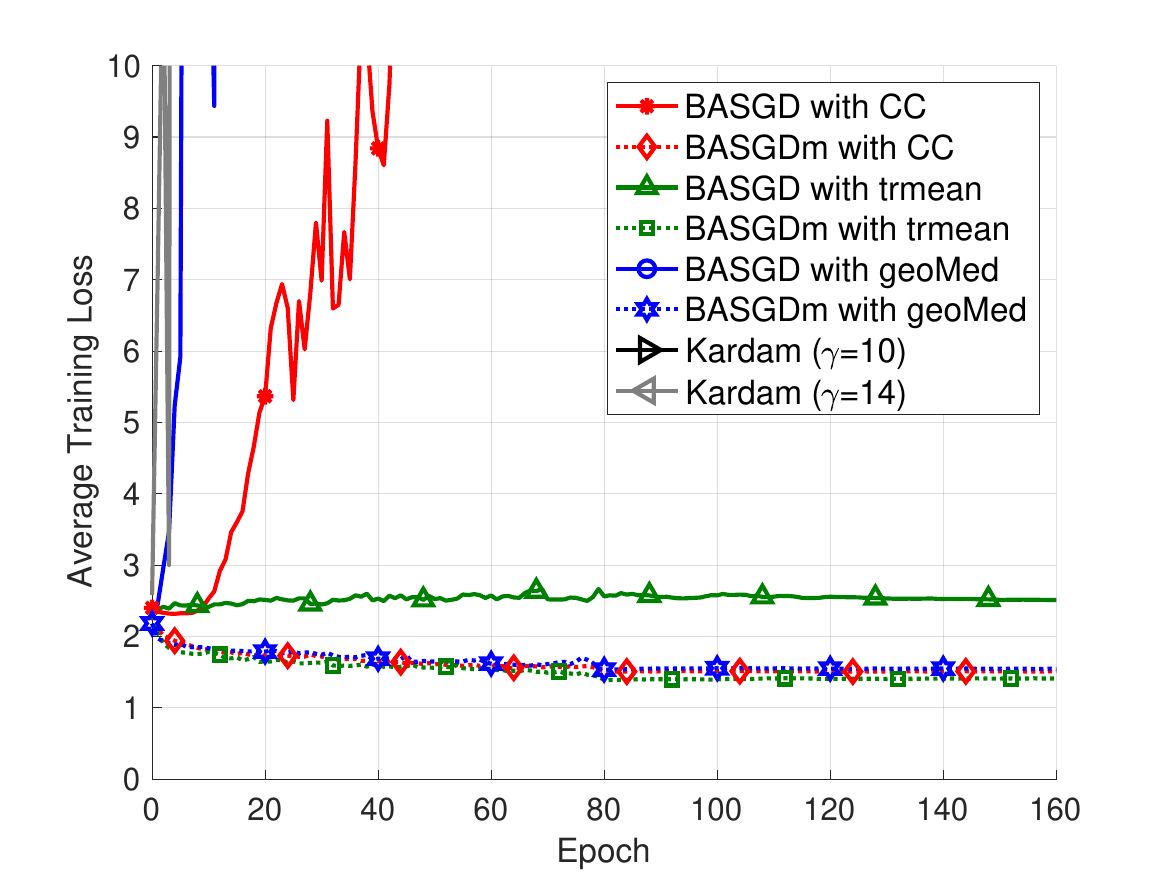}
  }
  \subfigure[$6$ Byzantine workers with ALIE attack]{
  \includegraphics[width=0.460\linewidth]{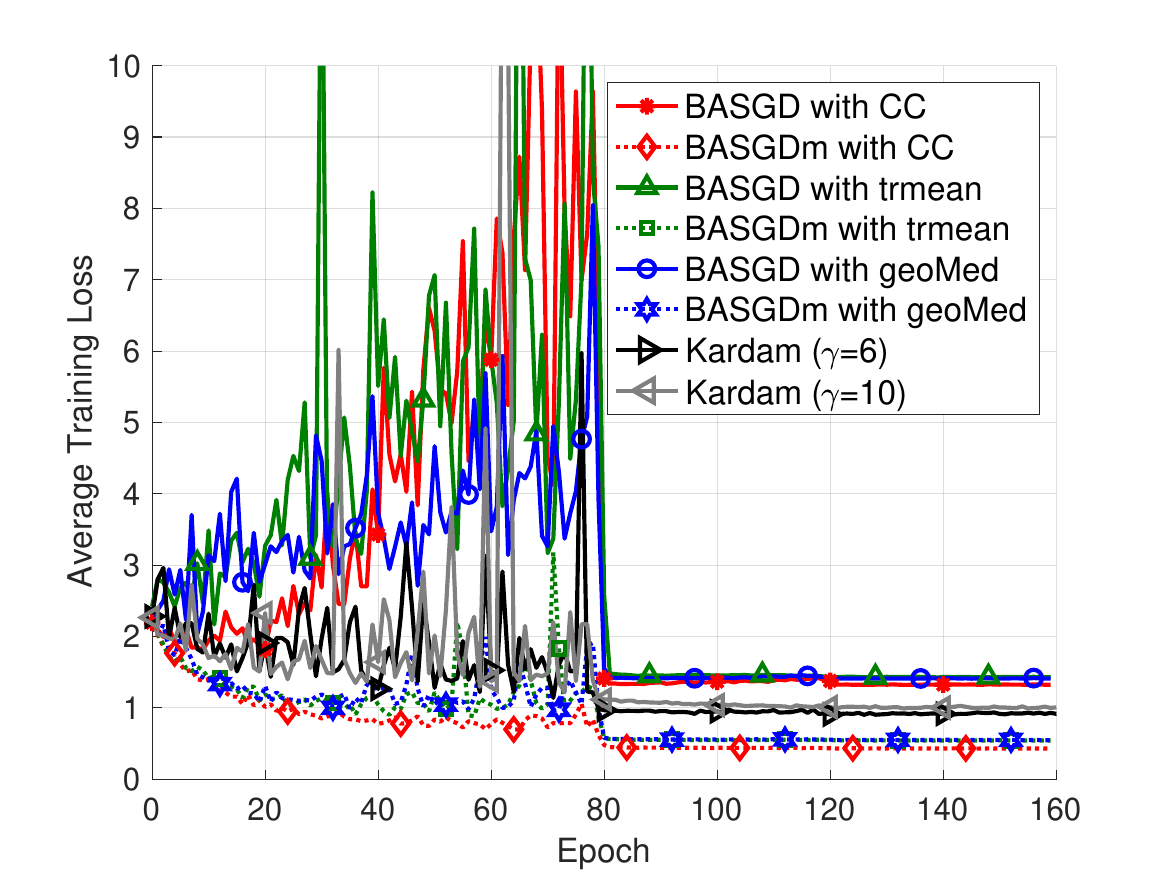}
  }
  \caption{Average training loss w.r.t. epochs under omniscient attacks. $B=10$ for BASGD (BASGDm) when there are $3$ Byzantine workers and $B=15$ for BASGD (BASGDm) when there are $6$ Byzantine workers. Some curves do not appear in the figure, because the value of loss function is extremely large.
  }\label{fig:IC_omniscient_loss}
  \end{center}
\end{figure}

Figure~\ref{fig:noAtk_loss}-\ref{fig:noAtk_loss_2}, Figure~\ref{fig:IC_non_omniscient_loss}, and Figure~\ref{fig:IC_omniscient_loss}
illustrate the average training loss w.r.t. epochs
when under no attack, non-omniscient attacks and omniscient attacks in the image classification task.
Please note that in Figure~\ref{fig:IC_non_omniscient_loss} and Figure~\ref{fig:IC_omniscient_loss},
some curves do not appear because the value of loss function is extremely large due to the Byzantine attack.
$\gamma$ is the hyper-parameter about the assumed number of Byzantine workers in Kardam.
The experimental results further support the conclusions of this work.

\clearpage
\bibliography{references}

\end{document}